\documentclass[twoside,11pt]{article}

\usepackage{blindtext}
\usepackage{jmlr2e}

\usepackage{enumitem}
\RequirePackage{amsmath,amsfonts}
\RequirePackage{graphicx}
\usepackage{booktabs}
\usepackage{indentfirst}
\usepackage{longtable}
\usepackage[section]{placeins}
\usepackage[utf8]{inputenc}
\usepackage{bbm}
\usepackage{mathrsfs} 
\usepackage{multirow}
\usepackage{float}
\usepackage{url} 
\usepackage{enumerate}
\usepackage{color}
\usepackage{times}
\usepackage[T1]{fontenc}
\usepackage{algorithm}
\usepackage{algorithmic}

\allowdisplaybreaks[4]

\DeclareMathOperator*{\argmin}{argmin}
\DeclareMathOperator*{\argmax}{argmax}

\DeclareMathOperator*{\Vol}{Vol}

\DeclareMathOperator*{\Var}{Var}
\def\T{{ \mathrm{\scriptscriptstyle T} }}
\newcommand{\RN}[1]{%
	\textup{\uppercase\expandafter{\romannumeral#1}}
}

\usepackage{lastpage}
\jmlrheading{25}{2024}{1-\pageref{LastPage}}{6/23; Revised
	11/23}{3/24}{23-0811}{Rui Qiu, Zhou Yu, and Ruoqing Zhu}
\ShortHeadings{Random Forest Weighted Local Fr\'echet Regression}{Qiu, Yu, and Zhu}
\firstpageno{1}

\begin{document}
	
	\title{Random Forest Weighted Local Fr\'echet Regression with Random Objects}

	\author{\name Rui Qiu 
		\email rqiu\_stat@outlook.com \\
		\addr School of Statistics, KLATASDS-MOE\\
		East China Normal University\\
		Shanghai 200062, China
		\AND
		\name  Zhou Yu\footnotemark[1]
		\email zyu@stat.ecnu.edu.cn \\
		\addr School of Statistics, KLATASDS-MOE\\
		East China Normal University\\
		Shanghai 200062, China
		\AND
		\name Ruoqing Zhu 
		\email rqzhu@illinois.edu \\
		\addr Department of Statistics\\
		University of Illinois at Urbana-Champaign\\
		Champaign, IL 61820, USA}
	
	\editor{Genevera Allen}
	
	\maketitle
	
	\renewcommand{\thefootnote}{\fnsymbol{footnote}}
    \footnotetext [1] {Corresponding author.}
	
	\begin{abstract}
		Statistical analysis is increasingly confronted with complex data from metric spaces.  \cite{petersen2019frechet} established a general paradigm of Fr\'echet regression with complex metric space valued responses and Euclidean predictors. However, the local approach therein involves nonparametric kernel smoothing and suffers from the curse of dimensionality. To address this issue, we in this paper propose a novel random forest weighted local Fr\'echet regression paradigm. The main mechanism of our approach relies on a locally adaptive kernel generated by random forests. Our first method uses these weights as the local average to solve the conditional Fr\'echet mean, while the second method performs local linear Fr\'echet regression, both significantly improving existing Fr\'echet regression methods. Based on the theory of infinite order U-processes and infinite order $M_{m_n}$-estimator, we establish the consistency, rate of convergence, and asymptotic normality for our local constant estimator, which covers the current large sample theory of random forests with Euclidean responses as a special case. Numerical studies show the superiority of our methods with several commonly encountered types of responses such as distribution functions, symmetric positive-definite matrices, and sphere data. The practical merits of our proposals are also demonstrated through the application to New York taxi data and human mortality data.
	\end{abstract}
	
	
	\section{Introduction} \label{sec1}
	In recent years, non-Euclidean statistical analysis has received increasing attention due to demands from modern applications, such as the covariance or correlation matrices for functional brain connectivity in neuroscience and probability distributions in CT hematoma density data. To this end,  \cite{hein2009robust} proposed nonparametric Nadaraya-Watson estimators for response variables being random objects,  which are random elements in general metric spaces that by default do not have a vector space structure.  \cite{petersen2019frechet} further introduced the general framework of Fr\'echet regression and established the methodology and theory for both global and local Fr\'echet regression analysis of complex random objects. \cite{chen2022uniform} continued to derive the uniform convergence rate of local Fr\'echet regression. \cite{yuan2012local} and \cite{lin2022additive} considered nonparametric modeling with responses being symmetric positive-definite matrices, which are a specific type of random object. These methods certainly build a concrete foundation of statistical modeling with non-Euclidean responses. However, methods mentioned above rely on nonparametric kernel smoothing and thus can be problematic when the dimension of predictor $X$ is relatively high, limiting the scope of Fr\'echet regression in real applications \citep{zhang2021dimension, ying2022frechet}. Recently, \cite{bhattacharjee2023single} and \cite{ghosal2023frechet} proposed single index Fr\'echet regression to resolve this dilemma but requires strong model assumptions.
	
	Random forest, as pioneered by Leo Breiman \citep{breiman2001random}, is a popular and promising tool for relatively high-dimensional statistical learning for Euclidean data. It is an ensemble model that combines the strength of multiple randomized trees. Moreover, trees can be generated parallelly, making random forests more attractive computationally. Random forests demonstrate substantial gains in various learning tasks compared to classical statistical methods, such as survival analysis \citep[e.g.,][]{ishwaran2008random, steingrimsson2019censoring}.  Theoretical research into random forests has gained considerable momentum in recent years due to their tremendous popularity. \cite{biau2008consistency} first proved the consistency of purely random forests for classification. For regression problems,  \cite{genuer2012variance} and \cite{arlot2014analysis} further made a complete analysis of the variance and bias of purely random forests. \cite{biau2012analysis} and \cite{gao2020towards} established the consistency and convergence rate of the centered random forests for regression and classification, respectively.  \cite{duroux2018impact} provided the convergence rate of $q$-quantile random forests. In particular,  \cite{klusowski2021sharp} improved the rate of the median random forests.  \cite{scornet2015consistency} proved the {$L^{2}$ consistency} of Breiman's original random forests for the first time under the assumption of additive model structure.  \cite{mentch2016quantifying} formulated random forests as infinite order incomplete U-statistics and studied their asymptotic normality.  \cite{wager2018estimation} further established the central limit theorem for random forests based on honest tree construction.
	
	However, most methodology developments, theoretical investigations, and real applications of random forests focus on classical Euclidean responses and predictors. In a recent study by \cite{yao2022ensemble}, a general framework based on forests was introduced for estimating a survival function considering time-varying covariates. Additionally, there is a significant interest in generalizing the random forests with metric space valued responses, which is expected to work better than existing Fr\'echet regression methods when the predictor dimension is moderately large. To this end, \cite{capitaine2019fr} proposed Fr\'echet trees and Fr\'echet random forests based on regression trees and Breiman's random forests. On the other hand, recent developments \citep[e.g.,][]{lin2006random, meinshausen2006quantile, bloniarz2016supervised, athey2019generalized, friedberg2020local} reveal the fact that random forests implicitly construct a kernel-type weighting function. This proliferation of work points toward a general synthesis between the core of nonparametric kernel smoothing and the ability to encompass locally data-adaptive weighting by random forests. Taking a step forward, we in this paper propose a novel random forest weighted local Fr\'echet regression paradigm with superior performance and desirable statistical properties. 
	
	Our major contributions are summarized from the following three perspectives. First, to the best of our knowledge, this is the first attempt to adopt random forests as a kernel for Fr\'echet regression. Our proposal called random forest weighted local constant Fr\'echet regression articulates a new formulation of Fr\'echet regression based on random forests that has an intrinsic relationship with classical nonparametric kernel regression. Second, compared to \cite{capitaine2019fr}, our method is more concise in terms of formulation, which allows us to take a substantial step towards the asymptotic theory of local Fr\'echet regression based on random forests rigorously. Following the line of research introduced by \cite{wager2018estimation} based on trees with honesty and other properties, the consistency and rate of convergence are derived based on the theory of infinite order $U$-statistics and $U$-processes. To study the asymptotic normality, we extend the current theory of finite order $M_m$-estimator to infinite order $M_{m_n}$-estimator. The new technical tools developed to establish the central limit theorem of infinite order $M_{m_n}$-estimator can be of independent interest. And our asymptotic normality result also covers that of random forests with Euclidean responses \citep{wager2018estimation} as a special case. Last but not least, the perspective from which we view the random forests facilitates the generalization of our method to the local linear version  \citep{bloniarz2016supervised, friedberg2020local}. This extension achieves better smoothness of the resulting estimator. The random forest weighted local constant Fr\'echet regression and local linear Fr\'echet regression collectively make up a coherent system and a new framework for Fr\'echet regression.
	
	The rest of the paper is organized as follows. In Section~\ref{sec2}, we give an overview of Fr\'echet regression and introduce the random forest weighted local constant Fr\'echet regression (RFWLCFR) method. In Section~\ref{sec3}, we establish the consistency and develop other asymptotic theories of RFWLCFR. In Section~\ref{sec4}, we present the random forest weighted local linear Fr\'echet regression (RFWLLFR) approach as the generalization of RFWLCFR and confirm its consistency in estimation. In {Section~\ref{add sec}, a novel measure of variable importance is introduced.} In Section~\ref{sec5}, we conduct comprehensive simulation studies to examine our proposals in different settings, including probability distributions, symmetric positive-definite matrices, and spherical data. In Section~\ref{sec6}, we apply our methods to the New York taxi data, where the response is a taxi ride network, and human mortality data, where the response is age-at-death distribution. Section~\ref{sec7} concludes the paper with some discussions. All proofs and additional material are presented in the appendices.

	\section{Proposed Method} \label{sec2}
	Before formally presenting our first method in Section~\ref{sec2.2}, we undertake some preliminary preparations. This encompasses furnishing a background introduction to Fr\'echet regression and explaining the process of constructing Fr\'echet trees, which is a fundamental constituent of our method.
	\subsection{Preliminaries}
	\subsubsection{Fr\'echet Regression}\label{sec2.1}
	Let $(\Omega, d)$ be a metric space equipped with a specific metric $d$. Let $\mathcal{R}^{p} $ be the $p$-dimensional Euclidean space. We consider a random pair $(X, Y) \sim F$, where $X \in \mathcal{R}^{p}$, $Y \in \Omega$ and $F$ is the joint distribution of $(X, Y)$. We denote the marginal distributions of $X$ and $Y$ as $F_X$ and $F_Y$, respectively. The conditional distributions $F_{X \mid Y}$ and $F_{Y \mid X}$ are also assumed to exist.  When $\Omega \subseteq \mathcal{R}$, the target of classical regression is to estimate the conditional mean
	$$
	m(x)=E \big(Y \mid X=x\big)=\underset{y \in \mathcal{R}}{\argmin} E\Big\{\big(Y-y\big) ^{2}\mid X=x\Big\}.
	$$
	By replacing the Euclidean distance with the intrinsic metric $d$ of $\Omega$,  conditional Fr\'echet mean \citep{petersen2019frechet} can then be defined as
	\begin{align*}
		m_{\oplus}(x)=\underset{y \in \Omega}{\argmin} M_{\oplus}(x,y)=\underset{y \in \Omega}{\argmin} E\big\{d^{2}(Y, y) \mid X=x\big\}.
	\end{align*}
	
	Given an i.i.d training sample $\mathcal{D}_{n}=\left\{\left(X_{i}, Y_{i}\right)\right\}_{i=1}^n$ with $(X_i, Y_i)\sim F$, the goal of Fr\'echet regression is to estimate $m_{\oplus}(x)$ in the sample level.  For this purpose, \cite{hein2009robust} generalized the Nadaraya-Watson regression to the Fr\'echet version as
	\begin{align*}
		\hat{m}_{\oplus}^{\textup{NW}}(x)=\underset{y \in \Omega}{\operatorname{argmin}} \frac{1}{n} \sum_{i=1}^{n} K_{h}\left(X_{i}-x\right) d^{2}\left(Y_{i}, y\right),
	\end{align*}
	where $K$ is a smoothing kernel such as the  Epanechnikov kernel or Gaussian Kernel and $h$ is a bandwidth, with $K_{h}(\cdot)=h^{-1} K(\cdot / h)$. \cite{petersen2019frechet}
	recharacterized the standard multiple linear regression and local linear regression as
	a function of weighted Fr\'echet means, and proposed global Fr\'echet regression and local Fr\'echet regression as
	\begin{align*}
		\hat{m}_{\oplus}(x)=\underset{y \in \Omega}{\operatorname{argmin}}  \frac{1}{n}\sum_{i=1}^{n} s_{i n}(x) d^{2}\left(Y_{i}, y\right),
	\end{align*}
	where $s_{i n}(x)$ has different expressions for global and local Fr\'echet regression. 
	
\begin{figure}[ht!]
		\centering
		\includegraphics[width = 1\linewidth]{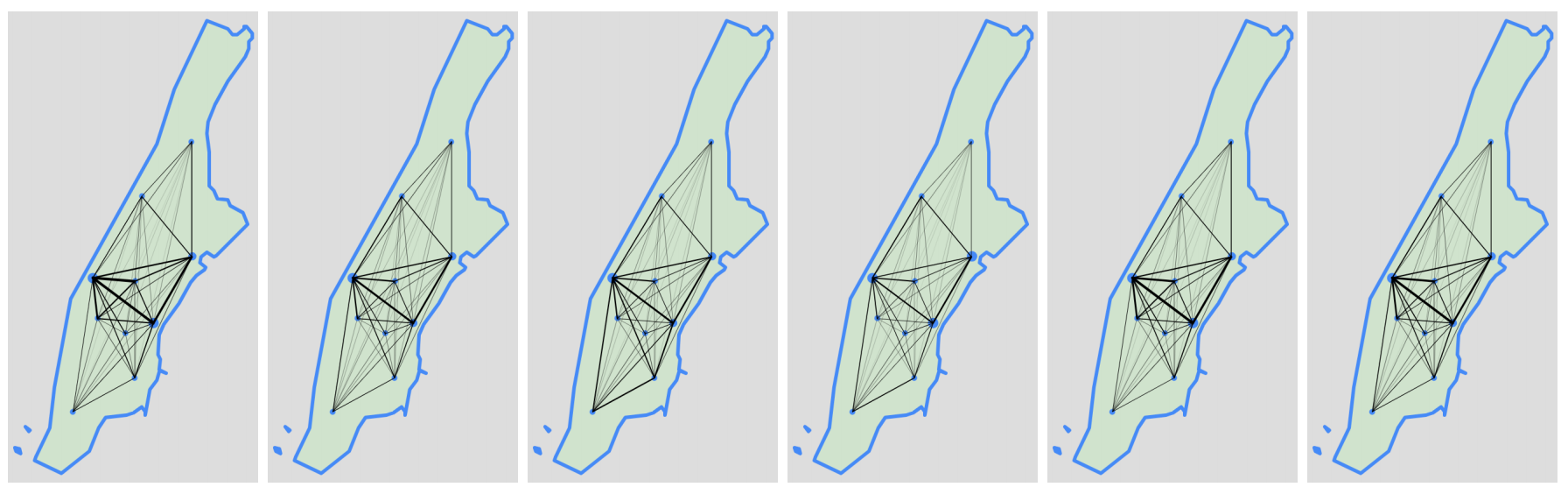}
		\caption{The first plot illustrates the flow statistics of yellow taxis in ten distinct zones of Manhattan, New York, during a certain time period. The thickness of the edges connecting vertices corresponds to the level of inter-zone traffic, while the size of vertices represents the total traffic volume within each zone. The remaining five plots from left to right are the predictions given by the global Fr\'echet regression, local Fr\'echet regression after dimension reduction, single index Fr\'echet regression, RFWLCFR and RFWLLFR.}
		\label{taxi network}
	\end{figure}

	The Nadaraya-Watson Fr\'echet regression and local Fr\'echet regression both involve kernel weighting function $K$ in the estimation procedure, which limits their applications when $p\geq 3$. To address this issue, we aim to borrow the strength of random forests to generate a more powerful weighting function for moderately large $p$. Figure~\ref{taxi network} depicts flow statistics and predictive outcomes of yellow taxi traffic in Manhattan, New York, while further details are provided in Section~\ref{sec6}. This problem can be formulated as a Fr\'echet regression problem where the response variable is a network (matrix) and $14$ predictor variables are considered. Notably, when $p=14$, both the Nadaraya-Watson Fr\'echet regression and local Fr\'echet regression techniques exhibit significant limitations. Here we employ Fr\'echet sufficient dimension reduction \citep{ying2022frechet} to help realize the local Fr\'echet regression. The global Fr\'echet regression, although not restricted by dimensionality, relies on the assumption of linearity for satisfactory performance. Instead, it is evident from Figure~\ref{taxi network} that the two methods we will propose in this paper have a higher prediction accuracy than global Fr\'echet regression.
	
	\subsubsection{Fr\'echet Trees}\label{Frechet Trees}
	A regression tree $T$ splits the input space recursively from the root node (the entire input space). At each split, the parent node is divided into two child nodes along a certain feature direction and a certain cutoff point, which are decided by a specific splitting criterion. After many splits, the child node becomes small enough to form a leaf node, and the sample data points within the leaf node are used to estimate the conditional (Fr\'echet) mean. 
	
	In this paper, we use Fr\'echet trees to refer to regression trees that handle metric space valued responses, regardless of their splitting criterion. Here we introduce an adaptive criterion---variance reduction splitting criterion, which uses information from both the predictor $X=(X^{(1)}, \ldots, X^{(p)})$ and response $Y$ in the node splitting decision. The impurity of $Y$ from a general metric space is no longer measured by the variance under the Euclidean distance. Instead, we use the Fr\'echet variance. A split on an internal node $A$ can be represented by a pair $(j, c),j\in\{1, \ldots,p\}$, indicating that $A$ is split at position $c$ along the direction of feature $X^{(j)}$.  We select the optimal $(j_{n}^{*},c_{n}^{*})$ to decrease the sample Fr\'echet variance as much as possible so that the sample points grouped in the same child node exhibit a high degree of similarity. Specifically, the splitting criterion is
	\begin{align*}
		\mathcal{L}_{n}(j, c)=\frac{1}{N_{n}(A)}\Bigg\{&\sum_{i: X_{i} \in A} d^{2}\left(Y_{i}, \bar{Y}_{A}\right)-\sum_{i: X_{i} \in A_{j,l}} d^{2}\left(Y_{i}, \bar{Y}_{A_{j,l}}\right)-\sum_{i: X_{i} \in A_{j,r}} d^{2}\left(Y_{i}, \bar{Y}_{A_{j,r}}\right)\Bigg\},
	\end{align*}
	where $A_{j,l}=\left\{x \in A: x^{(j)}<c\right\}, A_{j,r}=\left\{x \in A: x^{(j)} \geq c\right\}$, ${N_{n}(A)}$ is the number of samples falling into the node $A$, and $\bar{Y}_{A}={\argmin}_ {y \in \Omega}\sum_{i: X_{i} \in A} d^{2}\left( Y_{i},y\right)$, \emph{i.e.}, the sample Fr\'echet mean of $Y_i$'s associated to the samples belonging to the node $A$. $\bar{Y}_{A_{j,l}}$ and $\bar{Y}_{A_{j,r}}$ are defined similarly. Then the optimal split pair is decided by
	\begin{align*}
		(j_{n}^{*},c_{n}^{*})=\underset{j,c}{\argmax}\ \mathcal{L}_{n}(j, c).
	\end{align*}
	
	\subsection{Local Constant Method} \label{sec2.2}
	A single tree model may suffer from large bias or large variance depending on the tuning. To improve the predictive accuracy, we can aggregate multiple trees to form a random forest. The prediction error of random forests is closely related to the correlation among different trees. In addition to resampling the training data set for the growing of individual trees, auxiliary randomness is often introduced to further reduce the correlation between trees and thus improve the performance of random forests. For example, a subset of features is randomly selected before each split, and the split direction is designed based on the subset only. Here, we denote $\xi \sim \Xi$ as a source of auxiliary randomness.
	
	We first consider the classical random forests with Euclidean responses. And each tree is trained on a subsample $\mathcal{D}_{n}^b=\left\{\left(X_{i_{b,1}}, Y_{i_{b,1}}\right),\left(X_{i_{b,2}}, Y_{i_{b,2}}\right), \ldots, \left(X_{i_{b,s_{n}}}, Y_{i_{b,s_{n}}}\right)\right\}$ of the training data set $\mathcal{D}_{n}$, with $1 \leq i_{b,1}<i_{b,2}<\ldots<i_{b,s_{n}} \leq n$. Throughout the paper, we assume that the subsample size $s_n \rightarrow +\infty$ and $s_n/n \rightarrow 0$ as $n$ tends to infinity.
	Data resampling is done here without replacement (see \cite{scornet2015consistency, mentch2016quantifying, wager2018estimation}). The $b$th tree $T_{b}$ constructed by $\mathcal{D}_{n}^b$ and a random draw $\xi_{b} \sim \Xi$ gives an estimator of $m(x)$
	\begin{align*}
		T_{b}(x;\mathcal{D}_{n}^b,\xi_{b})=\frac{1}{N(L_{b}(x; \mathcal{D}_{n}^b,\xi_{b}))} \sum_{i: X_{i} \in L_{b}(x; \mathcal{D}_{n}^b,\xi_{b})}Y_{i},
	\end{align*}
	where $N(L_{b}(x; \mathcal{D}_{n}^b,\xi_{b}))$ is the number of samples in $L_{b}(x; \mathcal{D}_{n}^b,\xi_{b})$, the leaf node containing $x$ of $T_b$.
	For the random forest constructed by $B$ randomized trees,  $m(x)$ can be estimated by 
	\begin{align}\label{RF-1}
		\hat{r}(x)&=\frac{1}{B} \sum_{b=1}^{B}T_{b}(x;\mathcal{D}_{n}^b,\xi_{b})=\frac{1}{B} \sum_{b=1}^{B}\frac{1}{N(L_{b}(x;\mathcal{D}_{n}^b,\xi_{b}))} \sum_{i: X_{i} \in L_{b}(x;\mathcal{D}_{n}^b,\xi_{b})}Y_{i}.
	\end{align}
	In fact, we can view the random forest from another perspective and regard it as a weighted average of the training responses like
	\begin{align}\label{RF-2}
		\hat{r}(x)=\sum_{i=1}^{n} \alpha_{i}\left(x\right) Y_{i},
	\end{align}
	where $\alpha_{i}\left(x\right)=\frac{1}{B} \sum_{b=1}^{B} \frac{1\left\{X_{i} \in L_{b}(x; \mathcal{D}_{n}^b,\xi_{b})\right\}}{N(L_{b}(x;\mathcal{D}_{n}^b,\xi_{b}))}$ is defined as the random forest kernel.
	
	{We now generalize the Euclidean random forests to the Fr\'echet version when $\Omega$ is a general metric space. The generalization process involves two distinct approaches, each aligning with one of the two perspectives presented in \eqref{RF-1} and \eqref{RF-2}. We first rewrite the explicit expression of the random forest estimator \eqref{RF-1} as the implicit minimizer of some objective function
		\begin{align*}
			\hat{r}(x)=\underset{y \in \mathcal{R}}{\argmin}\frac{1}{B} \sum_{b=1}^{B}\Bigg[\underset{y^{\prime} \in \mathcal{R}}{\argmin}\Bigg\{\frac{1}{N(L_{b}(x; \mathcal{D}_{n}^b,\xi_{b}))} \sum_{i: X_{i} \in L_{b}(x; \mathcal{D}_{n}^b,\xi_{b})}\left(Y_{i}-y^{\prime}\right)^2\Bigg\}-y\Bigg]^2.
		\end{align*}
		Then the first generalization for metric space valued responses is simply replacing the Euclidean distance by the metric $d$ of $\Omega$, that is,
		\begin{align} \label{FRF--1}
			\hat{r}_{\oplus}^{(1)}(x)=\underset{y \in \Omega}{\argmin}\frac{1}{B} \sum_{b=1}^{B}d^2\left(\underset{y^{\prime} \in \Omega}{\argmin}\frac{1}{N(L_{b}(x; \mathcal{D}_{n}^b,\xi_{b}))} \sum_{i: X_{i} \in L_{b}(x; \mathcal{D}_{n}^b,\xi_{b})}d^2\left(Y_{i},y^{\prime}\right),y\right).
		\end{align}
		Alternatively, we can start from \eqref{RF-2} and rewrite the random forest estimator as
		$$
		\hat{r}(x)=\underset{y \in \mathcal{R}}{\argmin}\sum_{i=1}^{n} \alpha_{i}\left(x\right) \left(Y_{i}-y\right)^2.
		$$
		Then we propose the second generalization for metric space valued responses as
		\begin{align}\label{FRF--2}
			\hat{r}_{\oplus}^{(2)}(x)=\underset{y \in \Omega}{\argmin}\sum_{i=1}^{n} \alpha_{i}\left(x\right) d^2\left(Y_{i},y\right).
		\end{align}
		The first generalization \eqref{FRF--1} is actually the Fr\'echet random forest proposed by \cite{capitaine2019fr}. The idea behind it is to average the results of each Fr\'echet tree. However, our proposed second generalization  \eqref{FRF--2} looks more concise because it involves only one $``\argmin"$, which brings convenience to our theoretical derivation. To acquire the random forest kernel $\alpha_{i}\left(x\right)$, we still need to construct all Fr\'echet trees. It is worth noting that when $\Omega \subseteq \mathcal{R}$, \eqref{FRF--1}  and \eqref{FRF--2} are equivalent.  However, for a general metric space $\Omega$,  \eqref{FRF--1}  and \eqref{FRF--2} may not be the same. Therefore these are two different methods. Building upon the framework of the weighted Fr\'echet mean outlined in \eqref{FRF--2}, we are able to systematically establish the asymptotic theory. These results fill the theoretical gap left in \cite{capitaine2019fr} and encompass the theory for classical Euclidean random forests as a specific case. Additionally, this framework offers the potential for generalizing our method to a local linear version (see Section~\ref{sec4}).}
	
	In the generalization \eqref{FRF--2}, random forests produce the weighting function $\alpha_i(x)$ for Fr\'echet regression but do not participate in the prediction.  Since \eqref{FRF--2} is essentially a local constant estimator based on the random forest kernel, we call it random forest weighted local constant Fr\'echet regression (RFWLCFR). Our proposal is expected to outperform Nadaraya-Watson Fr\'echet regression estimator \citep{hein2009robust} and the local Fr\'echet regression estimator \citep{petersen2019frechet} for the following two reasons. Firstly, the random forest kernel can handle moderately large $p$. Secondly, if the split criterion of Fr\'echet trees depends on the response $Y$,  the random forest kernel will be adaptive in the sense that it incorporates the information of both $X$ and $Y$. Here we use a simulation example to illustrate the adaptiveness of a random forest kernel based on $100$ Fr\'echet trees with the variance reduction splitting criterion introduced before. 
	
	\begin{figure}[b!]
		\centering
		\includegraphics[width=0.5\linewidth]{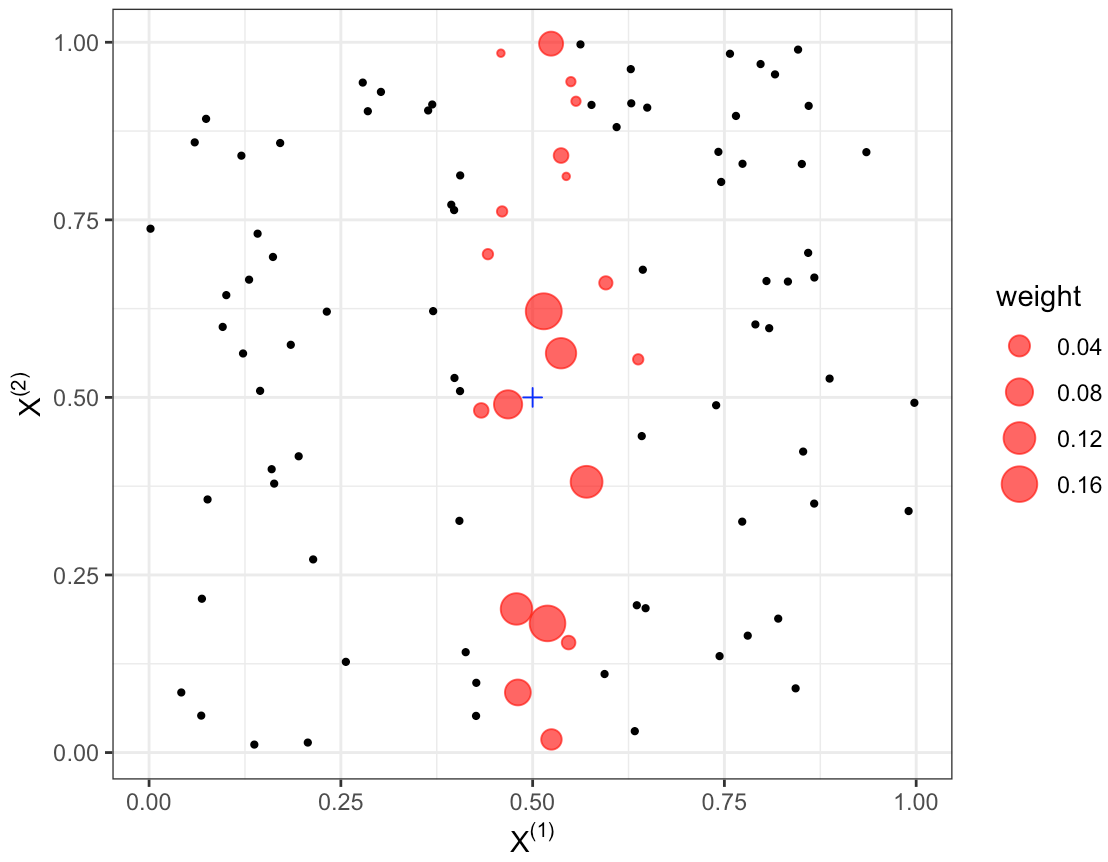}
		\caption{Weights given by the random forest kernel. Each point represents a training sample. The red points represent samples whose weights to $(0.5,0.5)$ are greater than $0$ and the diameter of these points indicates the size of the weights.}
		\label{figure:adaptive}
	\end{figure}
	
	\begin{example}
		Consider a Fr\'echet regression problem for a spherical response $Y$ equipped with the geodesic distance and a predictor $X=(X^{(1)}, X^{(2)}) \sim \mathcal{U}\left([0,1]^2\right)$ as
		\begin{equation}\label{example}
			\begin{aligned}
				Y=\big(&\sin (X^{(1)}+\varepsilon) \sin (X^{(1)}+\varepsilon),\sin (X^{(1)}+\varepsilon) \cos (X^{(1)}+\varepsilon), \cos (X^{(1)}+\varepsilon)\big)^{\T},
			\end{aligned}
		\end{equation}
		where $\epsilon \sim \mathcal{N}\left(0, 0.2^2\right)$. The values of the random forest kernel at $100$ training samples when making a prediction at the center $(0.5, 0.5)$ are displayed in Fig.~\ref{figure:adaptive}.
		It can be observed that the weights decay much more quickly along the $X^{(1)}$ direction and are less influenced by the value of $X^{(2)}$. Under the construction mechanism of the random forest, samples that are close to the target point in the $X^{(1)}$ direction are considered important for prediction, which is consistent with the fact that $Y$ is only relevant to $X^{(1)}$ in \eqref{example}. Unlike the random forest kernel, the Euclidean distance based kernel does not have such an adaptive nature, and the local neighborhoods determined by it for the target point will not spread out in irrelevant directions. 
	\end{example}

	\section{Theoretical Properties} \label{sec3}
	In this section, we first introduce the population objective form of our method RFWLCFR, and then establish its theoretical properties from three perspectives: consistency, convergence rate, and asymptotic distribution.
	\subsection{Population Target}
	Here we consider a random pair $(X, Y) \sim F$, where $X$ and $Y$ take values in $[0,1]^p$ and $\Omega$. To facilitate further theoretical investigations with the infinite order U-statistic and U-process tools, we follow \cite{wager2018estimation} and assume that $B \rightarrow \infty$ given infinite computing power. {Let 
		\begin{align}\label{infinite weight}
			\bar{\alpha}_{i}(x)= \binom{n}{s_n}^{-1} \sum_{k}E_{\xi \sim \Xi} \frac{1\left\{X_{i} \in L(x; \mathcal{D}_{n}^k,\xi)\right\}}{N(L(x; \mathcal{D}_{n}^k,\xi))},
		\end{align}
		where the summation about $k$ is taken over all $\binom{n}{s_n}$ subsamples of size $s_{n}$, $\mathcal{D}_n^k$ is the $k$th subsample of $\mathcal{D}_{n}$, and the expectation is taken about the random effect $\xi$.} $B \rightarrow \infty$ is equivalent to taking into account all Fr\'echet trees constructed by each subsample of $\mathcal{D}_{n}$ and all $\xi$ conditioned on each subsample, which leads to the random forest kernel \eqref{infinite weight}.
	Now we consider the infinite forest version
	\begin{align}\label{RFWLCFR-1}
		\hat{r}_{\oplus}(x)=\underset{y \in \Omega}{\argmin}\hat{R}_{n}(x,y)=\underset{y \in \Omega}{\argmin}\sum_{i=1}^{n}
		\bar{\alpha}_{i}\left(x\right) d^2\left(Y_{i},y\right),
	\end{align}
	and develop the corresponding large sample theory. It can be observed that
	\begin{align}\label{RFWLCFR-2}\nonumber
		\hat{r}_{\oplus}(x)
		=&\underset{y \in \Omega}{\argmin} \sum_{i=1}^{n}\Bigg[ \binom{n}{s_n}^{-1} \sum_{k}E_{\xi \sim \Xi} \frac{1\left\{X_{i} \in L(x; \mathcal{D}_{n}^k,\xi)\right\}}{N(L(x; \mathcal{D}_{n}^k,\xi))}\Bigg]
		d^2\left(Y_{i},y\right)\\
		=& \underset{y \in \Omega}{\argmin}\binom{n}{s_n}^{-1} \sum_{k}E_{\xi \sim \Xi}\Bigg\{\frac{1}{N(L(x;\mathcal{D}_n^k,\xi))} \sum_{i: X_{i} \in L(x;\mathcal{D}_n^k,\xi)}d^2\left(Y_{i},y\right)\Bigg\}. 
	\end{align}
	As $s_n\rightarrow \infty$ when $n\rightarrow \infty$, the objective function $\hat{R}_{n}(x,y)$ of $\hat{r}_{\oplus}(x)$ is an infinite order U-statistic with rank $s_{n}$ for any fixed $y \in \Omega$. The assumption that
	$B$ is large enough for Monte Carlo effects not to matter is inspired by \cite{scornet2015consistency, wager2018estimation, cevid2022distributional} and many other works. In practice, we can choose $B$ as large as possible. In Remark~\ref{remark 3.2}, we provide a discussion regarding the issue of finite $B$.
	
	Based on \eqref{RFWLCFR-1} and \eqref{RFWLCFR-2}, we define two population level versions of $\hat{r}_{\oplus}(x)$ as follows.
	\begin{equation}\label{P-RFWLCFR-2}
		\begin{aligned}
			\tilde{r}_{\oplus}(x)&=\underset{y \in \Omega}{\argmin}\tilde{R}_{n}(x,y)
			=\underset{y \in \Omega}{\argmin} \ n E \big\{\bar{\alpha}_{i}\left(x\right) d^2\left(Y_{i},y\right)\big\}\\
			&=\underset{y \in \Omega}{\argmin}E\Bigg\{\frac{1}{N(L(x;\mathcal{D}_n^k,\xi))} \sum_{i: X_{i} \in L(x;\mathcal{D}_n^k,\xi)}d^2\left(Y_{i},y\right)\Bigg\},
		\end{aligned}
	\end{equation}
	where the expectation is taken about all randomness. We can separate $d(\hat{r}_{\oplus}(x),m_{\oplus}(x))$ into the bias term $d(\tilde{r}_{\oplus}(x), m_{\oplus}(x))$ and the variance term $d(\hat{r}_{\oplus}(x),\tilde{r}_{\oplus}(x))$ for asymptotic analysis.
	
	\subsection{Consistency} \label{sec3.1}
	To study the pointwise consistency of $\hat{r}_{\oplus}(x)$, we assume the following regularity conditions.
	
	(A1) $(\Omega, d)$ is a bounded metric space, \emph{i.e.}, $\operatorname{diam}(\Omega)=\sup _{y_{1}, y_{2} \in \Omega} d\left(y_{1}, y_{2}\right)<\infty$.
	
	(A2) The marginal density $f$ of $X$, as well as the conditional densities $g_{y}$ of $X \mid Y=y$, exist and are bounded and continuous, the latter for all $y \in \Omega$. And $f$ is also bounded away from zero such that $0<f_{\min } \leq f$. Additionally, for any open $V \subseteq \Omega$, $\int_{V} \mathrm{~d} F_{Y \mid X}(x, y)$ is continuous as a function of $x$.
	
	(A3) $\operatorname{diam}\left(L(x)\right) \rightarrow 0$ in probability, where $L(x)$ is the leaf node containing $x$ of any Fr\'echet tree in the random forest.
	
	(A4) The object $m_{\oplus}(x)$ exists and is unique. For all $n$, $\tilde{r}_{\oplus}(x)$ and $\hat{r}_{\oplus}(x)$ exist and are unique, the latter almost surely. Additionally, for any $\varepsilon>0$,
	$$
	\begin{array}{l}
		\inf _{d\left(y, m_{\oplus}(x)\right)>\varepsilon}\left\{M_{\oplus}(x,y)-M_{\oplus}\left(x,m_{\oplus}(x)\right)\right\}>0,  \\
		\liminf _{n} \inf _{d\left(y, \tilde{r}_{\oplus}(x)\right)>\varepsilon}\left\{\tilde{R}_{n}(x,y)-\tilde{R}_{n}\left(x, \tilde{r}_{\oplus}(x)\right)\right\}>0.
	\end{array}
	$$
	
	Assumptions (A1), (A2) and (A4) are commonly used conditions to study the Fr\'echet regression, see \cite{petersen2019frechet}. If the termination condition for the growth of each Fr\'echet tree is that the number of samples in the leaf nodes does not exceed a certain constant, for example, the Fr\'echet tree is $\alpha$-regular, which will be mentioned in Section~\ref{sec3.2}, the assumption (A3) will hold (see Lemma 2 of \cite{wager2018estimation}). Similar conditions can also be found in \cite{denil2013consistency}. The assumption (A4) is also a regular condition to guarantee the consistency of M-estimators (see  Corollary 3.2.3 of \cite{van1996weak}). The simulation in Section~\ref{sec5} will consider three kinds of metric spaces:  probability distributions equipped with the Wasserstein metric, symmetric positive definite matrices equipped with Log-Cholesky metric or the affine-invariant metric, and unit sphere equipped with geodesic distance. The first two can satisfy the assumption (A4) naturally or under very weak conditions.  For the last one, the uniqueness of Fr\'echet means is generally not guaranteed but can be satisfied under certain circumstances, for example restricting the support of the underlying distribution. 
	
	In addition to assumptions (A1)--(A4), we further require that Fr\'echet trees are constructed with honesty and symmetry as defined in \cite{wager2018estimation}.  More instructions about honesty are given in  Appendix~\ref{sec A} and can easily be adapted to Fr\'echet trees.

	(a) (\textit{Honest}) The Fr\'echet tree is honest if the training examples whose responses have been used to decide where to place the splits can not be involved in the calculation of the random forest kernel.
	
	(b) (\textit{Symmetric}) The Fr\'echet tree is symmetric if the output of the tree does not depend on the order $(i = 1, 2, \dots)$ in which the training examples are indexed.
	
	\begin{theorem}\label{consistency}
		Suppose that  for a fix $x \in [0,1]^p$, (A1)--(A4) hold and Fr\'echet trees are honest and symmetric. Then $\hat{r}_{\oplus}(x)$  is pointwise consistent, that is,
		\begin{equation*}
			d(\hat{r}_{\oplus}(x), m_{\oplus}(x))=o_{p}(1).
		\end{equation*}
	\end{theorem}
	\begin{remark}\label{remark 3.2}
		Building infinite Fr\'echet trees is actually computationally unfeasible. This issue also represents a perennial challenge associated with U-statistics, particularly when dealing with large $n$. In fact, the above results also hold true with $B = B_n < \binom{n}{s_n}$ where $B_n \rightarrow \infty$ as $n \rightarrow \infty$. Consider 
		\begin{align*}
			\hat{r}_{\oplus}(x)=&\underset{y \in \Omega}{\argmin}\hat{R}_{n}(x,y)
			=\underset{y \in \Omega}{\argmin} \frac{1}{B_n} \sum_{b=1}^{B_n}\Bigg\{\frac{1}{N(L_{b}(x; \mathcal{D}_{n}^b,\xi_{b}))} \sum_{i: X_{i} \in L_{b}(x; \mathcal{D}_{n}^b,\xi_{b})}d^2\left(Y_{i},y\right)\Bigg\}.
		\end{align*}
		In this case, $\hat{R}_n(x,y)$ is called an incomplete infinite order U-statistic with a random kernel (about $\xi_{b}$) for each fixed $y \in \Omega$, which in general does not fit within the framework of infinite order U-statistics. Since the randomization parameter $\xi \sim \Xi$ here is  independent of the training sample $\mathcal{D}_{n}$, the consistency of $\hat{R}_n(x, y)-\tilde{R}_n(x, y)=o_p(1)$ for each $y \in \Omega$ still holds. Then by the same proof as Theorem~\ref{consistency}, $\hat{r}_{\oplus}(x)$ is still pointwise consistent with finite Fr\'echet trees. But the other tool of incomplete infinite order U-processes with a random kernel is not clear so far. So the convergence rate and asymptotic normality in the following content will still be developed under the constraint $B \rightarrow \infty$ due to the vast challenges of theoretical techniques.
	\end{remark}
	
	If the previous assumptions are suitably strengthened (see the assumption (U1--U4) before the proof of the following theorem in the Appendix~\ref{sec D}), we can further obtain the uniform convergence results for $\hat{r}_{\oplus}(x)$. Let $\Vert \cdot \Vert$ be the Euclidean norm on $\mathcal{R}^{p}$ and $J > 0$.
	\begin{theorem}\label{uniform consistency}
		Suppose that  (A1), (U1)--(U4) hold and Fr\'echet trees are honest and symmetric. Then
		\begin{equation*}
			\sup_{\Vert x \Vert \leq J}	d(\hat{r}_{\oplus}(x), m_{\oplus}(x))=o_{p}(1).
		\end{equation*}
	\end{theorem}
	
	\subsection{Rate of Convergence} \label{sec3.2}
	We proceed to analyze the convergence rates of the bias term and the variance term separately. The convergence rate of the bias term is closely related to the construction of Fr\'echet trees. Here we follow \cite{wager2018estimation} to place the following additional requirements on the construction of Fr\'echet trees.
	
	(c) (\textit{Random-split}) At each node split, marginalizing over $\xi$, the probability that $X^{(j)} (1\le j\le p)$ is selected as the split variable is bounded below by $\pi / p$ for some $0<\pi \leq 1$. 
	
	(d) (\textit{$\alpha$-regular}) After each splitting, each child node contains at least a fraction $\alpha>0$ of the available training examples which will be used to calculate the random forest kernel. Moreover, the tree stops growing if every leaf node contains only between $k$ and $2k- 1$ observations, where $k$ is some fixed integer. 
	
	To derive the convergence rates, some additional assumptions are required.  We start with some notations.
	Let  $Z_i=(X_i,Y_i)$ and $\mathcal{D}_n^k=\left(Z_{i_{k,1}}, Z_{i_{k,2}}, \ldots, Z_{i_{k,s_n}}\right)$, and define
	\begin{align*}
		&H_n(Z_{i_{k,1}}, \ldots, Z_{i_{k,s_n}}, y)=
		E_{\xi \sim \Xi}\Bigg[\frac{1}{N(L(x;\mathcal{D}_n^k,\xi))} \sum_{i: X_{i} \in L(x;\mathcal{D}_n^k,\xi)}\left\{d^2\left(Y_{i},y\right)-d^2\left(Y_{i},\tilde{r}_{\oplus}(x)\right)\right\}\Bigg].
	\end{align*}
	Consider the function class
	$$
	\mathcal{H}_{\delta}:=\left\{H_n(z_{1}, \ldots, z_{s_n}, y): d\left(y, \tilde{r}_{\oplus}(x)\right)<\delta\right\}.
	$$
	Let $Z_{i}^{0}=(X_{i},Y_{i})$, for $i=1, \ldots, n$, and let $\left\{Z_{i}^{1}\right\}_{i=1}^{n}$ be i.i.d.,  independent of $\left\{Z_{i}^{0}\right\}_{i=1}^{n}$ with the same distribution. For $\forall H_n(y_1), H_n(y_2) \in  \mathcal{H}_{\delta}$, define the following random pseudometric
	\begin{align*}
		&d_{j}\left(H_n(y_1), H_n(y_2)\right)=\frac{\sum_{k=1}^{n}\left|\sum_{a \in(n)_{s_n}: a_{1}=k} {H}_{n}\left(Z_{a_{j}}^{0,1} ; y_1\right)-{H}_{n}\left(Z_{a_{j}}^{0,1} ; y_2\right)\right|}{\sum_{a \in(n)_{s_n}} {G}_{\delta}\left(Z_{a_{j}}^{0,1}\right)},
	\end{align*}
	where $Z_{a_{j}}^{0,1}=\left(Z_{a_{1}}^{0}, \ldots, Z_{a_{j}}^{0}, Z_{a_{j+1}}^{1}, \ldots, Z_{a_{s_n}}^{1}\right)$, $(n)_{s_n}$ represents all the permutations of taking $s_n$ distinct elements from the set $\left\{1,2,\dots,n\right\}$, and ${G}_{\delta}$ is an envelope function  for $\mathcal{H}_{\delta}$ such that $|H_n| \leq {G}_{\delta}$ for every $H_n \in \mathcal{H}_{\delta}$.
	
	With the above preparation, the assumptions are specified as follows.
	
	(A5) For each $y$, $M_{\oplus}(x,y)$ is Lipschitz-continuous about $x$, and the Lipschitz constant has a common upper bound $K$.
	
	(A6) There exist $\delta_1>0, C_1>0$ and $\beta_1>1$, possibly depending on $x$, such that, whenever $d\left(y, m_{\oplus}(x)\right)<\delta_1$, we have $M_{\oplus}(x,y)-M_{\oplus}\left(x,m_{\oplus}(x)\right) \geq C_1 d\left(y, m_{\oplus}(x)\right)^{\beta_1}$.
	
	(A7) There exist constants $A$ and $V$ such that
	$$\max _{j \leq s_n} N\left(\varepsilon, d_{j}, \mathcal{H_{\delta}}\right) \leq A \varepsilon^{-V}$$
	as $\delta \rightarrow 0$ for any $\epsilon \in (0,1]$, where  $N\left(\varepsilon, d_{j}, \mathcal{H_{\delta}}\right)$ is the $\varepsilon$-covering number of the function class $\mathcal{H}_{\delta}$ based on the pseudometric $d_j$ we introduce.
	
	(A8) There exist $\delta_2>0, C_2>0$ and $\beta_2>1$, possibly depending on $x$, such that, whenever $d\left(y, \tilde{r}_{\oplus}(x)\right)<\delta_2$, we have
	$$
	\liminf _{n}\left\{\tilde{R}_{n}(x,y)-\tilde{R}_{n}\left(x,\tilde{r}_{\oplus}(x)\right)-C_{2} d\left(y, \tilde{r}_{\oplus}(x)\right)^{\beta_{2}}\right\} \geq 0.
	$$
	
	The Lipschitz continuity in the assumption (A5) allows us to control the bias term by restricting the diameter of the sample space represented by the leaf node. The assumption (A7) along with the pseudometric $d_j$ were proposed by \cite{heilig1997empirical} and \cite{heilig2001limit}  to establish the maximal inequality of infinite order U-processes. From the perspective of empirical process, (A7) regulates $\mathcal{H_{\delta}}$ a Euclidean class. Knowing that a class of functions is Euclidean aids immensely in establishing the convergence rate of the variance term.  The assumptions (A6) and (A8) comes from \cite{petersen2019frechet}. (A8) is also an extension of the condition that controls the convergence rate of $M$-estimators. Please refer to Theorem 3.2.5 of \cite{van1996weak} for more details. {First, we establish the rate for the bias term as follows.
		\begin{lemma}\label{bias rate}
			Suppose that for a fixed $x \in [0,1]^p$, (A1), (A2), (A4), (A5) and (A6) hold,  and the Fr\'echet trees are $\alpha$-regular, random-split and honest. Then, as $n \rightarrow \infty$, provided that $\alpha \leq 0.2$, we have
			$$
			d(\tilde{r}_{\oplus} (x),m_{\oplus}(x))=O\left(s_n^{-\frac{1}{2} \frac{\log \left(1-\alpha\right)}{\log (\alpha)} \frac{\pi}{p} \frac{1}{\beta_1-1}}\right).
			$$
		\end{lemma}
		\begin{remark}
			Consider the special case when $\Omega \subseteq \mathcal{R}$. Then
			\begin{align*}
				M_{\oplus}(x,y)-M_{\oplus}\left(x,m_{\oplus}(x)\right)
				=\left\{y-m_{\oplus}(x)\right\}^2.
			\end{align*}
			That is to say, the assumption (A6) holds when $\beta_1=2$. By the above lemma \ref{bias rate},
			$$d(\tilde{r}_{\oplus} (x),m_{\oplus}(x))=O\left(s_n^{-\frac{1}{2} \frac{\log \left(1-\alpha\right)}{\log (\alpha)} \frac{\pi}{p} }\right) $$
			as $n \rightarrow \infty$. This rate coincides with Theorem 3 of \cite{wager2018estimation}, which indicates that our asymptotic bias result generalizes that of Euclidean random forests to random forests with metric space valued responses.
	\end{remark}}
	
	{Then we turn to the convergence rate of the variance term.
		\begin{lemma}\label{variance rate}
			Suppose that for a fixed $x \in [0,1]^p$, (A1), (A4), (A7) and (A8) hold, and the Fr\'echet trees are symmetric. Then we have
			$$
			d(\hat{r}_{\oplus}(x),\tilde{r}_{\oplus}(x))=O_p\left(\left(\frac{s_n^2 \log s_n}{n}\right)^{\frac{1}{2 \left(\beta_2-1\right)}}\right).
			$$
		\end{lemma}
		\begin{remark}
			When $\Omega \subseteq \mathcal{R}$, and if the  trees are honest, then
			$$ \tilde{R}_n(x,y)=E\left[E\left\{\left(Y-y\right)^2 \mid X \in L(x)\right\}\right] \quad \text{and} \quad \tilde{r}_{\oplus}(x)=E\left[E\left\{Y|X \in L(x)\right\}\right].$$
			And we can further get 
			\begin{align*}
				&\tilde{R}_n(x,y)-\tilde{R}_n\left(x,\tilde{r}_{\oplus}(x)\right)\\
				=&E\left[E\left\{\left(Y-y\right)^2 \mid X \in L(x)\right\}\right] -E\left[E\left\{\left(Y-\tilde{r}_\oplus(x)\right)^2 \mid X \in L(x)\right\}\right] \\
				=&y^2-2yE\left[E\left\{Y \mid X \in L(x)\right\}\right]+\left(E\left[E\left\{Y \mid X \in L(x)\right\}\right]\right)^2\\
				= & \left\{y-\tilde{r}_{\oplus}(x)\right\}^2,
			\end{align*}
			which indicates that $\beta_2=2$ in the assumption (A8) for Euclidean responses. By the result of Lemma \ref{variance rate},  $d(\hat{r}_{\oplus}(x),\tilde{r}_{\oplus}(x))=O_{p}\big((s_n^2 \log s_n/n)^{1/2}\big)$. The rate derived here is slower than $(s_n/n)^{1/2}$ as described in our asymptotic normality result (Remark~\ref{remark 2} in Appendix~\ref{sec E}). The reason is that the existing maximal inequalities of infinite order U-processes are not strong enough. More refined tools of infinite order U-processes are expected to be developed to further improve this convergence rate.
	\end{remark}}
	
	Combining Lemma~\ref{bias rate} and Lemma~\ref{variance rate}, we get the convergence rate for $\hat{r}_{\oplus} (x)$.
	\begin{theorem}\label{rate}
		Suppose that for a fixed $x \in [0,1]^p$, (A1), (A2), (A4)--(A8)  hold,  and Fr\'echet trees  are $\alpha$-regular, random-split, honest and  symmetric. If $\alpha \leq 0.2$, then
		\begin{align*}
			&d(\hat{r}_{\oplus} (x),m_{\oplus}(x))=O_p\left(s_n^{-\frac{1}{2} \frac{\log \left(1-\alpha\right)}{\log (\alpha)} \frac{\pi}{p} \frac{1}{\beta_1-1}}+\left(\frac{s_n^2 \log s_n}{n}\right)^{\frac{1}{2 \left(\beta_2-1\right)}}\right).
		\end{align*}
	\end{theorem}

	\subsection{Asymptotic Normality} \label{sec3.3}
	There are two major challenges in deriving the asymptotic normality of $\hat{r}_{\oplus}(x)$. On the one hand, $\hat{r}_{\oplus}(x)$ has no explicit expression like Euclidean random forests and it is not the classical $M$-estimator, but  $M_{m}$-estimator \citep{bose2018u} and even $M_{m_n}$-estimator with infinite order U-processes. So we have to deal with the most general $M_{m_n}$-estimator,  where $m_n$ diverges to infinity. On the other hand,  the $M_{m_n}$-estimator here takes value not in Euclidean space but in general metric space, which will also bring difficulties to the study of asymptotic limiting distribution. To address the first difficulty, we will generalize the result in section $2.5$ of  \cite{bose2018u} to acquire the probability representation and asymptotic normality of the $M_{m_n}$-estimator. As for the second issue, the seminal work of \cite{bhattacharya2017omnibus} and \cite{bhattacharya2003large,bhattacharya2005large} concluded that the map of the sample Fr\'echet mean is asymptotically normally distributed around the map of the Fr\'echet mean under certain assumptions. We follow their developments in combination with our developed asymptotic tool for $M_{m_n}$-estimator to establish the asymptotic normality of $\hat{r}_{\oplus}(x)$ finally. {While these difficulties have been reasonably addressed, the conditions under which asymptotic normality holds are technical and not easily verifiable in practice. Consequently, the asymptotic normality result is primarily intended for theoretical completeness and may not be convenient for practical statistical inference. Therefore we have placed this section in the Appendix~\ref{sec E} for interested readers. It is noteworthy that our result encompasses the asymptotic normality of the Euclidean random forest as a special case. In addition, the theory developed for $M_{m_n}$-estimator based on infinite order U-processes and U-Statistics is of independent interest. Overall, the development of a more practically significant asymptotic distribution theory remains a challenging endeavor for our problem. When $Y$ comes from specialized metric spaces such as Riemannian manifolds or Hilbert spaces, the incorporation of additional space properties may lead to further breakthroughs. However, it is beyond the scope of this paper.}
	
	\section{Local Linear Smoothing}\label{sec4}
	In Section~\ref{sec2}, we have proposed RFWLCFR, which is a Nadaraya-Watson type estimator using the random forest kernel. A very natural extension is to carry out local linear Fr\'echet regression further. The local linear estimator is more flexible and accurate in capturing smooth signals. Local Fr\'echet regression proposed by \cite{petersen2019frechet} is a novel local linear estimator adapted to cases with metric space valued responses. Similar to the classical local linear estimator, their method still suffers from the curse of dimensionality. This section proposes the second method called random forest weighted local linear Fr\'echet regression (RFWLLFR), which adopts the random forest kernel to local linear Fr\'echet regression.
	
	\cite{bloniarz2016supervised} and \cite{friedberg2020local} considered a local linear regression with Euclidean responses based on the random forest kernel
	\begin{equation}\label{Euclidean local linear}
		\left(\hat{\beta}_{0}, \hat{\beta}_{1}\right)=\underset{\beta_{0}, \beta_{1}}{\operatorname{argmin}} \frac{1}{n} \sum_{i=1}^{n} {\alpha}_{i}(x)\left\{Y_{i}-\beta_{0}-\beta_{1}^{\T}\left(X_{i}-x\right)\right\}^{2},
	\end{equation}
	where $\alpha_{i}\left(x\right)=\frac{1}{B} \sum_{b=1}^{B} \frac{1\left\{X_{i} \in L_{b}(x; \mathcal{D}_{n}^b,\xi_{b})\right\}}{N(L_{b}(x;\mathcal{D}_{n}^b,\xi_{b}))}$. Then define the random forest weighted local linear estimator as
	\begin{equation}\label{RF based local}
		\hat{l}(x)=\hat{\beta}_{0}=e_{1}^{\T} (\tilde{X}^{\T}A\tilde{X})^{-1}\sum_{i=1}^{n}\left(\begin{array}{c}
			1 \\
			X_{i}-x
		\end{array}\right) {\alpha}_{i}(x) Y_{i}
	\end{equation}
	where
	\begin{gather*}
		\tilde{X}:=\left(\begin{array}{cccc}1 & (X_{1}-x)^{\T} \\ 1 & (X_{2}-x)^{\T}  \\ \vdots & \vdots \\ 1 & (X_{n}-x)^{\T} \end{array}\right),
		A:=\operatorname{diag}\left({\alpha}_{1}(x), \ldots,{\alpha}_{n}(x)\right),
		e_{1}:=\left(1, 0, \cdots, 0\right)^{\T}.
	\end{gather*}
	To generalize \eqref{RF based local} with metric space valued responses, we rewrite it as an implicit form
	$$
	\hat{l}(x)=\underset{y \in \mathcal{R}}{\argmin}  \ e_{1}^{\T} (\tilde{X}^{\T}A\tilde{X})^{-1}\sum_{i=1}^{n}\left(\begin{array}{c}
		1 \\
		X_{i}-x
	\end{array}\right) \alpha_{i}(x) (Y_{i}-y)^2.
	$$
	Replacing the Euclidean distance  with a metric $d$,  the RFWLLFR for a general metric space $(\Omega,d)$  is proposed as
	\begin{equation} \label{RFWLLFR}
		\hat{l}_{\oplus}(x)=\underset{y \in \Omega}{\argmin} \ e_{1}^{\T} (\tilde{X}^{\T}A\tilde{X})^{-1}\sum_{i=1}^{n}\left(\begin{array}{c}
			1 \\
			X_{i}-x
		\end{array}\right) {\alpha}_{i}(x) d^2(Y_{i},y).
	\end{equation}
	
	Our proposed local constant and local linear methods are both to calculate weighted Fr\'echet means, but the weights of the local linear method may be negative. Due to the complexity of RFWLLFR, we only study the consistency of $\hat{l}_{\oplus}(x)$ here. To this end, we assume the following conditions.
	
	(A9) $X \sim \mathcal{U}\left([0,1]^{p}\right)$, the uniform distribution on $[0,1]^p$.
	
	(A10) $N(L_{b}(x; \mathcal{D}_{n}^b,\xi_{b})) \rightarrow \infty$ for $b=1,\dots,B$.
	
	(A11) The Fr\'echet trees are trained in such a way that for each $y \in \Omega$
	$$
	\max _{1 \leq i \leq n, 1 \leq b \leq B}\left[1\big\{X_{i} \in L_{b}(x; \mathcal{D}_{n}^b,\xi_{b})\big\} \left|M_{\oplus}(X_i,y)-M_{\oplus}(x,y)\right|\right] \stackrel{p}{\rightarrow} 0.
	$$
	That is, the leaf node containing $x$ shrinks such that the maximal variation of the function $M_{\oplus}(\cdot,y)$ within a cell shrinks to 0 in probability for each $y \in \Omega$.
	
	(A12) $m_{\oplus}(x)$ and $\hat{l}_{\oplus}(x)$ exist and are unique, the latter almost surely. For any $\varepsilon>0$,
	\begin{align*}
		\inf _{d\left(y, m_{\oplus}(x)\right)>\varepsilon}\left\{M_{\oplus}(x,y)-M_{\oplus}\left(x,m_{\oplus}(x)\right)\right\}>0.
	\end{align*}
	
	Assumptions (A9)-(A11) are similar to the conditions used in \cite{bloniarz2016supervised} to establish the consistency of a nonparametric regression estimator using random forests as adaptive nearest neighbor generators. The assumption (A11) is a general condition,  which can be deduced from assumptions (A2) and (A3). It is important to note that assumption (A10) is not required for the consistency of our local constant method.  But it provides a guarantee that the law of large numbers can be used for the samples in the leaf node, which derives the consistency of $\hat{l}_{\oplus}(x)$ even when $B$ in the random forest kernel is a fixed constant. 
	Another reasonable interpretation is that the random forest provides weights for the final local linear regression and should not be used to model strong, smooth signals to prevent overfitting phenomena. In other words, the Fr\'echet trees that form the forest here should not grow too deep. These can also be observed in Figure~\ref{deep} from Appendix~\ref{sec B.2}, where moderately grown trees can improve the performance of $\hat{l}_{\oplus}(x)$. But our local constant method often requires deeper trees. In addition to the assumptions above, the honesty condition is still necessary to prove the consistency of $\hat{l}_{\oplus}(x)$.
	\begin{theorem} \label{consistency of RFWLLFR}
		Suppose that for a fixed $x \in [0,1]^{p}$,  (A1), (A9)--(A12) hold and Fr\'echet trees are honest. Then $\hat{l}_{\oplus}(x)$ is pointwise consistent, that is,
		\begin{equation*}
			d(\hat{l}_{\oplus}(x), m_{\oplus}(x))=o_{p}(1).
		\end{equation*}
	\end{theorem}
	
	\begin{remark}
		As can be seen from Remark \ref{remark 3.2}, we have weakened the requirement of our local constant method for the number $B$ of trees, which eliminates the gap between theoretical investigations and practical applications. Certainly, if we reimpose assumptions of the local linear method here on it, a parallel proof guarantees that $B$ can be further weakened to a fixed constant.
	\end{remark}
	
\begin{figure}[t!]
		\centering
		\includegraphics[width=0.8\linewidth]{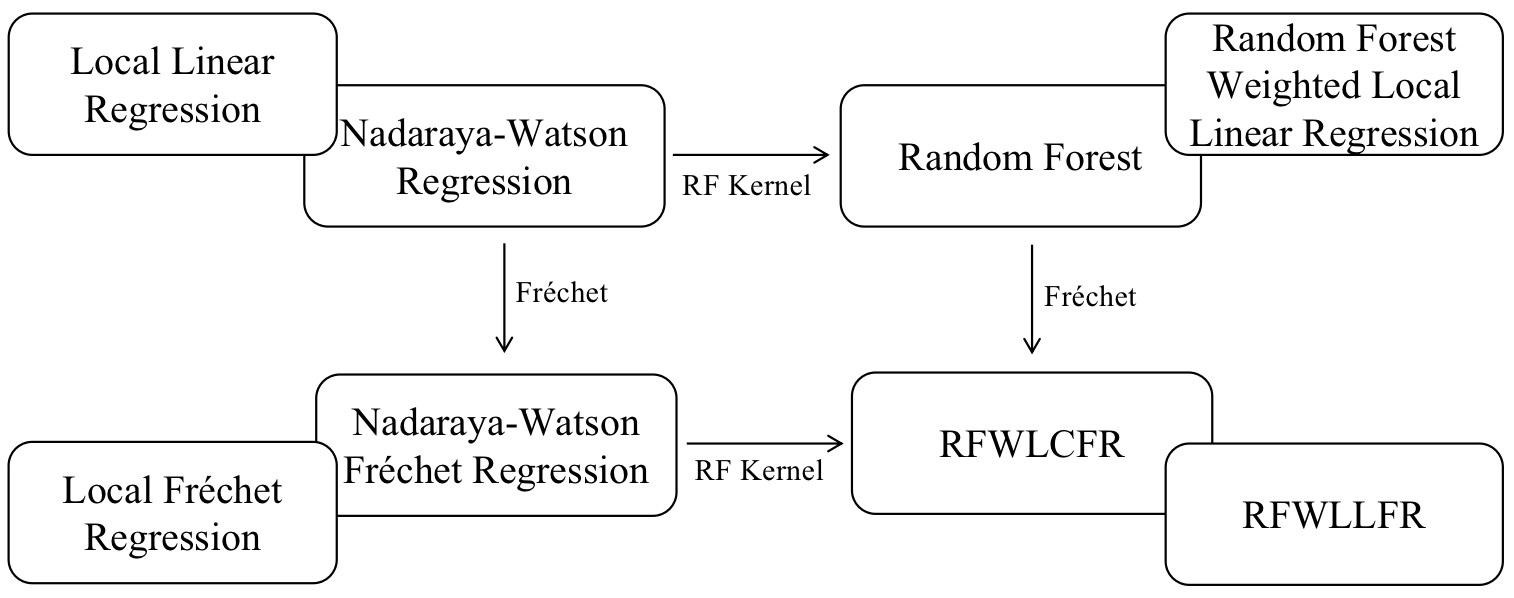}
		\caption{The relationships among the eight local estimators.}
		\label{pic}
	\end{figure}

	After introducing our two methods, we briefly summarize all relevant methods. The classical Nadaraya-Watson regression,  random forest,  Nadaraya-Watson Fr\'echet regression \citep{hein2009robust} and RFWLCFR are essentially all local constant estimators. The classical local linear regression, random forest weighted local linear regression \citep{bloniarz2016supervised,friedberg2020local}, local Fr\'echet regression \citep{petersen2019frechet} and RFWLLFR are essentially all local linear estimators. The relationship among the eight types of estimators is shown in Fig.~\ref{pic}.
	
	\section{Forest Kernel-based Permutation Variable Importance} \label{add sec}
	{As a byproduct of the classical random forest algorithm, one can effectively assess the importance of each variable using the out-of-bag (oob) samples \citep{breiman2001random}, which refer to those observations not used in constructing the tree. Specifically, when the $b$th tree is grown,  the oob samples are passed down the tree, and the prediction accuracy is recorded. Then, the $j$th variable values are randomly permuted within the oob samples, and accuracy is re-evaluated. The average decrease in accuracy, resulting from this permutation, serves as a measure of the importance of variable $j$ in the random forest.  The above process can be readily extended to non-Euclidean scenarios by replacing the Euclidean distances with general distances. A notable disadvantage of this type of permutation-based measure is that it is formulated at the tree level, and averaged over the forest. Hence this method does not correctly reflect the reduced accuracy on the forest kernel due to the permutation. Here, we introduce a novel method that uses the forest kernel for oob permutation variable importance.}
	
	{For each sample $(X_i, Y_i)$ from the entire training data $\mathcal{D}_{n}$,  we can collect the Fr\'echet trees whose construction $(X_i, Y_i)$ did not participate in. Consequently, $(X_i, Y_i)$ serves as a natural test point for a small forest composed of these trees.  We predict the response of $X_i$ by RFWLCFR based on the kernel generated by this small forest and record the prediction error. Note that the oob prediction mechanic here is also similar to the ones used in the jackknife confidence interval \citep{wager2014confidence}. These errors are then averaged over all training samples as a baseline. To quantify the importance of the $j$th variable, we randomly permute the value for the variable $j$ within $\mathcal{D}_{n}$. The prediction error is recalculated by treating each permuted observation as a testing point. The decrease in accuracy on the permuted data compared to the baseline is a variable importance measure for variable $j$. The complete algorithm is summarized in Algorithm~\ref{importance}. Although Algorithm~\ref{importance} incurs more computational overhead compared to the classical tree-level measure, as it involves identifying unrelated trees and making predictions for each observation,  the extra computational cost is not significant. The advantage of our method lies in its higher precision, given that the prediction process is executed based on a forest-level kernel. In addition, repeating the algorithm multiple times and averaging the variable importance can enhance result stability. In cases where there is an excessive number of features, the variable importance ranking obtained from Algorithm~\ref{importance} can guide us in discarding less important variables, thereby mitigating the challenges arising from the curse of dimensionality.}
	
	\begin{algorithm}[H]
		\caption{: Variable importance calculation}
		\label{importance}
		\begin{algorithmic}
			\STATE \textbf{Inputs:} A training set $\mathcal{D}_{n}=\{(X_i, Y_i)\}_{i=1}^{n}$, number of Fr\'echet trees $B$.
			\vspace{0.2cm}
			\STATE  \textbf{Step 1.} Construct a random forest consisting of $B$ Fr\'echet trees $\{T_{b}(x;\mathcal{D}_{n}^b,\xi_{b})\}_{b=1}^B$ based on $\mathcal{D}_{n}$, which generate the random forest kernel for the achievement of RFWLCFR.
			\STATE  \textbf{Step 2.}
			\FOR {$i=1$ to $n$} 
			\STATE Identify the collection $\mathcal{T}_i$ of Fr\'echet trees whose growth $(X_i, Y_i)$ did not participate in: $\mathcal{T}_i=\{T_{b}(x;\mathcal{D}_{n}^b,\xi_{b}): 1 \leq b \leq B, (X_i, Y_i) \notin \mathcal{D}_{n}^b\}$.
			\STATE Predict the response of $X_i$ with RFWLCFR, denoted by $\hat{r}_{\oplus}^{oob}(X_i)$, based on the random forest kernel provided by $\mathcal{T}_i$.
			\ENDFOR
			\STATE Record the mean square error:
			$
			R_{0}=  \frac{1}{n} \sum_{i=1}^{n} d^2(\hat{r}_{\oplus}^{oob} (X_i),Y_i).
			$
			\STATE  \textbf{Step 3.} 
			\FOR {$j=1$ to $p$} 
			\STATE  Permute the values for the $j$th variable randomly in $\{X_i\}_{i=1}^n$ and repeat Step 2 with the permuted data and the same $\mathcal{T}_i, 1 \leq i \leq n$, acquired in Step 2; Record the corresponding mean square error $R_{j}$.
			\ENDFOR
			\STATE  \textbf{Step 4.} Calculate the variable importance for the $j$th variable: $\operatorname{VI}(X^{(j)})=R_{j}-R_{0}, 1 \leq j \leq p$.
		\end{algorithmic}
	\end{algorithm}

	\section{Simulations} \label{sec5}
	In this section, we consider three Fr\'echet regression scenarios including probability distributions, symmetric positive definite matrices, and spherical data to evaluate the performance of the two methods proposed in this paper. We include the global Fr\'echet regression (GFR) and local Fr\'echet regression (LFR)  \citep{petersen2019frechet},  and the Fr\'echet random forest (FRF) \citep{capitaine2019fr} for comparisons. {Additionally, the single index Fr\'echet regression (IFR) \citep{bhattacharjee2023single} is also considered for a setting of the single index model with symmetric positive-definite matrix responses.} Throughout this section, GFR and LFR can be implemented by R-package ``frechet'' \citep{frechet2020}. FRF can be implemented by the R-package ``FrechForest'' \citep{FrechForest2021} with a slight modification by adding the three new types of responses and their corresponding metrics into the package. {And Julia code for the implementation of IFR can be found in the GitHub platform.}  Our RFWLCFR and RFWLLFR are also implemented in R. For simplicity, the Fr\'echet trees in our simulations are not necessarily honest. All random forests are constructed by $100$ Fr\'echet trees. There are three hyperparameters for each Fr\'echet tree: the size $s_n$ of each subsample, the depth of Fr\'echet trees and the number of features randomly selected at each internal node. The choice of $s_n$ is very tedious and time-consuming. Here we instead acquire all subsamples by sampling from the training data set $\mathcal{D}_{n}$ with replacement, which is commonly used in random forest codes.  When the size $n$ of $\mathcal{D}_{n}$ is large enough, each subsample is expected to have the fraction $(1 - 1/e) \approx 63.2\%$ of the unique examples of $\mathcal{D}_{n}$. We consider $3 \sim \lceil \log_2 n \rceil$ for the range of tuning about the depth of Fr\'echet trees, where $n$ is the number of training samples. For a fair comparison, each method chooses the hyperparameters by cross-validation. 
	
	In the following simulations, each setting is repeated $100$ times. For the $r$th Monte Carlo test,
	$\hat{m}_{\oplus}^r$ denotes the fitted function based on the method $\hat{m}_{\oplus}$ and the quality of the estimation is measured quantitatively by the mean squared error
	$$
	\operatorname{MSE}_{r}(\hat{m}_{\oplus})=  \frac{1}{100} \sum_{i=1}^{N} d^2(\hat{m}_{\oplus}^r(X_i),m_{\oplus}(X_i))
	$$
	based on $100$ new testing points.
	
	\subsection{Fr\'echet Regression for Distributions}\label{sec 5.1}
	Let $(\Omega,d)$ be the metric space of probability distributions on $\mathcal{R}$ with finite second order moments and the quadratic Wasserstein metric $d_W$. For two such distributions $Y_{1}$ and $Y_{2}$, the squared Wasserstein distance is defined by
	\begin{align} \label{Wasserstein distance}
		d_W^{2}\left(Y_{1}, Y_{2}\right)=\int_{0}^{1}\left\{Y_{1}^{-1}(t)-Y_{2}^{-1}(t)\right\}^{2} d t,
	\end{align}
	where $Y_{1}^{-1}$ and $Y_{2}^{-1}$ are quantile functions corresponding to $Y_{1}$ and $Y_{2}$, respectively.
	
	Let $X_1,\ldots,X_n \sim \mathcal{U}\left([0, 1]^p\right)$, and we generate random normal distribution $Y$ by
	\begin{align*}
		Y =\mathcal{N}\left(\mu_Y, \sigma_Y^2\right),
	\end{align*}
	where $\mu_Y$ and $\sigma_Y$ are random variables dependent on $X$ as described in the following.
	
	Setting I-1:
	\begin{align*}
		\mu_{Y} \sim \mathcal{N}\left(5\beta^{\T} X-2.5, \sigma^2\right), \quad \sigma_{Y}=1.
	\end{align*}
	We consider four situations of the dimension of $X$ : $p=2, 5, 10, 20$.\\
	(i) $p=2$:	$\beta=\left(0.75, 0.25\right)$;\\
	(ii) $p=5,10$:	$\beta=\left(0.1, 0.2, 0.3, 0.4,  0, \dots, 0\right)$;\\
	(iii) $p=20$: $\beta=\left(0.1, 0.2, 0.3, 0.4,  0,\dots, 0, 0.1, 0.2, 0.3, 0.4\right)/2$.
	
	Setting I-2:
	\begin{gather*}
		\mu_{Y} \sim \mathcal{N}\left(\sin\left(4 \pi \beta_1^{\T} X\right) \left(2\beta_2^{\T} X-1\right), \sigma^2\right), \quad
		\sigma_{Y}=2 \vert e_1^{\T}X-e_2^{\T}X\vert,
	\end{gather*}
	where $e_i$ is a vector of zeros with $1$ in the $i$th element. We also consider four situations.\\
	(i) For $p=2$: $\beta_1=(0.75,0.25),  \beta_2=(0.25,0.75)$.\\
	(ii) For $p=5,10,20$: $\beta_1=\left(0.1, 0.2, 0.3, 0.4,  0, \dots, 0\right),  \beta_2=\left(0, \dots, 0, 0.1, 0.2, 0.3, 0.4\right)$.
	
	We set $n=100,200$ for $p=2$; $n=200,500$ for $p=5$; $n=500,1000$ for $p=10$ and $n=1000,2000$ for $p=20$. For computation simplicity, the quantile function of each $Y_i$ is discretized as the $21$ quantile points corresponding to the equispaced grids on $[0, 1]$. It can then be verified from \eqref{Wasserstein distance} that the Wasserstein distance between the two distributions is actually the Euclidean distance between their quantile points. Therefore, our RFWLCFR and the method FRF will have the same output.
	
	\begin{table}[b!]
		\centering
		\resizebox{0.83\columnwidth}{!}{
			\begin{tabular}{ccccccccccccccc}
				\toprule
				Model & $(p,n)$  & GFR & LFR & RFWLCFR/FRF & RFWLLFR\\
				\midrule
				\multirow{8}{*}{I-1}
				&$(2, 100)$ & \textbf{0.0014}  (0.0012) & 0.0688  (0.0707) & 0.0269 (0.0071) & 0.0031   (0.0015)  \\
				&$(2, 200)$ & \textbf{0.0006}  (0.0006) & 0.0452  (0.0363) & 0.0179 (0.0037) & 0.0014   (0.0008)  \\
				&$(5, 200)$ & \textbf{0.0011} (0.0006) & NA & 0.0468  (0.0085) & 0.0028  (0.0010)  \\
				&$(5, 500)$ & \textbf{0.0005}  (0.0003) & NA & 0.0299 (0.0049) & 0.0011   (0.0004)  \\
				&$(10, 500)$ & \textbf{0.0009}  (0.0004) & NA &0.0401  (0.0059) &0.0019  (0.0005)  \\
				&$(10, 1000)$ & \textbf{0.0004}  (0.0002) & NA &0.0284  (0.0039) & 0.0009  (0.0002)  \\
				&$(20, 1000)$ & \textbf{0.0009} (0.0003) & NA & 0.0542  (0.0088) &0.0015  (0.0004)  \\
				&$(20, 2000)$ & \textbf{0.0004}  (0.0001) & NA & 0.0444  (0.0059) & 0.0007   (0.0002)  \\
				\midrule
				\multirow{8}{*}{I-2}
				&$(2, 100)$ & 0.3045  (0.0306) & 0.0944  (0.0883) &0.0410  (0.0090) & \textbf{0.0317}  (0.0141)  \\
				&$(2, 200)$ & 0.3023  (0.0278) & 0.0745  (0.1550) & 0.0254  (0.0050) & \textbf{0.0186} (0.0073)  \\
				&$(5, 200)$ & 0.2482 (0.0206) & NA & 0.0772 (0.0136) & \textbf{0.0719} (0.0121)  \\
				&$(5, 500)$ & 0.2335 (0.0241) & NA &0.0557 (0.0087) & \textbf{0.0502} (0.0083)  \\
				&$(10, 500)$ &0.2416 (0.0261) & NA &\textbf{0.1053} (0.0195) & 0.1134 (0.0171)  \\
				&$(10, 1000)$ & 0.2438  (0.0297) & NA & \textbf{0.0901} (0.0174) & 0.0927 (0.0148)  \\
				&$(20, 1000)$ & 0.2442 (0.0245) & NA & \textbf{0.1399} (0.0260) & 0.1554 (0.0197)  \\
				&$(20, 2000)$ & 0.2456 (0.0286) & NA &\textbf{0.1257} (0.0238) & 0.1337 (0.0194)  \\
				\bottomrule
			\end{tabular}
		}
		\caption{Average MSE (standard deviation) of different methods for setting I-1,2 with $\sigma=0.2$ for $100$ simulation runs. Bold-faced numbers indicate the best performers.}
		\label{table:dis_result}
	\end{table}
	
	For the fairness of comparison, we begin with the setting I-1 where data are generated in a linear way. It is compliant with the initial assumptions of GFR. The results are recorded in Table~\ref{table:dis_result}. Not surprisingly, GFR is the best performer, followed by RFWLLFR. It is worth noting that in all cases, RFWLLFR performs best when the depth of Fr\'echet trees is $3$ (under the constraint $3 \sim \lceil \log_2 n \rceil$) and only one feature is randomly selected at each internal node. In fact, if the input space is not partitioned, \emph{i.e.}, the Fr\'echet trees are of depth $1$, RFWLLFR will do the same thing as GFR except RFWLLFR has a subsampling process from the training data. For setting I-2, as the Fr\'echet regression function is nonlinear, the performance of GFR is the worst. And we observe that the performance of GFR can not be improved significantly by simply increasing the number of training samples.  For the low-dimensional case,  the best performance is concentrated in  RFWLLFR, indicating that it is easier to capture nonlinear signals. But as the dimension of $X$ increases, the requirement of data size increases rapidly for local methods and the fitting becomes more challenging.  Instead, a more straightforward method RFWLCFR begins to outperform RFWLLFR. LFR is a local method relying heavily on kernel smoothing.  The corresponding function in ``frechet'' package can only handle the case where the dimension of $X$ is less than $3$. When $p=2$, the method LFR  does not perform as well as  RFWLCFR and RFWLLFR. For the effect of noise size $\sigma$ and the case that the components of $X$ are correlated, please refer to Appendix~\ref{sec B.2}.
	
\begin{figure}[t!]
		\centering
		\includegraphics[width=0.8\linewidth]{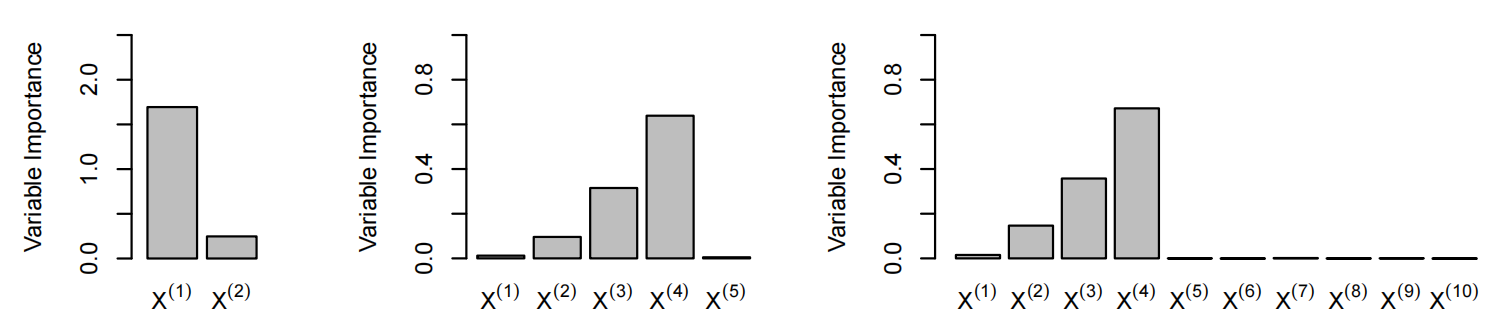}
		\includegraphics[width=0.8\linewidth]{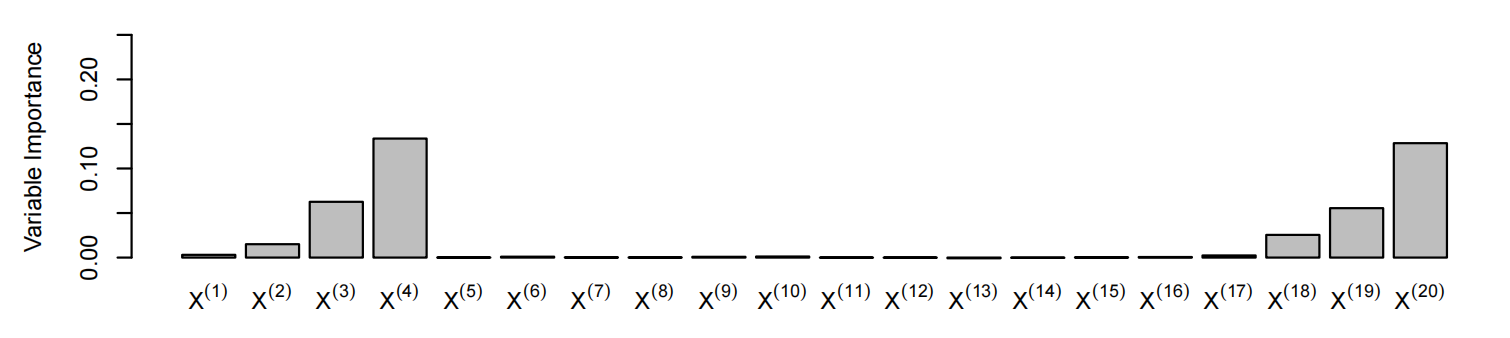}
		\caption{The variable importance for all variables of setting I-1.}
		\label{score_dis}
	\end{figure}

	{As variable importance is readily apparent in setting I-1, we use this setting to examine the variable importance measure introduced in Section~\ref{add sec}. Our analysis concentrates on four cases of $(n, p)$: $(2, 100), (5, 200), (10, 500)$ and $(20, 1000)$. Following the Algorithm~\ref{importance}, we calculate the importance for all variables in one simulation run, as presented in Figure~\ref{score_dis}. It is evident that the importance rankings closely align with the truth. This indicates the ability of our method to accurately discern the order of importance among the feature variables for predicting the response values.}
	
	\subsection{Fr\'echet Regression for Symmetric Positive-definite Matrices}
	Let $(\Omega,d)$  be the metric space  $\mathcal{S}_m^{+}$ of $m \times m$ symmetric positive-definite (SPD) matrices endowed with metric $d$. There are many options for metrics, this section focuses on the Log-Cholesky metric and the affine-invariant metric. For a matrix $Y$, let $\lfloor Y\rfloor$ denote the strictly lower triangular matrix of $Y$,  $\mathbb{D}(Y)$ denote the diagonal part of $Y$ and $\|Y\|_{F}$ denote the Frobenius norm. It is well known that if $Y$ is a symmetric positive-definite matrix, there is a lower triangular matrix $P$ whose diagonal elements are all positive such that $PP^{\T}=Y$.  This $P$ is called the Cholesky factor of $Y$, devoted as $\mathscr{L}(Y)$.
	
	For an $m \times m$ symmetric matrix $A$, $\exp (A)=I_{m}+\sum_{j=1}^{\infty} \frac{1}{j !} A^{j}$ is a symmetric positive-definite matrix. Conversely, for a symmetric positive-definite matrix $Y$, the matrix logarithmic map is $\log (Y)= A$ such that $\exp (A)=Y$.
	
	For two symmetric positive-definite matrices $Y_{1}$ and $Y_{2}$, the Log-Cholesky metric \citep{lin2019riemannian} is defined by
	\begin{align*}
		d_L(Y_1, Y_2)=d_{\mathcal{L}}(\mathscr{L}(Y_1), \mathscr{L}(Y_2)),
	\end{align*}
	where $d_{\mathcal{L}}(P_1,P_2)=\{\left\| \lfloor P_1 \rfloor-\lfloor P_2 \rfloor\right\|_{F}^{2}+\left\| \log \mathbb{D}(P_1)-\log \mathbb{D}(P_2)\right\|_{F}^{2}\}^{1 / 2}.$
	And the affine-invariant metric \citep{moakher2005differential, pennec2006riemannian} is defined by
	\begin{align*}
		d_A(Y_1, Y_2)=\left\|\log \left(Y_1^{-1 / 2} Y_2 Y_1^{-1 / 2}\right)\right\|_{F}.
	\end{align*}
	In practical applications, we need to choose the appropriate metric according to the need.  The Log-Cholesky metric is faster in the calculation, while the affine-invariant metric has the congruence invariance property. More discussion refers to \cite{lin2019riemannian}.
	
	We generate $X_1,\ldots,X_n$ from the uniform distribution $\mathcal{U}([0, 1]^p)$. And the response $Y$ is generated via symmetric matrix variate normal distribution \citep{zhang2021dimension}. Consider the simplest case, we say an $m \times m$ symmetric matrix $A \sim \mathcal{N}_{m m}(M ; \sigma^2)$ if $A=\sigma Z+M$ where $M$ is an $m \times m$ symmetric matrix  and  $Z$ is an  $m \times m$ symmetric random matrix with independent $\mathcal{N}(0, 1)$ diagonal elements and $ \mathcal{N}(0, 1/2)$ off-diagonal elements. We consider the following settings with $Y$ being SPD matrices.
	
	Setting II-1:
	$$
	\log (Y) \sim \mathcal{N}_{m m}\left(D(X), \sigma^{2}\right)
	$$
	with 
	$D(X)= \left(\begin{array}{cc}
		1 & \rho(X) \\
		\rho(X) & 1
	\end{array}\right),  \rho(X)= \cos\left(4  \pi (\beta^{\T}X) \right)$.
	The choice of $\beta$ corresponds to $p=2, 5, 10, 20$ is the same as setting I-1.

	Setting II-2:
	$$
	\log (Y) \sim \mathcal{N}_{m m}\left( D(X), \sigma^{2}\right)
	$$
	with $D(X)=\left(\begin{array}{ccc}
		1 & \rho_{1}(X) & \rho_{2}(X) \\
		\rho_{1}(X) & 1 & \rho_{1}(X) \\
		\rho_{2}(X) & \rho_{1}(X) & 1
	\end{array}\right), \rho_1(X)=0.8 \cos\left(4  \pi (\beta_1^{T}X) \right)$ and $\rho_2(X)= 0.4\cos$ $\left(4  \pi (\beta_2^{T} X) \right)$. The choice of $(\beta_1, \beta_2)$ corresponds to $p=2, 5, 10, 20$ is the same as setting I-2.

\begin{table}[t!]
	\centering
	\resizebox{0.83\columnwidth}{!}{
		\begin{tabular}{ccccccccccccccc}
			\toprule
			Model & $(p,n)$  & GFR & LFR & IFR & RFWLCFR/FRF & RFWLLFR\\
			\midrule
			\multirow{8}{*}{II-1}
			&$(2, 100)$ & 1.264 (0.116) & \textbf{0.088} (0.039) &0.981 (0.280)& 0.177  (0.061) & 0.128 (0.072)  \\
			&$(2, 200)$ & 1.209 (0.089) & \textbf{0.038} (0.016) &0.863 (0.283) & 0.082  (0.019) & 0.054 (0.019)  \\
			&$(5, 200)$ & 1.281 (0.115) & NA &1.202 (0.180)& 0.507  (0.082) & \textbf{0.397} (0.096)  \\
			&$(5, 500)$ & 1.267 (0.104) & NA &1.167 (0.216) & 0.299  (0.046) & \textbf{0.190} (0.040)  \\
			&$(10, 500)$ & 1.279 (0.104) & NA &1.252 (0.115) & 0.586  (0.101) & \textbf{0.575} (0.101)  \\
			&$(10, 1000)$ & 1.253 (0.096) & NA &1.242 (0.099)& 0.420  (0.084) & \textbf{0.407} (0.082)  \\
			&$(20, 1000)$ & 0.973 (0.114) & NA &0.971(0.110) & 0.623  (0.088) & \textbf{0.485} (0.075)  \\
			&$(20, 2000)$ & 0.956 (0.122) & NA &0.591(0.073)& 0.522  (0.076) & \textbf{0.379} (0.058)  \\
			\midrule
			\multirow{8}{*}{II-2}
			&$(2, 100)$ & 1.932 (0.135) & \textbf{0.240} (0.098) &NA& 0.367  (0.068) & 0.283 (0.079)  \\
			&$(2, 200)$ & 1.898 (0.152) & \textbf{0.109} (0.020) &NA& 0.188  (0.035) & 0.143 (0.032)  \\
			&$(5, 200)$ & 1.980 (0.136) & NA &NA& 0.855  (0.114) & \textbf{0.674} (0.116)  \\
			&$(5, 500)$ & 1.935 (0.143) & NA &NA& 0.543  (0.065) & \textbf{0.369} (0.048)  \\
			&$(10, 500)$ & 1.971 (0.136) & NA &NA& 1.057  (0.155) & \textbf{1.056} (0.134)  \\
			&$(10, 1000)$ & 1.949 (0.140) & NA &NA& 0.845  (0.134) & \textbf{0.839} (0.119)  \\
			&$(20, 1000)$ & 1.970 (0.116) & NA &NA& \textbf{1.251} (0.200) & 1.362 (0.171)  \\
			&$(20, 2000)$ & 1.962 (0.140) & NA &NA& \textbf{1.071} (0.178) & 1.191 (0.158)  \\
			\bottomrule
		\end{tabular}
	}
	\caption{Average MSE (standard deviation) of different methods for setting II-1,2 with $\sigma=0.2$ and Log-Cholesky metric over $100$ simulation runs. Bold-faced numbers indicate the best performers.}
	\label{table:cov_Cho}
\end{table}

We again compare our methods with GFR, LFR and FRF. {Additionally, since the setting II-1 is a single index model, we include IFR as another competitor.} Since the Log-Cholesky distance between two symmetric positive-definite matrices is essentially the Frobenius distance between the matrices after some transformations. Therefore, similar to the regression for distributions, RFWLCFR and FRF would have the same output. Results about setting II-1,2 with Log-Cholesky metric are shown in Table~\ref{table:cov_Cho}.  {In setting II-1, IFR performs better than GFR, but still falls short of our two methods. IFR learns coefficients of a single index model within $500$ randomly generated direction vectors. This finite traversal optimization approach incurs a loss in precision.} Taken together, LFR has the best performance when $p=2$. But when $p > 2$, RFWLLFR performs the best in most cases. The results with the affine-invariant metric summarized in Table~\ref{table:cov_Rim} also advocate RFWLLFR except for the high dimensional case of setting II-2. Moreover, for the affine-invariant metric, we observe slight differences between RFWLCFR and FRF.

	\begin{table}[t!]
		\centering
		\resizebox{0.83\columnwidth}{!}{
			\begin{tabular}{ccccccccccccccc}
				\toprule
				Model & $(p,n)$ & FRF & RFWLCFR & RFWLLFR\\
				\midrule
				\multirow{8}{*}{II-1}
				&$(2, 100)$ & 0.164133 (0.041988) & 0.164132 (0.041991) & \textbf{0.124380} (0.046016)  \\
				&$(2, 200)$ & 0.080703 (0.015133) & 0.080707 (0.015134) & \textbf{0.063739} (0.016001)  \\
				&$(5, 200)$ & 0.408030 (0.063235) & 0.408010 (0.063239) & \textbf{0.332478} (0.073828)  \\
				&$(5, 500)$ & 0.243664 (0.035725) & 0.243662 (0.035725) & \textbf{0.166254} (0.029714)  \\
				&$(10, 500)$ & 0.501018 (0.079189) & 0.500904 (0.079182) & \textbf{0.464811} (0.079191) \\
				&$(10, 1000)$ & 0.366349 (0.069003) & 0.366257 (0.068997) & \textbf{0.331014} (0.067757)  \\
				&$(20, 1000)$ & 0.502299 (0.073056) & 0.502238 (0.073048) & \textbf{0.390836} (0.057455)  \\
				&$(20, 2000)$ & 0.425098 (0.062546) & 0.425046 (0.062547) & \textbf{0.304537} (0.044358)  \\
				\midrule
				\multirow{8}{*}{II-2}
				&$(2, 100)$ & 0.286285 (0.059663) & 0.286285 (0.059655) & \textbf{0.238548} (0.062099)  \\
				&$(2, 200)$ & 0.154080 (0.025816) & 0.154085 (0.025813) & \textbf{0.128763} (0.023595)  \\
				&$(5, 200)$ & 0.627837 (0.088528) & 0.627769 (0.088545) & \textbf{0.499721} (0.087616)  \\
				&$(5, 500)$ & 0.396012 (0.046278) & 0.395999 (0.046281) & \textbf{0.280171} (0.035996)  \\
				&$(10, 500)$ & 0.794744 (0.128331) & 0.794220 (0.128334) & \textbf{0.762684} (0.100195)  \\
				&$(10, 1000)$ & 0.638420 (0.110770) & 0.637765 (0.110686) & \textbf{0.605356} (0.089589)  \\
				&$(20, 1000)$ & 0.927755 (0.147346) & \textbf{0.927363} (0.147369) & 0.991135 (0.111894)  \\
				&$(20, 2000)$ & 0.795528 (0.141159) & \textbf{0.795009} (0.141150) & 0.878269 (0.118397)  \\
				\bottomrule
			\end{tabular}
		}
	\caption{Average MSE (standard deviation) of different methods for setting II-1,2 with $\sigma=0.2$ and affine-invariant metric over $100$ simulation runs. Bold-faced numbers indicate the best performers.}
	\label{table:cov_Rim}
	\end{table}
   
   \begin{figure}[t!]
   	\centering
   	\includegraphics[width=0.8\linewidth]{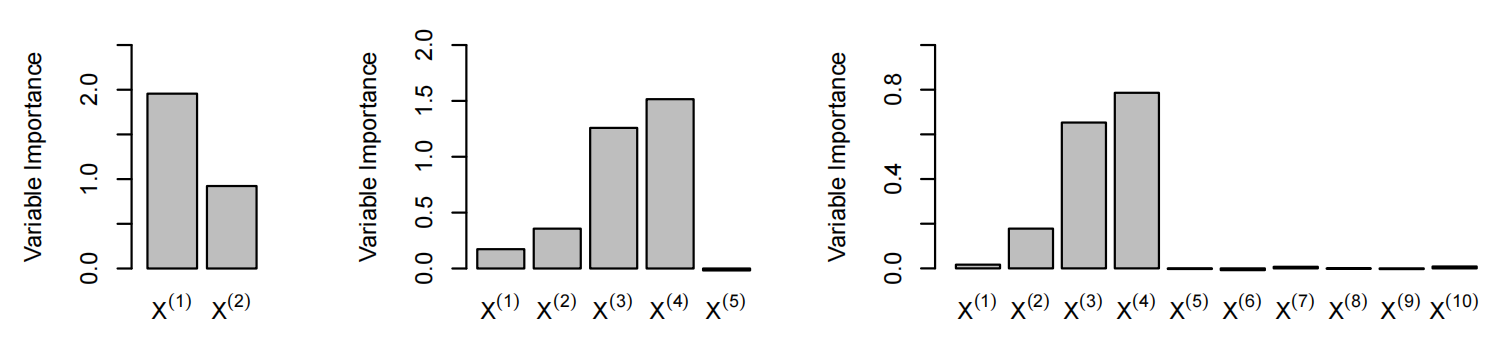}
   	\includegraphics[width=0.8\linewidth]{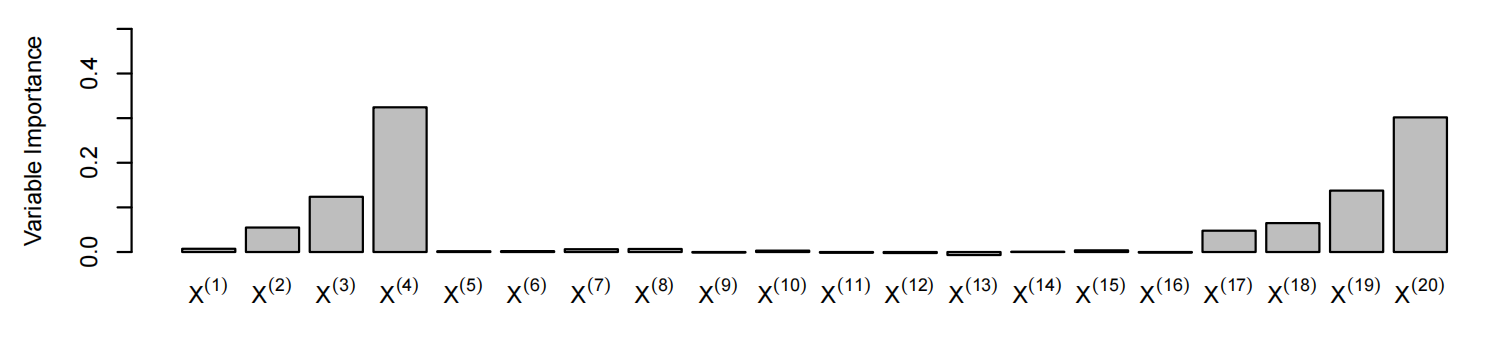}
   	\caption{The variable importance for all variables of setting II-1.}
   	\label{score_spd}
   \end{figure}
	
	{Similar to Section~\ref{sec 5.1}, we further evaluate the validity of the proposed variable importance method in the scenario of matrix responses. We consider the setting II-1 with the Log-Cholesky metric and still pick $(n, p)$ for $(2, 100), (5, 200), (10, 500)$ and $(20, 1000)$. The results are illustrated in Figure~\ref{score_spd}. Once again, it can be observed that our method produces nearly accurate importance rankings for the variables. }

	\subsection{Fr\'echet Regression for Spherical Data}
	Let $(\Omega,d)$ be the metric space $\mathbb{S}^2$ of sphere data endowed with the geodesic distance $d_{g}$. For two points $Y_1, Y_2 \in \mathbb{S}^2$, the geodesic distance is defined by
	\begin{align*}
		d_{g} \left(Y_{1}, Y_{2}\right)=\arccos \left(Y_{1}^{\T}Y_{2}\right).
	\end{align*}
	We generate i.i.d $X_1,\ldots,X_n \sim \mathcal{U}\left([0, 1]^p\right)$.  And $Y_{i}$ are generated by the following two settings.
	
	\begin{table}[b!]
		\centering
		\resizebox{0.83\columnwidth}{!}{
			\begin{tabular}{ccccccccccccccc}
				\toprule
				Model & $(p,n)$ & FRF & RFWLCFR & RFWLLFR\\
				\midrule
				\multirow{8}{*}{III-1}
				&$(2, 100)$ & 0.032329 (0.007413) & 0.032327 (0.007409) & \textbf{0.019839} (0.005842)  \\
				&$(2, 200)$ & 0.023078  (0.003967) & 0.023079 (0.003967) & \textbf{0.010781}  (0.002704)  \\
				&$(5, 200)$ &0.030237 (0.004693) & 0.030237 (0.004693) & \textbf{0.017935} (0.003892)  \\
				&$(5, 500)$ & 0.021684 (0.002725) & 0.021683 (0.002724) & \textbf{0.012164} (0.001998)  \\
				&$(10, 500)$ &0.035012 (0.004252) & 0.035018 (0.004253) & \textbf{0.013339} (0.001915)  \\
				&$(10, 1000)$ & 0.027184 (0.003063) & 0.027188 (0.003065) & \textbf{0.010851} (0.001389)  \\
				&$(20, 1000)$ & 0.034698 (0.003879) & 0.034701 (0.003879) & \textbf{0.033069} (0.004379)  \\
				&$(20, 2000)$ & 0.029362 (0.003503) & 0.029362 (0.003502) & \textbf{0.028837} (0.003453)  \\
				\midrule
				\multirow{8}{*}{III-2}
				&$(2, 100)$ &0.010498 (0.002954) &0.010501  (0.002954) & \textbf{0.004226} (0.001735)  \\
				&$(2, 200)$ & 0.007982 (0.001836) & 0.007984 (0.001837) & \textbf{0.003150} (0.000956)  \\
				&$(5, 200)$ &0.008893  (0.001612) &0.008894 (0.001612) & \textbf{0.005192}  (0.001418)  \\
				&$(5, 500)$ & 0.006562 (0.001216) &0.006563 (0.001216) & \textbf{0.002946} (0.000637)  \\
				&$(10, 500)$ & 0.008936 (0.001378) & 0.008938 (0.001378) & \textbf{0.005266} (0.000962)  \\
				&$(10, 1000)$ & 0.007110 (0.000951) & 0.007111 (0.000952) & \textbf{0.004190} (0.000626)  \\
				&$(20, 1000)$ & 0.009005 (0.001258) & 0.009006 (0.001259) & \textbf{0.006311} (0.000975)  \\
				&$(20, 2000)$ &0.007427 (0.000938) &0.007429 (0.000938) & \textbf{0.005117} (0.000692)  \\
				\bottomrule
			\end{tabular}
		}
	\caption{Average MSE (standard deviation) of different methods for setting III-1,2 over 100 simulation runs. Bold-faced numbers indicate the best performers.}
	\label{table:sph_result}
	\end{table}
	
	Setting III-1:
	Let the Fr\'echet regression function be
	\begin{align*}
		m_{\oplus}\left(X\right)=\big(&\{1-(\beta_{1}^{\T}X)^{2}\}^{1 / 2} \cos (\pi (\beta_{2}^{\T}X)), \{1-(\beta_{1}^{\T}X)^{2}\}^{1 / 2} \sin (\pi (\beta_{2}^{\T}X)), \beta_{1}^{\T}X\big)^{\T}.
	\end{align*}
	We generate binary Normal noise $\varepsilon_{i}$ on the tangent space $T_{m_{\oplus}\left(X_{i}\right)} \mathbb{S}^2$, then map $\varepsilon_{i}$ back to $\mathbb{S}^2$ by Riemannian exponential map to get $Y_i$. Specifically, we first independently generate $\delta_{i 1}, \delta_{i 2} \stackrel{i i d}{\sim} \mathcal{N}\left(0,0.2^{2}\right)$, then let $\varepsilon_{i}=\delta_{i 1} v_{1}+\delta_{i 2} v_{2}$, where $\left\{v_{1}, v_{2}\right\}$ forms an orthogonal basis of tangent space $T_{m_{\oplus}\left(X_{i}\right)} \mathbb{S}^2$. Then $Y_{i}$ can be generated by
	$$
	Y_{i}=\operatorname{Exp}_{m_{\oplus}\left(X_{i}\right)}\left(\varepsilon_{i}\right)=\cos \left(\left\|\varepsilon_{i}\right\|\right) m_{\oplus}\left(X_{i}\right)+\sin \left(\left\|\varepsilon_{i}\right\|\right) \frac{\varepsilon_{i}}{\left\|\varepsilon_{i}\right\|},
	$$
	where $\|\cdot\|$ is the Euclidean norm. Consider the following four kinds of dimensions\\
	(i) $p=2$: $\beta_1=(1, 0),  \beta_2=(0, 1)$;\\
	(ii) $p=5,10,20$: $\beta_1=\left(0.1, 0.2, 0.3, 0.4,  0, \dots, 0\right),  \beta_2=\left(0, \dots, 0, 0.1, 0.2, 0.3, 0.4\right)$.
	
	Setting III-2:
	Consider the following model
	\begin{align*}
		Y_{i}=\big(&\sin (\beta_{1}^{\T}X_i+\varepsilon_{i 1}) \sin (\beta_{2}^{\T}X_i+\varepsilon_{i 2}), \sin (\beta_{1}^{\T}X_i+\varepsilon_{i 1}) \cos (\beta_{2}^{\T}X_i+\varepsilon_{i 2}), \cos (\beta_{1}^{\T}X_i+\varepsilon_{i 1})\big)^{\T},
	\end{align*}
	where the random noise $\varepsilon_{i 1}, \varepsilon_{i 2} \stackrel{i i d}{\sim} \mathcal{N}\left(0,0.2^{2}\right)$ are generated independently. The four situations corresponding to $p=2, 5, 10, 20$ are the same as setting III-1.
	
\begin{figure}[t!]
		\centering
		\includegraphics[width=0.9\linewidth]{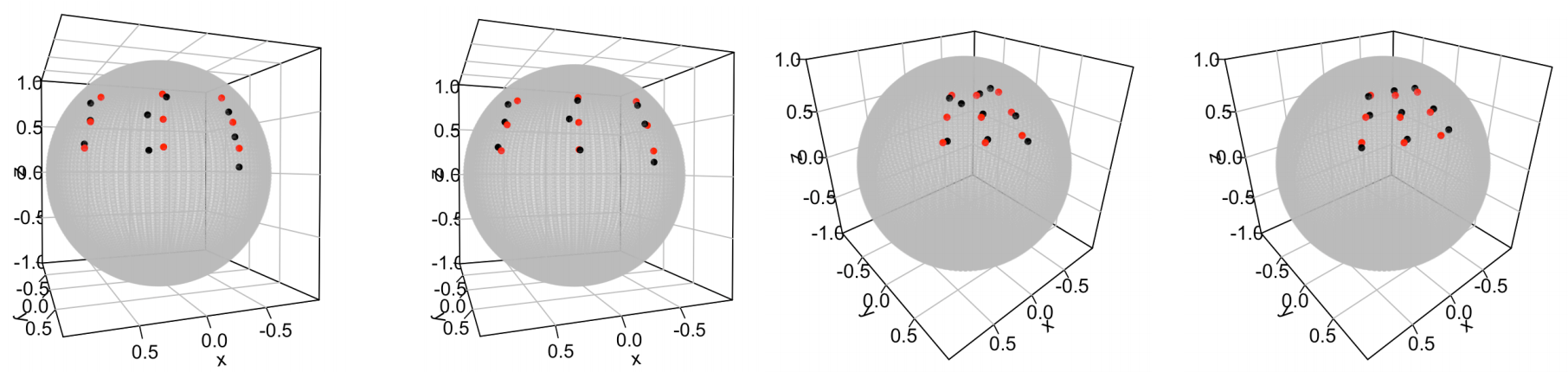}
		\caption{The plots of predictions of ${m}_{\oplus}(x)$ given by RFWLCFR (the 1st and 3rd panels) and RFWLLFR (the 2nd and 4th panels) in a simulation run of $p=2,n=200$. The left two panels show the results of setting III-1, while the right two show the results of setting III-2. The red points represent real points, and the black points represent predicted points.}
		\label{plot:sph_result}
	\end{figure}

	Setting III-1  is similar to \cite{petersen2019frechet} and \cite{zhang2021dimension}, and setting III-2 is similar to \cite{ying2022frechet}. For Fr\'echet regression with sphere data, we focus on the comparison of FRF, RFWLCFR and RFWLLFR. RFWLCFR and FRF will have different outputs under the geodesic distance $d_g$. We summarize the results in Table~\ref{table:sph_result}. RFWLLFR performs best in all cases. 
	To vividly describe the performance of RFWLCFR and RFWLLFR, Figure~\ref{plot:sph_result} exhibits the prediction of nine given testing points with $p=2$ and $n=200$ for setting III-1,2, which verifies the advantage of RFWLLFR.
	
	\section{Real Application} \label{sec6}
	In this section, we use New York taxi data and mortality data to validate the advanced performance of our methods in practical applications. 
	\subsection{New York Taxi Data}
	The New York City Taxi and Limousine Commission provides detailed records on yellow taxi rides, including pick-up and drop-off dates and times, pick-up and drop-off locations, trip distances, payment types, and other information. The data can be downloaded from \url {https://www1.nyc.gov/site/tlc/about/tlc-trip-record-data.page}. In line with \cite{dubey2020functional}, we transform the raw data into network data  (adjacency matrices), where nodes represent zones and edges are weighted by the number of taxi rides that picked up in one zone and dropped off in another within a single hour. Specifically, we take the following steps to gather adjacency matrices:
	
	1. Due to resource constraints, we only use data from January and February 2019 (59 days).
	
	2. We further filter the observations to only include pick-ups and drop-offs that occurred within Manhattan (excluding islands).
	
	3. We divide Manhattan into 10 zones and labeled them according to \cite{dubey2020functional}. Details can be found in Appendix~\ref{sec C}. Then each network has $10$ nodes, and the corresponding adjacency matrix has dimensions $10 \times 10$.
	
	4. For each hour, we collect the number of pairwise connections between nodes based on pick-ups and drop-offs, which corresponds to the weights between nodes. We normalize the weights by the maximum edge weight in each hour, scaling them to the range $[0,1]$.
	
	We acquire a total of $1416$ adjacency matrices of $10 \times 10$, which describe the taxi movements between zones in Manhattan. To facilitate the Fr\'echet regression analysis for network responses, we transform these matrices into SPD matrices by applying the matrix exponential mapping $\exp(\cdot)$ to them. Additionally, from the taxi data, we collect nine potential features with values averaged over each hour:
	\begin{itemize}
		\item {\em Ave. Distance}: mean distance traveled, standardized
		\item {\em Ave. Fare}: mean fare, standardized
		\item {\em Ave. Passengers}: mean number of passengers, standardized
		\item {\em Ave. Tip}: mean tip, standardized
		\item {\em Cash}: sum of cash indicators for type of payment, standardized
		\item {\em Credit}: sum of credit indicators for type of payment, standardized
		\item {\em Dispute}: sum of  dispute indicators for type of payment, standardized
		\item {\em Free}: sum of free indicators for type of payment, standardized
		\item {\em Late Hour}: indicator for the hour being between 11pm and 5am, standardized 
	\end{itemize}
	We also gather weather data for January and February $2019$ from \url {https://www.wunderground.com/history/daily/us/ny/new-york-city/KLGA/date}, which yields $5$ weather variables as potential features: 
	\begin{itemize} 
		\item {\em Day’s Ave. Temp}: daily mean temperature, standardized
		\item {\em Day’s Ave. Humid}: daily mean humidity, standardized
		\item {\em Day’s Ave. Wind}: daily mean wind speed, standardized
		\item {\em Day’s Ave. Press}: daily mean barometric pressure, standardized
		\item {\em Day’s Total Precip}: daily total precipitation, standardized
	\end{itemize}
	
	In total, we have $14$ potential features. The data set consisting of $1416$ samples is partitioned randomly into three parts for Fr\'echet regression: a training set of size $850$, a validation set of size $283$, and a testing set of size $283$, following a ratio of $6:2:2$. We train GFR, LFR, IFR, RFWLCFR, and RFWLLFR on the training set using the Log-Cholesky metric and fine-tune their hyperparameters on the validation set. Subsequently, we retrain these methods on the combined training and validation sets of size $1133$ using the selected hyperparameters to obtain the final models. Their performance is evaluated on the testing set by computing the mean squared errors based on the Log-Cholesky metric.  {However, the R-package for LFR is only applicable when the dimension of the feature space does not exceed $2$. This limitation restricts the use of LFR in the current regression task. An effective solution is to consider the Fr\'echet sufficient dimension reduction method called the weighted inverse regression ensemble (WIRE), as proposed by \cite{ying2022frechet}.  Unfortunately, the structural dimension of the central space is estimated to be $3$. To ensure the availability of LFR, we are compelled to discard the third sufficient dimension reduction direction and project the original $14$-dimensional feature vector onto a $2$-dimensional subspace. This inevitably results in some loss of information. Specifically, the first two sufficient dimension reduction directions obtained through WIRE are as follows:
		\begin{equation}\label{taxi direction}
			\begin{aligned}
				\hat{\beta}_1=(-0.911, 0.049, &-0.152, -0.095, 0.303, 0.013, 0.035, -0.079,\\ &0.157, 0.076, 0.007, -0.021, 0.058, 0.033)^{\T};\\
				\hat{\beta}_2=(0.125, -0.057, &0.178, 0.354, 0.740, 0.258, -0.091, -0.220,\\ & -0.390, -0.014, 0.020, -0.005, -0.027, 0.001)^{\T}.
			\end{aligned}
		\end{equation}
		The two directions transform $X$ into a $2$-dimensional vector $(\hat{\beta}_1^{\T}X, \hat{\beta}_2^{\T}X)$ as the input of LRF. Additionally, the above result of the Fr\'echet sufficient dimension reduction shows that the underlying model can not be a single index model. However, despite this, we still intend to employ the IFR method for the sake of comparative analysis.}
	
	\begin{figure}[t!]
		\centering
		\includegraphics[width = 0.98\linewidth]{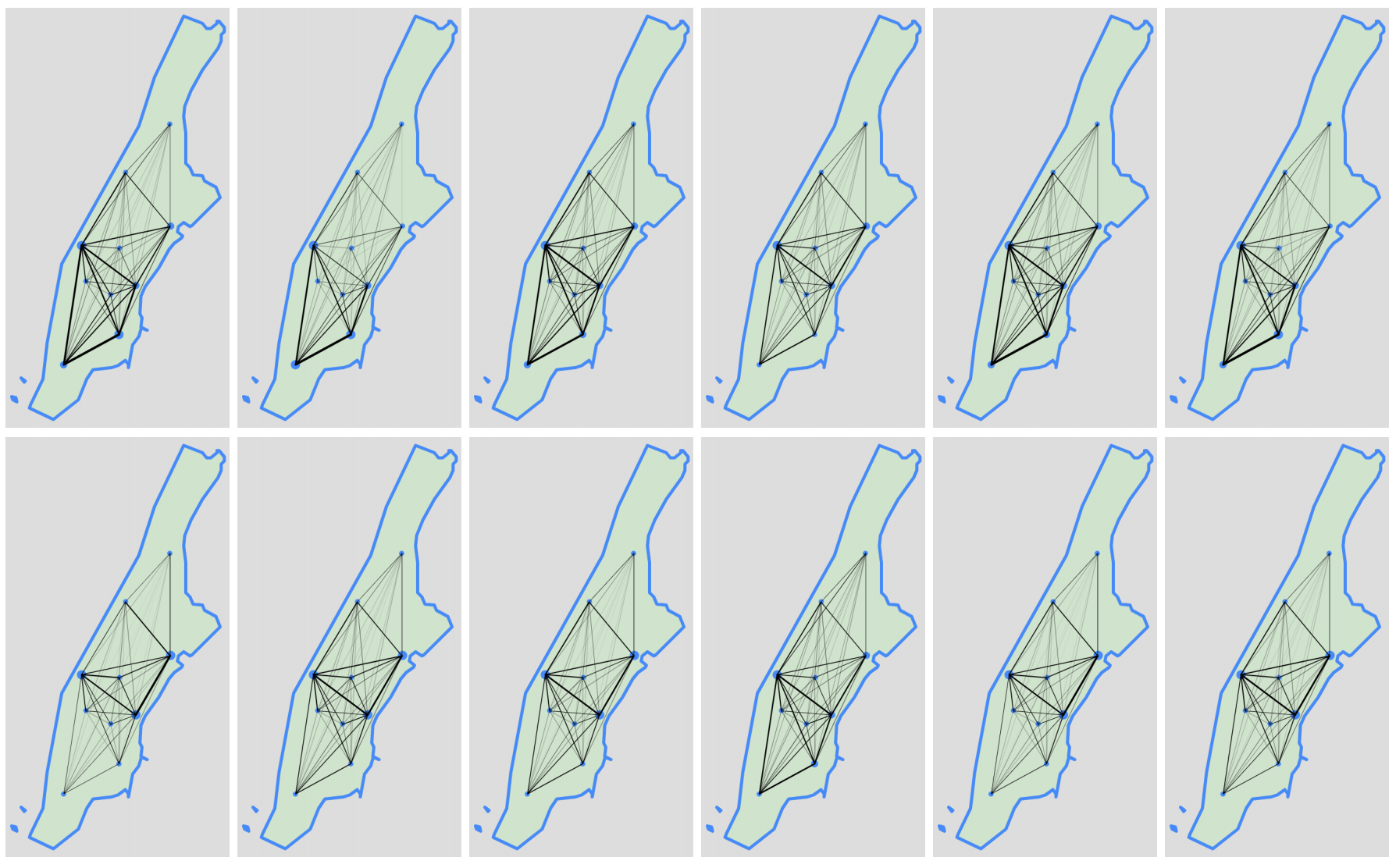}
		\caption{True (left) and fitted (right, in order of GFR, LFR  after dimension reduction, IFR, RFWLCFR, RFWLLFR) networks with $10$ zones for the remaining two test samples. The thickness of the edges connecting vertices corresponds to their weights, while the size of vertices represents the total traffic volume within each zone.}
		\label{taxi network2}
	\end{figure}

   \begin{figure}[t!]
   	\centering
   	\includegraphics[width = 13cm]{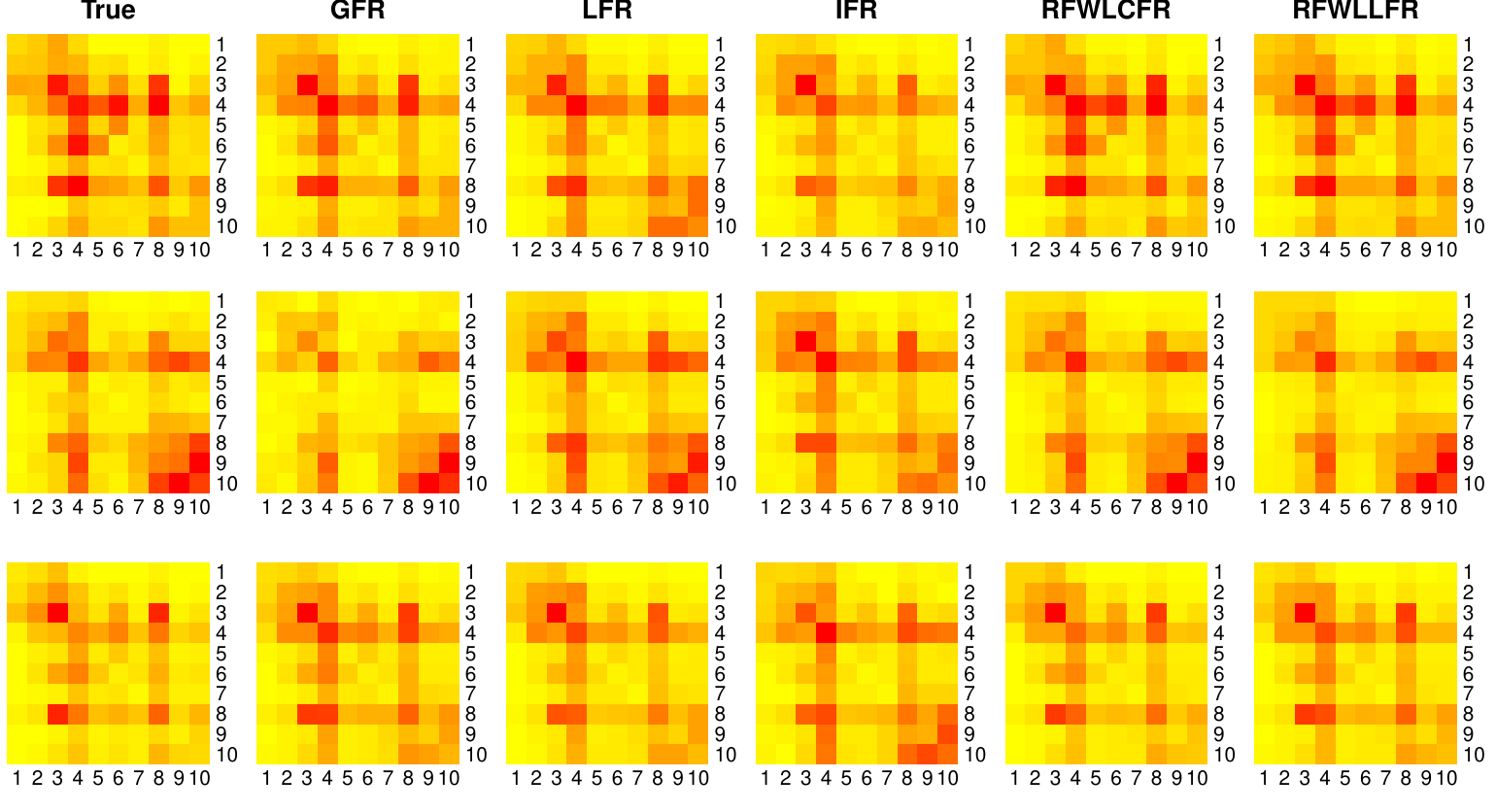}
   	\includegraphics[width = 0.8cm]{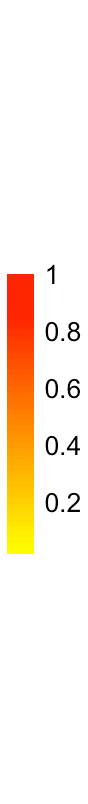}
   	\caption{Networks with $10$ zones represented as heatmaps for the three test samples.}
   	\label{heatmap}
   \end{figure}
	
	{With the above preparations, we obtain the following testing errors:  $1.377$ for GFR, $0.944$ for LFR, $2.620$ for IFR, $0.568$ for RFWLCFR, and $0.576$ for RFWLLFR. In light of the results, RFWLCFR exhibits the highest prediction accuracy, followed closely by RFWLLFR. The superior performance of LFR compared to GFR strongly suggests the presence of a nonlinear regression relationship. The poor performance of IFR can be easily understood, as the central space discussed before has a structural dimension of $3$, rendering the single index method ineffective. As expected, IFR performs worse than LFR, which benefits from two sufficient dimension reduction directions.} We can further obtain the predicted taxi ride networks by applying the inverse mapping (matrix logarithmic map) $\log(\cdot)$ to the predicted matrices given by the methods described above. To visually illustrate the disparities in the outcomes of these methods, we randomly select three samples from the testing set and plot their corresponding true and predicted networks, as depicted in Figure~\ref{taxi network2}. Please note that the plots for the first sample have been previously shown in Figure~\ref{taxi network} and are therefore excluded here. Obviously, all the regression methods effectively capture the structural characteristics and weight information of true networks.
	But our methods stand out in their capacity to handle intricate details, yielding predicted outcomes that closely approximate the true networks. Corresponding heatmaps (Figure~\ref{heatmap}) have also been generated to complement the visualization. These results numerically and visually verify the superiority of our methods and demonstrate their potential in complex network learning.

    \begin{figure}[t]
    	\centering
    	\includegraphics[width=0.65\linewidth]{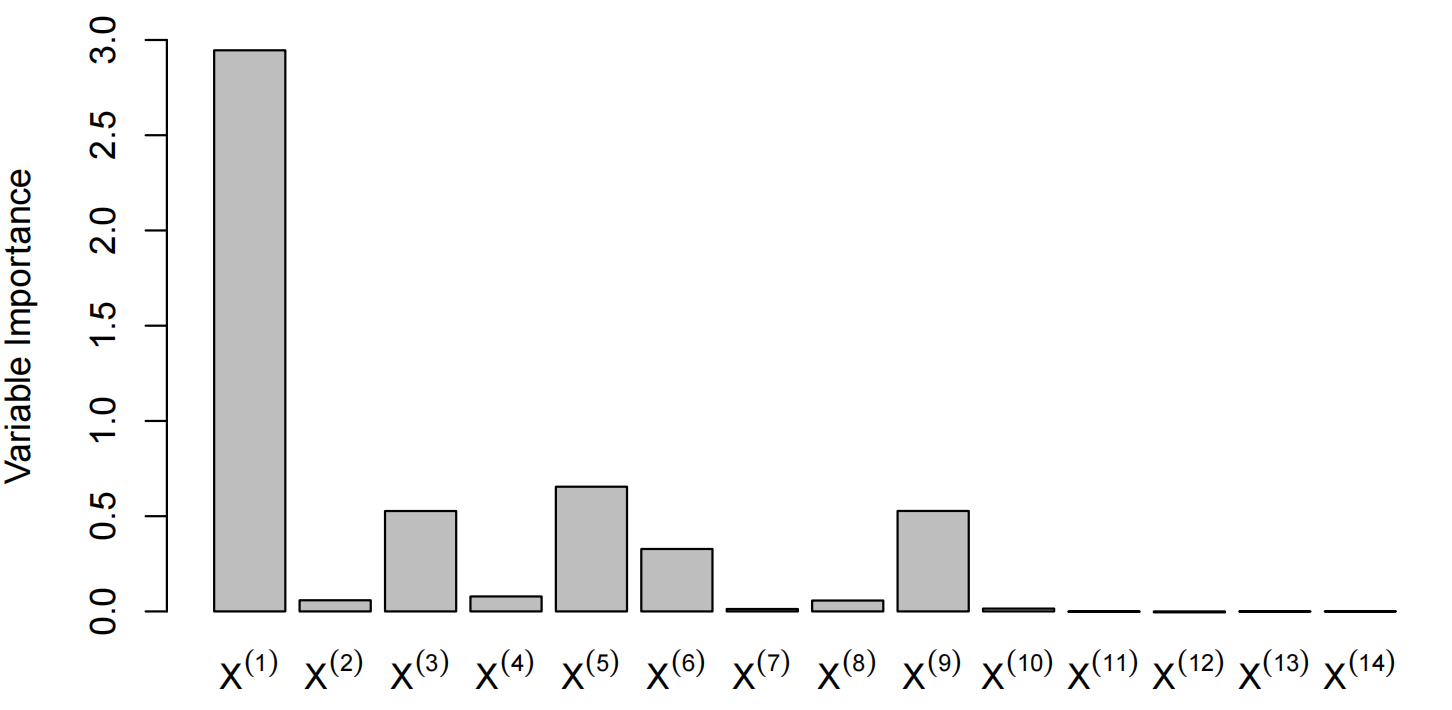}
    	\caption{The variable importance for $14$ features of New York taxi data.}
    	\label{score_taxi}
    \end{figure}
	
	{We now measure the importance of each feature using Algorithm~\ref{importance}.  The result is shown in Figure~\ref{score_taxi}. This ranking of importance appears to be reasonable. Firstly, five weather features ($X^{(10)}$ to $X^{(14)}$) are deemed unimportant. One main reason is that daily averages of these weather features may not accurately capture their hourly impact on taxi traffic. And our data is sourced from the consistent winter season in which daily weather conditions are relatively stable. The limited variability in weather conditions may account for the constrained explanatory power of these weather features in capturing fluctuations in taxi traffic. Another evidence support is the observation that the last five coefficients of the first two sufficient dimension reduction directions, $\beta_1$ and $\beta_2$ in \eqref{taxi direction}, are all notably small. Among these five features, the two most significant ones are Day’s Ave. Temp ($X^{(10)}$) and Day’s Total Precip ($X^{(14)}$), as adverse weather conditions such as low temperatures or rain are more likely to hinder travel. Secondly, disputed and free rides are infrequent events in the raw data, and as such, their influence on the taxi ride network is consequently limited. This helps explain the relatively low importance assigned to Dispute ($X^{(7)}$) and Free ($X^{(8)}$) among the $14$ features. Finally, it can be observed that the four most crucial features are Ave. Distance ($X^{(1)}$),  Cash ($X^{(5)}$), Late Hour ($X^{(9)}$), and Ave. Passengers ($X^{(3)}$). This result can also be explained intuitively. Travel distance and passenger count are typically the two main considerations when people decide whether to choose a taxi for their trip. Cash, as a primary mode of payment, plays a pivotal role in people's travel decisions.  Late Hour emerges as a critical factor shaping taxi ride networks, as late-night travel activity is significantly different from other times. }
	
\begin{figure}[t!]
		\centering
		\includegraphics[width=0.65\linewidth]{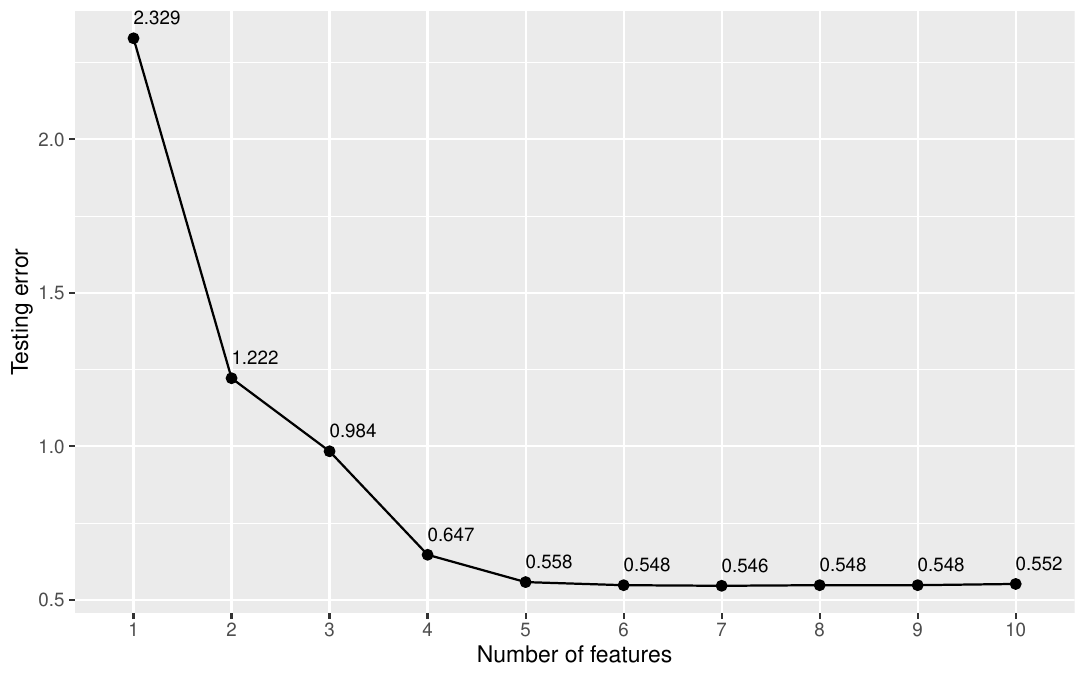}
		\caption{The testing errors with the different number of features.}
		\label{taxi_selection}
	\end{figure}

	{Based on the established variable importance ranking, we further illustrate how to perform variable selection. Here we judge which features can be omitted by comparing the performance of RFWLCFR with different feature subsets on the testing set.  However, evaluating the performance of Fr\'echet regression separately for all possible subsets of the $14$ features is a cumbersome task. If we leverage the feature importance ranking, this task will be greatly simplified. Specifically, we first remove the four least important features ($X^{(11)}$ to $X^{(14)}$). Then we initiate the selection process with the most important feature, and add other features, one by one, to the candidate feature subset in order of their importance among the remaining $10$ variables. We conduct testing with each candidate subset. In this way, we only need $10$ experiments to determine the most appropriate feature subset as the result of our selection. The entire data set consisting of $1416$ samples is divided into training, validation, and testing sets in the same way as before. In the $j$th experiment, we intercept the most important $j$ features to implement RFWLCFR, tune the hyperparameters using the validation set, and calculate the testing error using the testing set. The results of $10$ experiments are summarized in Figure~\ref{taxi_selection}. Notably,  the Fr\'echet regression with the first seven most important features achieves the lowest testing error. Adding additional features does not improve prediction accuracy but instead increases computational complexity and complicates interpretation. Consequently, the final selected seven variables are Ave. Distance, Cash, Late Hour, Ave. Passengers, Credit, Ave. Tip, and Ave. Fare. In Figure~\ref{taxi_selection}, the testing error does not exhibit a clear upward trend as the number of features increases. This once again illustrates, to some extent, that our random forest based method can adaptively identify valuable features, and the presence of less relevant features will not interfere too much with its accuracy.}
	
	\subsection{Mortality Data}
	Taking distribution as the outcome of interest allows us to obtain more information than summary statistics. In this part, we apply the two proposed methods to deal with the Fr\'echet regression problem for human mortality distribution. Just like \cite{zhang2021dimension}, we also consider the following $9$ predictor variables that have been standardized:  (1) Population Density: population per square Kilometer; (2) Sex Ratio: number of males per 100 females in the population; (3) Mean Childbearing Age: the average age of mothers at the birth of their children; (4)Gross Domestic Product (GDP) per Capita; (5) Gross Value Added (GVA) by Agriculture: the percentage of agriculture, hunting, forestry, and fishing activities of gross value added; (6) Consumer price index: treat 2010 as the base year; (7) Unemployment Rate; (8) Expenditure on Health (percentage of GDP); (9) Arable Land (percentage of total land area). These variables involve population, economy, health, and geography factors in 2015, which are closely related to human mortality. The data are collected from United Nation Databases (\url {http://data.un.org/}) and UN World Population Prospects $2019$ Databases (\url {https://population.un.org/wpp/Download}).  The life table considered here contains the number of deaths for each single age group from 162 countries in 2015.  We treat the life table data as histograms of death versus age, with bin width equal to one year (the results of the subsequent analysis are similar when bin width is set to five years).  Then the package ``frechet'' helps to transform the histograms into smoothed probability density functions.  For comparison, we try to consider GFR, LFR, IFR. Similarly, before using LFR, WIRE \citep{ying2022frechet} is adopted to achieve sufficient dimension reduction.  The first four largest singular values of the WIRE matrix \citep{ying2022frechet} are $4.486, 0.785, 0.101, 0.066$, and the structural dimension of central space is determined to be $2$. The first two sufficient dimension reduction directions obtained by WIRE are
	\begin{gather*}
		\hat{\beta}_1=(-0.092, -0.084, 0.009, -0.429, 0.806, 0.048, 0.104, -0.364,  -0.076)^{\T};\\
		\hat{\beta}_2=(0.103, 0.079, 0.641, 0.566, 0.415, -0.017, 0.130, 0.243, 0.066)^{\T}.
	\end{gather*}
	The two directions transform the $9$-dimensional feature $X$ into a $2$-dimensional vector $(\hat{\beta}_1^{\T}X, \hat{\beta}_2^{\T}X)$ as the input of LRF. {This also implies that the single index method is inappropriate. Here we drop the use of the IFR method.}
	
	We then perform $ 9$-fold testing to evaluate the performance of all Fr\'echet regression methods. Specifically, we divide the $162$ countries into $9$ parts evenly and conduct $9$ training runs. For each run, one of the $9$ parts is chosen as the testing set and the rest as the training set.  The test errors (mean squared errors based on the Wasserstein distance) obtained for nine runs are averaged for each method under the best choice of hyperparameters. The average test errors are recorded as $56.51$ for GFR, $41.66$ for LFR, $31.79$ for RFWLCFR, and $36.20$ for RFWLLFR. RFWLCFR has the best performance. These large errors reflect that inadequate sample size and large variation across countries increase the difficulty of the $9$-dimensional Fr\'echet regression problem.
	
	\begin{figure}[b!]
		\centering
		\includegraphics[width=0.76\linewidth]{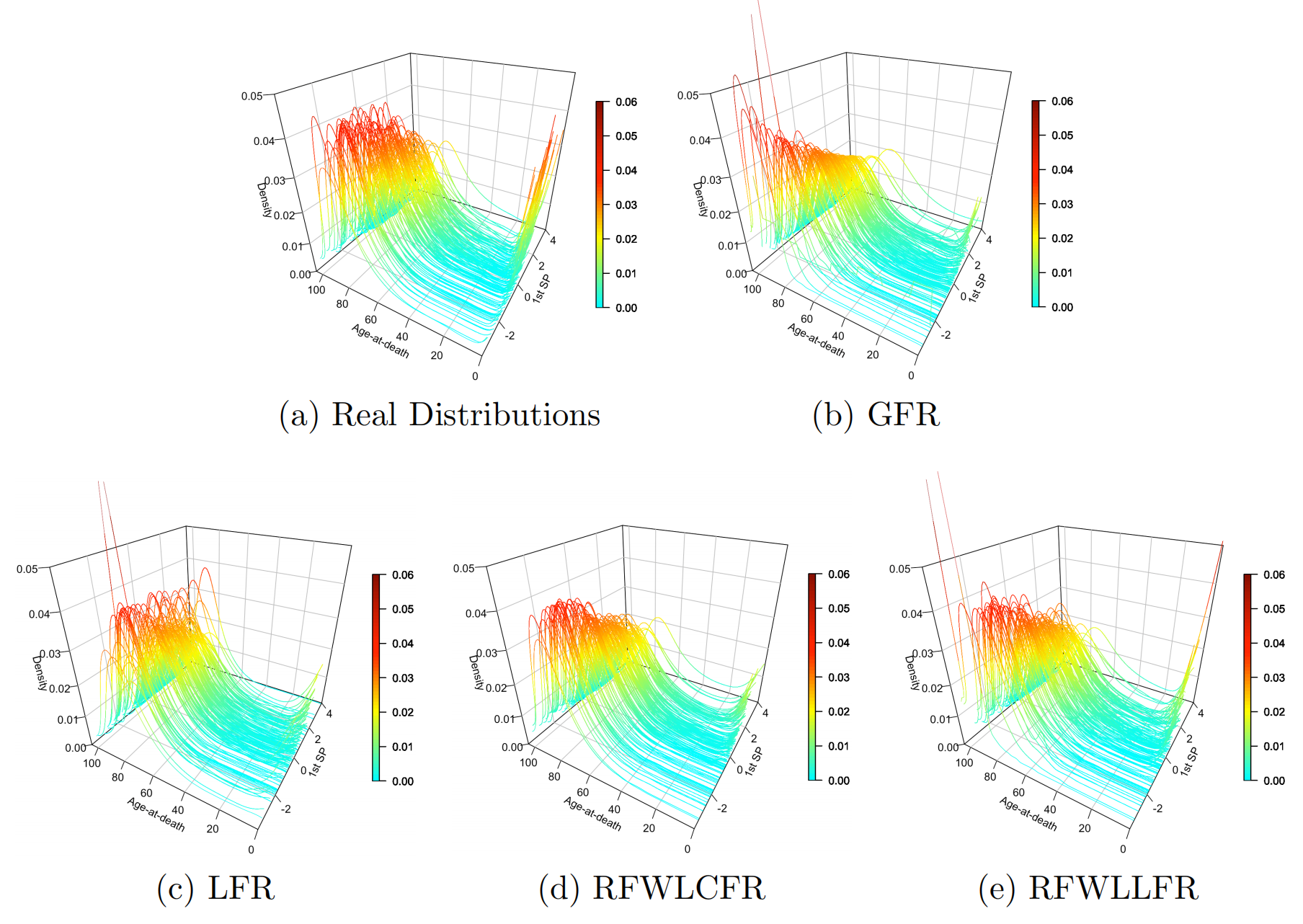}
		\caption{The plot (a) is the real mortality distributions against ${\hat{\beta}_1}^{\T}X$, and the remaining plots are the distributions predicted by each method against ${\hat{\beta}_1}^{\T}X$.}
		\label{plot: real data}
	\end{figure}
	
	To show the performance of each method more vividly, we aid the analysis by plotting the mortality density predictions against the first sufficient predictor ${\hat{\beta}_1}^{\T}X$ (see Fig.~\ref{plot: real data}). For reference, plot (a) of Fig.~\ref{plot: real data} is the smooth real density functions fitted according to the human mortality data. Observations reveal that the first sufficient predictor represents the development degree of a country. Countries with large ${\hat{\beta}_1}^{\T}X$ have backward medical levels and indigent living conditions, resulting in higher infant mortality and lower life expectancy. It is clear from plots that the distribution in the elderly age region ($80 \sim 100$ years old) and the infant age region (near $0$ years old) is challenging to be well estimated. 
	GFR performs poorly in both regions. It underestimates the infant mortality rate but successfully exhibits a tendency for the distribution to concentrate towards the elderly age region as the first sufficient predictor decreases.
	Compared with GFR, LFR based solely on two sufficient dimension reduction directions improves predictions of the elderly age region. However, the shape of the predicted mortality density function does not change significantly with the first sufficient predictor.
	In terms of the overall visual effect, the predictions of RFWLCFR are closest to the real distributions,  but still suffer from large deviations for the infant age region in density functions.
	Among all local methods, RFWLLFR has the best performance in the infant age region. While RFWLLFR performs slightly inferior to RFWLCFR in the elderly age region, but much better than LFR.
	Overall, RFWLCFR has a remarkable advantage in this real data application. It again reflects the fact that RFWLLRF is not always the best choice, especially for complex regression problems with a small amount of data. RFWLCFR tends to be more robust and accurate in these cases.

	\section{Discussion} \label{sec7}
	We propose two highly flexible and complementary locally weighted Fr\'echet regression methods for random object responses residing in a general metric space coupled with relatively high-dimensional Euclidean predictors. These methods employ adaptive random forest weights that effectively mitigate the curse of dimensionality, leading to significantly improved prediction accuracy compared to classical kernel weights.  The two methods certainly extend random forests to the case with metric space valued responses. In addition, our theoretical findings include the most up-to-date result for random forests with Euclidean responses as a special case. Our proposals are supported by strong numerical performance, as demonstrated in both simulation studies and real data applications. In the present work, we focus on the theory of random forest weighted local constant Fr\'echet regression estimator. Theoretical investigation into the random forest weighted local linear Fr\'echet regression estimator including convergence rate and asymptotic normality is challenging for future research. 
	
	For Fr\'echet regression, our two methods only use the most basic information of a metric space. So our methods have a wide range of applicability. When the metric space is a specific Riemannian manifold, more information can be considered in the construction of the model. Taylor expansions can be implemented on the tangent plane of the Riemannian manifold based on some specific geometric structure. And some advanced statistical tools designed for responses lying on the Riemannian manifold were developed, like the intrinsic local polynomial regression \citep{yuan2012local} and the manifold additive model \citep{lin2022additive}. For metric space being a specific Hilbert space, vector operations and an inner product structure are available, which inspires several promising nonparametric Hilbertian regressions such as \cite{jeon2020additive} and \cite{jeon2022locally}. For the above two types of responses, we can consider a nonparametric regression framework based on the random forest kernel in the future. When more information of the output space is considered, models are expected to be more specific and targeted.

	
	\acks{
		
		We sincerely thank the action editor and two reviewers for their valuable comments and constructive suggestions. The research of Rui Qiu and the corresponding author Zhou Yu is supported by the National Key R\&D Program of China (Grant No. 2021YFA1000100 and 2021YFA1000101), the National Natural Science Foundation of China (Grant No. 12371289) and the Shanghai Pilot Program for Basic Research (Grant No. TQ20220105). The research of Ruoqing Zhu is supported by NSF grant 2210657.}
	

	\appendix
	\section{More Explanation of Some Concepts}\label{sec A}
	This section serves to provide a more detailed explanation of certain concepts mentioned in the main text.
	\subsection{Random Forest Kernel}
	Nadaraya-Watson Fr\'echet regression \citep{hein2009robust} and local Fr\'echet regression \citep{petersen2019frechet} are based on the rationality that $(X, Y)$ should be informative for $m_{\oplus}(x)$ if $X$ is close to $x$ (assume the function $m_{\oplus}$ has some degree of smoothness). The smoothing kernel $K_h(X_i-x)$ is exactly used to weight the contribution of each $(X_i, Y_i)$ to the estimation of $m_{\oplus}(x)$ according to the proximity of $X_i$ to $x$. But if the predictor contains some irrelevant variables, using the classical kernel smoothing functions often has unsatisfactory performance. 
	
	Different from the above kernel, the random forest kernel has a different mechanism for generating local weights. Fr\'echet trees produce local relationships among samples by recursively partitioning the input space. In addition to helping combat the curse of dimensionality, the random forest kernel can be adaptive if the partition process makes use of the information from responses $Y$, for example,  the variance reduction splitting criterion introduced above.  For the sample points divided into the same child node (or leaf), it is required that not only the distance of $X$ is close to each other, but also the sample Fr\'echet variance of $Y$ is small. It encourages Fr\'echet trees to select more relevant variables to divide the sample space. So the contribution value of $(X_i, Y_i)$ given by the random forest kernel $\alpha_{i}(x)$ is jointly determined by both the information of $X_i$ and $Y_i$. The random forest kernel prefers to assign a high weight to sample points that share a similar value of the response.
	
	Assuming each tree is trained with at most $k$ sample points per leaf node, and the full training data is used for each tree, \cite{lin2006random} introduced a paradigm to understand Euclidean random forests by considering any sample points falling into the leaf $L(x)$ be a $k$ potential nearest neighbor ($k$-PNN) of $x$.  When the splitting scheme of trees depends on the response, $k$-PNNs are chosen by an adaptive selection scheme. And a $k$-potential nearest neighbor can be made a $k$ nearest neighbor by choosing a reasonable distance metric but not a simple Euclidean distance. The adaptive nature of the random forest kernel can be reflected in this way. A similar analysis can be extended to non-Euclidean cases.

	\subsection{Honest Tree}
	The honesty assumption is the largest divergence between the theory and applications of random forests. However, it is necessary for pointwise asymptotic theoretical analysis as it can help to eliminate bias. Similar assumptions have been used in many literatures \citep{friedberg2020local, bloniarz2016supervised, denil2013consistency, biau2012analysis}. The training examples whose $Y_{i}$’s are used for prediction are called prediction points, while the training examples whose $Y_{i}$'s are used to construct the tree are called structure points. A random forest is honest if it is composed of honest trees. The advantage of such random forests is that the model construction process and prediction process are independent. This brings great convenience to the analysis of the theoretical property of random forests.  \cite{wager2018estimation} achieved both consistency and the central limit theorem of honest random forests provided that the honesty assumption guarantees the following critical relationship for the bias analysis
	$$
	E\big\{T_{b}(x ; \mathcal{D}_{n}^{b}, \xi_{b})\big\}=E\big[E\big\{Y \mid X \in L_{b}(x; \mathcal{D}_{n}^{b}, \xi_{b})\big\}\big],
	$$
	where $T_{b}(x ; \mathcal{D}_{n}^{b}, \xi_{b})$ is the prediction at $x$ of the tree $T_{b}$ constructed by a subsample $\mathcal{D}_{n}^{b}$ and a random draw $\xi_{b} \sim \Xi$, and  $L_{b}(x; \mathcal{D}_{n}^{b}, \xi_{b})$ is the corresponding leaf node containing $x$ of $T_{b}$.
	Moreover, their theoretical results can apply to a wide range of random forest algorithms, including the classical variance reduction splitting criterion. 
	
	Any method of constructing an honest tree can be applied exactly to the Fr\'echet trees. The simplest way to achieve honesty is that the splitting rule of trees only depends on the predictor $X$ like purely random forests \citep{arlot2014analysis, genuer2012variance}. When treating the random forest as a local weighting generator, all training examples in $L_{b}(x)$ can be used to calculate the random forest kernel. If the information of $Y$ is also considered, then the double-sample tree (outlined in Procedure 1 of \cite{wager2018estimation}) is a common approach to generate an honest tree. It divides the training sample into two non-overlapping parts: the structure set $\mathcal{J}_{b}$ and the prediction set $\mathcal{I}_{b}$ satisfying $|\mathcal{J}_{b}|=\lceil s_{n} / 2\rceil$ and $|\mathcal{I}_{b}|=\lfloor s_{n} / 2\rfloor$. During the tree growing process, the splits are chosen using any data from the sample $\mathcal{J}_{b}$  and $X$-observations from the sample $\mathcal{I}_{b}$, but without using $Y $-observations from the sample $\mathcal{I}_{b}$. And the estimation of leaf-wise responses only adopts $Y$-observations from the sample $\mathcal{I}_{b}$. When treating the random forest as a local weighting generator, we only consider the subset $\{(X_{i}, Y_{i}): (X_{i}, Y_{i}) \in L_{b}(x)\}$ of $\mathcal{I}_{b}$  to calculate the random forest kernel;  For more discussion, please refer to section 2.4 and Appendix B of \cite{wager2018estimation} or section 2.3 of \cite{friedberg2020local}.  It is worth noting that if honesty is achieved by double-sample trees, another condition $\alpha$-regular mentioned in the paper should be satisfied for the sample $\mathcal{I}_{b}$. This is why $X$-observations from the sample $\mathcal{I}_{b}$ may be used during the construction of trees.

	\subsection{A Little Remark about RFWLLFR}
	Consider a special case with $p=1$, then RFWLLFR estimator $\hat{l}_{\oplus}(x)$ has the following equivalent expression.
	\begin{align*}
		\hat{l}_{\oplus}(x)&=\underset{y \in \Omega}{\argmin}\sum_{i=1}^{n}e_{1}^{\T} (\tilde{X}^{\T}A\tilde{X})^{-1}\left(\begin{array}{c}
			1 \\\nonumber
			X_{i}-x
		\end{array}\right){\alpha}_{i}(x) d^2(Y_{i},y)\\
		&=\underset{y \in \Omega}{\argmin}\sum_{i=1}^{n}e_{1}^{\T}\left(\begin{array}{cc}\sum_{i=1}^{n}{\alpha}_{i}(x) & \sum_{i=1}^{n}{\alpha}_{i}(x)(X_{i}-x)  \\ \sum_{i=1}^{n}{\alpha}_{i}(x)(X_{i}-x) & \sum_{i=1}^{n}{\alpha}_{i}(x)(X_{i}-x)^2 \end{array}\right)^{-1} \\
	    &\quad \quad \quad \quad \quad \quad \quad \quad \quad \quad \quad \quad \quad \quad \quad \quad \quad \quad  \quad \quad  \quad \left(\begin{array}{c}
			1 \\
			X_{i}-x
		\end{array}\right){\alpha}_{i}(x) d^2(Y_{i},y)\\
		&=\underset{y \in \Omega}{\argmin}\frac{1}{n}\sum_{i=1}^{n}e_{1}^{\T}\left(\begin{array}{cc}\hat{\mu}_{0} & \hat{\mu}_{1}  \\  \hat{\mu}_{1} &  \hat{\mu}_{2} \end{array}\right)^{-1}\left(\begin{array}{c}
			1 \\\nonumber
			X_{i}-x
		\end{array}\right) {\alpha}_{i}(x) d^2(Y_{i},y)\\
		&=\underset{y \in \Omega}{\argmin}\frac{1}{n}\sum_{i=1}^{n}e_{1}^{\T}\frac{1}{\hat{\mu}_{0}\hat{\mu}_{2}-\hat{\mu}_{1}^2}\left(\begin{array}{cc}\hat{\mu}_{2} & -\hat{\mu}_{1}  \\  -\hat{\mu}_{1} &  \hat{\mu}_{0} \end{array}\right)\left(\begin{array}{c}
			1 \\
			X_{i}-x
		\end{array}\right) {\alpha}_{i}(x) d^2(Y_{i},y)\\
		&=\underset{y \in \Omega}{\argmin}\frac{1}{n}\sum_{i=1}^{n}\frac{1}{\hat{\mu}_{0}\hat{\mu}_{2}-\hat{\mu}_{1}^2}\left\{\hat{\mu}_{2}-\hat{\mu}_{1}\left(X_{i}-x\right)\right\}{\alpha}_{i}(x) d^2(Y_{i},y)\\
		&=\underset{y \in \Omega}{\argmin}\frac{1}{n}\sum_{i=1}^{n}t_{in}(x)d^2(Y_{i},y),
	\end{align*}
	where  $t_{in}(x)=\frac{1}{\hat{\mu}_{0}\hat{\mu}_{2}-\hat{\mu}_{1}^2}{\alpha}_{i}(x)\left\{\hat{\mu}_{2}-\hat{\mu}_{1}\left(X_{i}-x\right)\right\}$, $\hat{\mu}_{j}=\frac{1}{n}\sum_{i=1}^{n}{\alpha}_{i}(x)\left(X_{i}-x\right)^{j}$.
	Apart from the weight generating function, this form coincides with the local  Fr\'echet regression proposed by \cite{petersen2019frechet}. Acquiring local Fr\'echet regression estimators directly from corresponding explicit Euclidean forms as we do may be more straightforward.

	\section{Asymptotic Normality}\label{sec E}
	This section is dedicated to deriving the asymptotic normality of RFWLCFR. We first generalize the theory of $M_m$-estimator in \cite{bose2018u} to the case that $m$ tends to infinity along with $n$.
	\begin{definition}
		Let $Z_{1}, Z_{2}, \ldots, Z_{m_n}$ be i.i.d. $\mathcal{Z}$-valued random variables and $\theta \in \mathcal{R}^q $. A real-valued measurable function $f_n(z_1, z_2, \ldots,z_{m_n}, \theta)$  is symmetric
		in the arguments $z_1, z_2, \ldots,z_{m_n}$ for each $n$. Define
		$$
		Q_{n}(\theta)=E f_n\left(Z_{1}, Z_{2}, \ldots, Z_{m_n}, \theta\right)
		$$
		and
		$$
		\theta_{n}=\underset{\theta \in \mathcal{R}^q}{\argmin} Q_{n}(\theta).
		$$
		$\theta_{n}$ is called the $M_{m_n}$-parameter.
	\end{definition}
	
	\begin{definition}
		Let $Z_{1}, Z_{2}, \ldots, Z_{n}$ be a sequence of i.i.d. observations. Define
		$$
		\hat{Q}_{n}(\theta)=\binom{n}{m_n}^{-1} \sum_{1 \leq i_{1}<i_{2} <\ldots<i_{m_n} \leq n} f_n\left(Z_{i_{1}}, Z_{i_{2}}, \ldots, Z_{i_{m_n}}, \theta\right)
		$$
		and
		$$
		\hat{\theta}_{n}=\underset{\theta \in \mathcal{R}^q}{\argmin} \hat{Q}_{n}(\theta).
		$$
		$\hat{\theta}_{n}$ is called the $M_{m_n}$-estimator of $\theta_{n}$.
	\end{definition}
	
	In the above definition, a hidden assumption is $m_n/n \rightarrow 0$. Actually, $\hat{Q}_{n}(\theta)$ is an infinite order U-process about $\theta$. Since $\hat{Q}_{n}(\theta)$ is the sample analogue of $Q_{n}(\theta)$,  $\hat{\theta}_{n}$ is  a reasonable estimator of $\theta_{n}$. When $m_n=1$, $\hat{\theta}_{n}$ is the classical M-estimator. When  $m_n=m$ is a fixed positive integer, $\hat{\theta}_{n}$ is the $M_m$-estimator studied in \cite{bose2018u}.  By an appropriate
	selection theorem,  it is often possible to choose a measurable version of $\hat{\theta}_{n}$. We always work with such a version. Please refer to section $2.3$ of  \cite{bose2018u} for more details.

	Let $g_n$ be  a measurable sub-gradient of $f_n(z_1, z_2, \ldots,z_{m_n}, \theta)$ about $\theta$. Define 
	$$
	K_n=\Var\left[E\left\{g_n\left(Z_{1}, Z_{2},   \ldots Z_{m_n}, \theta_{n}\right) \mid Z_{1}\right\}\right],
	$$
	$$
	U_{n} =\binom{n}{m_n}^{-1} \sum_{1 \leq i_{1} < i_{2} < \ldots<i_{m_n} \leq n} g_n\left(Z_{i_{1}}, Z_{i_{2}}, \ldots, Z_{i_{m_n}}, \theta_{n}\right).
	$$
	In order to achieve the asymptotic normality of the $M_{m_n}$-estimator, we need the following assumptions.
	
	(i) $f_n\left(z_{1}, z_{2}, \ldots, z_{m_n}, \theta\right)$ is measurable in $\left(z_{1}, z_{2}, \ldots, z_{m_n}\right)$ and convex  in $\theta$.
	
	(ii) $Q_n(\theta)$ is finite for each $\theta$.
	
	(iii) $\theta_{n}$ exists and is unique, and  $f_n\left(z_{1}, z_{2}, \ldots, z_{m_n}, \theta\right)$ is twice differentiable on an appropriate neighborhood of $\theta_{n}$.
	
	(iv) $E\left|g_n\left(Z_{1}, Z_{2}, \ldots, Z_{m_n}, \theta_n\right)\right|^{2}<C$ for some constant $C$, and $m_n \lambda_{\min}(K_n)\nrightarrow 0$, where $\lambda_{\min}(K_n) $ denotes the smallest eigenvalue of $K_n$.
	
	(v) $H_n=\nabla^{2} Q\left(\theta_{n}\right)$ exists and is positive definite and $\lambda_{\min }(H_n)\nrightarrow 0$.

	Through the definition, we know that the $M_{m_n}$-estimator is an implicit solution to the infinite order U-process. Under the above assumptions, we can derive $\hat{\theta}_n$ a weak representation through the linearization of the infinite order U-statistic $U_n$.  Therefore, the asymptotic normality of infinite order U-statistics determines the asymptotic normality of the $M_{m_n}$-estimator $\hat{\theta}_n$. \cite{mentch2016quantifying} gave sufficient conditions for the asymptotic normality of such U-statistics. However, these conditions can not hold simultaneously. \cite{diciccio2022clt}  then developed conditions that can be verified on the basis of \cite{mentch2016quantifying}. But both of their results require that the order of the infinite order U-statistics is $o(\sqrt{n})$. \cite{peng2019asymptotic} further improved this result when the order of the infinite order U-statistics is $o(n)$. \cite{wager2018estimation} also gave the same rate (ignoring the $\log$-factors) when focusing on random forests with some additional requirements on the construction of trees. Our assumption (iv) here is to ensure the asymptotic normality of $U_n$ by the result of \cite{peng2019asymptotic}. And assumptions (i), (ii), (iii) and (v) are adaptions of that for studying $M_{m}$ estimator in \cite{bose2018u}.
	
	\begin{theorem}\label{Mm normal}
		Suppose that assumptions (i)-(v) hold, then for any sequence of measurable minimizers $\{\hat{\theta}_{n}, n\ge 1\}$,
		
		(a) $\hat{\theta}_{n}-\theta_{n}=-H_n^{-1} U_{n}+o_p\left(\frac{\sqrt{m_{n}}}{\sqrt{n}}\right)$,
		
		(b) $\sqrt{n}\Lambda_n^{-1/2}\left(\hat{\theta}_{n}-\theta_{n}\right) \stackrel{d}{\longrightarrow} \mathcal{N}\left(0, I\right)$, where
		$$
		\Lambda_n=m_n^2H_n^{-1}K_nH_n^{-1}.
		$$
	\end{theorem}
	
	Theorem \ref{Mm normal} establishes the asymptotic normality of $M_{m_n}$-estimator, which is a generalization of the central limit theorem of the $M_{m}$-estimator given in \cite{bose2018u}. Next, we derive the asymptotic normality of RFWLCFR by applying this result. Here we continue to adopt the expressions (\ref{RFWLCFR-2}) and (\ref{P-RFWLCFR-2}). Let
	$$h_n\left(Z_{i_{k,1}}, Z_{i_{k,2}}, \ldots, Z_{i_{k,s_n}}, y\right)= E_{\xi \sim \Xi}\Bigg\{\frac{1}{N(L(x;\mathcal{D}_n^k,\xi))} \sum_{i: X_{i} \in L(x;\mathcal{D}_n^k,\xi)}d^2\left(Y_{i},y\right)\Bigg\}
	$$
	where $Z_i=(X_i,Y_i)$ and $\mathcal{D}_n^k=\left(Z_{i_{k,1}}, Z_{i_{k,2}}, \ldots, Z_{i_{k,s_n}}\right)$. 
	Then
	$$
	\hat{r}_{\oplus}(x)=\underset{y \in \Omega}{\argmin}\hat{R}_{n}(x,y)=\underset{y \in \Omega}{\argmin}\binom{n}{s_n}^{-1} \sum_{1 \leq i_{1}<i_{2} < \ldots<i_{s_n} \leq n} h_n\left(Z_{i_{1}}, Z_{i_{2}}, \ldots, Z_{i_{s_n}}, y\right),
	$$
	$$
	\tilde{r}_{\oplus}(x)=\underset{y \in \Omega}{\argmin}\tilde{R}_{n}(x,y)=\underset{y \in \Omega}{\argmin}E h_n\left(Z_1, Z_2, \ldots, Z_{s_n}, y\right).
	$$

	Unfortunately, since $y$ is not in the Euclidean space,  the
	derivative about $y$  can't be computed and $\hat{r}_{\oplus}(x)-\tilde{r}_{\oplus}(x)$ has no sense. In order to apply the Theorem~\ref{Mm normal}, we
	consider mapping $y$ to the Euclidean space locally and establish asymptotic normality, which is the standard procedure for the asymptotic analysis of sample Fr\'echet mean as introduced in \cite{bhattacharya2017omnibus,bhattacharya2003large,bhattacharya2005large}. We assume the following conditions to establish the central limit theorem of the proposed RFWLCFR.
	
	(A13) $h_n\left(z_{1}, z_{2}, \ldots, z_{s_n}, y\right)$ is measurable in $(z_{1}, z_{2}, \ldots, z_{s_n})$ and $\tilde{R}_{n}(x,y)<\infty$ for each $y$.
	
	(A14) $\tilde{r}_{\oplus}(x)$ exists and is unique.
	
	(A15) $\tilde{r}_{\oplus}(x) \in G$ for large $n$, where $G$ is a measurable subset of $\Omega$.  And there is a homeomorphism $\phi:G \rightarrow U$, where $U$ is an open subset of $\mathcal{R}^{q}$ for some $q \geq 1$, and $G$ is given its relative topology on $\Omega$. Also
	$$
	u \mapsto f_n(z_{1}, z_{2}, \ldots, z_{s_n},u)=h_n\left(z_{1}, z_{2}, \ldots, z_{s_n}, \phi ^{-1}(u)\right)
	$$
	is twice differentiable on an appropriate neighborhood of $\phi(\tilde{r}_{\oplus}(x))$.
	
	(A16) $f_n(z_{1}, z_{2}, \ldots, z_{s_n},u)$ is convex in $u$.
	
	(A17) Let $g_n$ be a measurable sub-gradient of $f_n$ about $u$. Define
	$$
	K_n=\Var\left\{E\left[g_n\left(Z_{1}, Z_{2}, \ldots Z_{s_n}, \phi(\tilde{r}_{\oplus}(x))\right) \mid Z_{1}\right]\right\},
	$$
	then $E\left|g_n\left(Z_{1}, Z_{2},\ldots, Z_{s_n},  \phi(\tilde{r}_{\oplus}(x))\right)\right|^{2}<C$ for some constant $C$,  and $s_n  \lambda_{\min}(K_n)\nrightarrow 0$.
	
	(A18) $H_n=\nabla^{2}Ef_n(Z_{1}, Z_{2}, \ldots, Z_{s_n},\phi(\tilde{r}_{\oplus}(x)))$
	exits and is positive definite, and $\lambda_{\min }(H_n)\nrightarrow 0$.
	
	The assumption (A15) is crucial just like the assumption for establishing the asymptotic normality of sample Fr\'echet mean as suggested in \cite{bhattacharya2017omnibus}. {For example, when $(\Omega, d_g)$ is a $q$-dimensional complete Riemannian manifold with metric tensor $g$ and geodesic distance $d_g$, we can choose the  Riemannian logarithmic map at $\tilde{r}_{\oplus}(x)$ as the homeomorphism $\phi$, which is defined on a neighborhood of $\tilde{r}_{\oplus}(x)$ onto its image $U$ in the tangent space at $\tilde{r}_{\oplus}(x)$.} Other assumptions are adaptions of that of Theorem~\ref{Mm normal}.
	
	\begin{theorem} \label{asymptotic normal}
		Suppose that for a fixed $x \in [0,1]^p$, (A1), (A4), (A13)--(A18) hold, and the Fr\'echet trees are symmetric. Then $\phi(\hat{r}_{\oplus}(x))$ is asymptotically normal, i.e.,
		$$\sqrt{n}\Lambda_n^{-1/2}\left\{\phi(\hat{r}_{\oplus}(x))-\phi(\tilde{r}_{\oplus}(x))\right\} \rightarrow \mathcal{N} \left(0, I\right),$$
		where
		$\Lambda_n=s_n^2H_n^{-1}K_nH_n^{-1}$ with $K_n, H_n$ defined in assumption (A17) and (A18).
	\end{theorem}

	{If the  homeomorphism $\phi$ in Theorem \ref{asymptotic normal} is further  Lipschitz-continuous, by Lemma~\ref{variance rate},
		$$\lVert  \phi(\tilde{r}_{\oplus} (x))-\phi(m_{\oplus}(x))\rVert \leq L d(\tilde{r}_{\oplus} (x),m_{\oplus}(x))=O\left(s_n^{-\frac{1}{2} \frac{\log \left(1-\alpha\right)}{\log (\alpha)} \frac{\pi}{p} \frac{1}{\beta_1-1}}\right),$$
		where $L$ is the Lipschitz constant. Under a suitable range of orders for $s_n$ with respect to $n$, 
		$
		\lVert  \phi(\tilde{r}_{\oplus} (x))-\phi(m_{\oplus}(x))\rVert / \lVert \Lambda_n^{1/2} / \sqrt{n} \lVert
		$
		can converge to zero. Then we can get the asymptotic normality of $\phi(\hat{r}_{\oplus}(x))$ about $\phi(m_{\oplus}(x))$ by Theorem~\ref{asymptotic normal} and Slutsky's theorem. The following takes the Euclidean case as a simple example to illustrate it.}
	
	\begin{remark}\label{remark 2}
		Consider the special case when $\Omega \subseteq \mathcal{R}$. For responses that are Euclidean, we naturally choose $\phi$ as the identity mapping. Then we have
		\begin{align*}
			f_n(Z_{i_{k,1}}, \ldots, Z_{i_{k,s_n}}, u)&=h_n\left(Z_{i_{k,1}}, \ldots, Z_{i_{k,s_n}}, u\right)\\
			&= E_{\xi \sim \Xi}\Bigg\{\frac{1}{N(L(x;\mathcal{D}_n^k,\xi))} \sum_{i: X_{i} \in L(x;\mathcal{D}_n^k,\xi)}\left(Y_{i}-u\right)^2\Bigg\}.
		\end{align*}
		And we further get
		$$Q_n(u)=E\Bigg\{\frac{1}{N(L(x;\mathcal{D}_n^k,\xi))} \sum_{i: X_{i} \in L(x;\mathcal{D}_n^k,\xi)}\left(Y_{i}-u\right)^2\Bigg\},$$
		$$\hat{Q}_n(u)=\binom{n}{s_n}^{-1} \sum_{k} E_{\xi \sim \Xi}\Bigg\{\frac{1}{N(L(x;\mathcal{D}_n^k,\xi))} \sum_{i: X_{i} \in L(x;\mathcal{D}_n^k,\xi)}\left(Y_{i}-u\right)^2\Bigg\}.$$
		In addition,
		\begin{gather*}
			u_n=\underset{u \in U}{\argmin}  Q_n(u)=E\Bigg\{\frac{1}{N(L(x;\mathcal{D}_n^k,\xi))} \sum_{i: X_{i} \in L(x;\mathcal{D}_n^k,\xi)}Y_{i}\Bigg\},\\
			\hat{u}_n=\underset{u \in U}{\argmin}  \hat{Q}_n(u)=\binom{n}{s_n}^{-1} \sum_k E_{\xi \sim \Xi}\Bigg\{\frac{1}{N(L(x;\mathcal{D}_n^k,\xi))} \sum_{i: X_{i} \in L(x;\mathcal{D}_n^k,\xi)}Y_{i}\Bigg\}.
		\end{gather*}
		Since $g_n$ is the sub-gradient of $f_n$ about $u$, we have
		$$
		g_n(Z_{i_{k,1}}, \ldots, Z_{i_{k,s_n}}, u)=-2 \left[E_{\xi \sim \Xi}\Bigg\{\frac{1}{N(L(x;\mathcal{D}_n^k,\xi))} \sum_{i: X_{i} \in L(x;\mathcal{D}_n^k,\xi)}Y_{i}\Bigg\}-u \right].
		$$
		Let $\zeta_{1,n}=\Var \left[ E \left\{E_{\xi \sim \Xi}\left(\frac{1}{N(L(x;\mathcal{D}_n^{*},\xi))} \sum_{i: X_{i} \in L(x;\mathcal{D}_n^{*},\xi)}Y_{i}\right) \mid Z_{1}\right\}\right]$ with $\mathcal{D}_n^{*}=(Z_1,\dots,Z_{s_n})$,
		then $K_n=4 \zeta_{1,n}$ and $\Lambda_n=s_n^2H_n^{-1}K_nH_n^{-1}=s_n^2  \zeta_{1,n} $. If $s_n  \zeta_{1,n}\nrightarrow 0$ and the assumption (A17) holds, by Theorem~\ref{asymptotic normal} we have
		$$\frac{\sqrt{n}\left(\hat{u}_{n}-u_{n}\right)}{\sqrt{s^2_n  \zeta_{1,n}}} \stackrel{d}{\longrightarrow} \mathcal{N}\left(0, 1\right),$$
		\emph{i.e.},
		\begin{align}\label{European random forest}
			\frac{\sqrt{n}\left\{\hat{r}_{\oplus}(x)-\tilde{r}_{\oplus}(x)\right\}}{\sqrt{s^2_n  \zeta_{1,n}}} \stackrel{d}{\longrightarrow} \mathcal{N}\left(0, 1\right).
		\end{align}
		Here $\hat{r}_{\oplus}(x)$ is exactly the prediction of Euclidean random forest at $x$, and \eqref {European random forest} is the standard results of Euclidean random forests \citep{mentch2016quantifying, peng2019asymptotic, diciccio2022clt}. Since $s_n \zeta_{1,n}$ is bounded and $s_n  \zeta_{1,n}\nrightarrow 0$, it holds that $\hat{r}_{\oplus}(x)-\tilde{r}_{\oplus}(x)=O_{p}\big((s_n/n)^{1/2}\big)$.
		
		For the Euclidean case, $\beta_1=2$ and  Lemma~\ref{bias rate} gives
		$$\left| \tilde{r}_{\oplus} (x)-m_{\oplus}(x) \right|=O\left(s_n^{-\frac{1}{2} \frac{\log \left(1-\alpha\right)}{\log (\alpha)} \frac{\pi}{p} }\right). $$
		Let $s_n=n^{\beta}$, then
		$$
		\frac{\left| \tilde{r}_{\oplus} (x)-m_{\oplus}(x) \right|}{\sqrt{s^2_n  \zeta_{1,n}/n}} = O\left(n^{\frac{1}{2}\left[1-\beta\left\{1+\frac{\log \left(1-\alpha\right)}{\pi^{-1} p \log \left(\alpha\right)}\right\}\right]}\right).
		$$
		The right-hand-side converges to zero provided that
		$$
		\beta>\left\{1+\frac{\log \left(1-\alpha\right)}{\pi^{-1} p \log \left(\alpha\right)}\right\}^{-1}=1-\left\{1+\frac{p}{\pi} \frac{\log \left(\alpha\right)}{\log \left(1-\alpha\right)}\right\}^{-1}.
		$$
		Then by \eqref{European random forest} and Slutsky's theorem,
		$$\frac{\sqrt{n}\left\{\hat{r}_{\oplus}(x)-m_{\oplus}(x)\right\}}{\sqrt{s^2_n  \zeta_{1,n}}} \stackrel{d}{\longrightarrow} \mathcal{N}\left(0, 1\right)$$
		when 
		$$s_{n} \asymp n^{\beta} \quad \text{for some} \quad \beta_{\min }:=1-\left\{1+\frac{p}{\pi} \frac{\log \left(\alpha\right)}{\log \left(1-\alpha\right)}\right\}^{-1}<\beta<1.
		$$ 
		This result coincides with Theorem 1 of \cite{wager2018estimation}. Therefore, our asymptotic normality established for Fr\'echet regression with metric space valued responses includes their result for Euclidean random forests as a special case.
	\end{remark}
	
	\section{Additional Simulations}\label{sec B}
	Here we specifically introduce the simplified splitting criterion used by the Fr\'echet tree construction in our simulations, and add more simulations for responses being distributions or symmetric positive-definite matrices.
	\subsection{Simplified Adaptive Splitting Criterion for Fr\'echet Trees}
	The process introduced in Section~\ref{Frechet Trees} to find the optimal split is accurate but computationally intensive. In all simulation experiments of this paper, we adopt another efficient way introduced by \cite{capitaine2019fr}. A split on an internal node $A$ along the direction of feature $j$ is any couple of distinct elements $(c_{j,l}, c_{j,r})$. The partition associated
	with elements $(c_{j,l}, c_{j,r})$ is defined by $$A_{j,l}=\left\{x \in A: |x^{(j)}-c_{j,l}| \leq |x^{(j)}-c_{j,r}|\right\}$$ and $$A_{j,r}=\left\{x \in A: |x^{(j)}-c_{j,r}|<|x^{(j)}-c_{j,l}|\right\},$$ which generate the left and right child nodes of the node $A$. Again, let
	\begin{align*}
		\mathcal{H}_{n}(j)=\frac{1}{N_{n}(A)}\left\{\sum_{i: X_{i} \in A} d^{2}\left(Y_{i}, \bar{Y}_{A}\right)-\sum_{i: X_{i} \in A_{j,l}} d^{2}\left(Y_{i}, \bar{Y}_{A_{j,l}}\right)-\sum_{i: X_{i} \in A_{j,r}} d^{2}\left(Y_{i}, \bar{Y}_{A_{j,r}}\right)\right\}.
	\end{align*}
	Then the optimal split $(c_{j_{n}^{*},l}, c_{j_{n}^{*},r})$ is decided by
	\begin{align*}
		j_{n}^{*}=\underset{j}{\argmax}\ \mathcal{H}_{n}(j).
	\end{align*}
	To determinate the representatives $(c_{j,l}, c_{j,r})$, the $2$-means algorithm ($k$-means with $k$ = 2) can be implemented on the $j$th component of the sample points falling into the node $A$.

	\subsection{Fr\'echet Regression for Distributions}\label{sec B.2}
\begin{figure}[t!]
		\centering
		\includegraphics[width=0.6\linewidth]{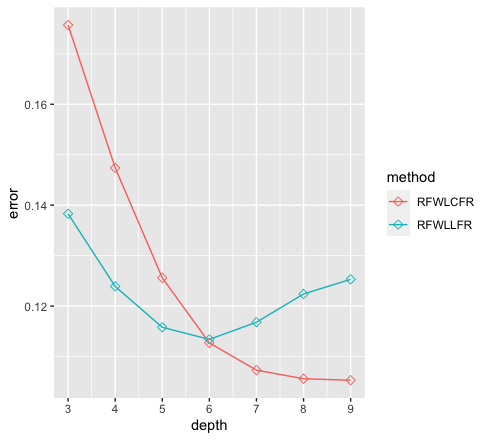}
		\caption{The influence of depth of Fr\'echet trees on average MSE for (10,500) from setting I-2.}
		\label{deep}
	\end{figure}

	RFWLCFR is similar in nature to random forests, so it prefers using deeper Fr\'echet trees. As for RFWLLFR, knowing that a more powerful local linear regression will be used for the final model fitting, it is not reasonable to capture too much signal from the data during the construction of the Fr\'echet trees. So shallower trees are often used to avoid overfitting for RFWLLFR. Figure~\ref{deep} shows the effect of the depth of  Fr\'echet trees on the performance of our two methods based on setting I-2 with $p=10$ and $n=500$. RFWLLFR achieves the optimal performance when the depth is six, while RFWLCFR prefers deeper Fr\'echet trees.
	
	We also select several combinations of $(n, p)$ from setting I-2 to study the effect of noise size $\sigma$ on the performance of each method. The results are summarized in Table~\ref{table:dis_sigma}. It can be seen that GFR is almost unaffected by noise. LFR performs poorly when the noise level is high. And our proposed RFWLCFR and RFWLLFR are still better than GFR and LFR in general.
	
	\begin{table}[t!]
		\centering
		\resizebox{0.83\columnwidth}{!}{
			\begin{tabular}{ccccccccccccccc}
				\toprule
				$(p,n)$ & $\sigma$  & GFR & LFR & RFWLCFR/FRF & RFWLLFR\\
				\midrule
				\multirow{3}{*}{(2,200)}
				&$\sigma=0.1$ & 0.3026   (0.0283) & 0.0261   (0.0198) & 0.0158   (0.0036) & \textbf{0.0084}    (0.0041)  \\
				&$\sigma=0.2$ &  0.3023  (0.0278) & 0.0745  (0.1550) & 0.0254  (0.0050) & \textbf{0.0186} (0.0073)  \\
				&$\sigma=0.5$ & 0.3032    (0.0274) & 0.3341  (0.3412) & 0.0895  (0.0181) & \textbf{0.0786}  (0.0209)  \\
				\midrule
				\multirow{3}{*}{(5,500)}
				&$\sigma=0.1$ & 0.2331  (0.0240) & NA &0.0515  (0.0080) & \textbf{0.0437}   (0.0074)  \\
				&$\sigma=0.2$ & 0.2335 (0.0241) & NA  &0.0557 (0.0087)  & \textbf{0.0502} (0.0083)  \\
				&$\sigma=0.5$ & 0.2363  (0.0245) & NA & \textbf{0.0850}  (0.0128) & 0.0946 (0.0144)  \\
				\midrule
				\multirow{3}{*}{(10,1000)}
				&$\sigma=0.1$ & 0.2434  (0.0295) & NA &  \textbf{0.0870} (0.0175) & 0.0879 (0.0145)  \\
				&$\sigma=0.2$ & 0.2438  (0.0297) & NA & \textbf{0.0901} (0.0174) & 0.0927 (0.0148)  \\
				&$\sigma=0.5$ & 0.2462   (0.0302) & NA & \textbf{0.1103}  (0.0195) & 0.1251 (0.0171)  \\
				\midrule
				\multirow{3}{*}{(20,2000)}
				&$\sigma=0.1$ & 0.2452    (0.0285) & NA & \textbf{0.1227}  (0.0225) & 0.1300 (0.0191)  \\
				&$\sigma=0.2$ & 0.2456 (0.0286) & NA & \textbf{0.1257} (0.0238) & 0.1337 (0.0194) \\
				&$\sigma=0.5$ & 0.2479  (0.0288) & NA & \textbf{0.1401}   (0.0260) & 0.1654 (0.0235)  \\
				\bottomrule
			\end{tabular}
		}
	\caption{Average MSE (standard deviation) of different methods for (2,200), (5,500), (10,1000), (20,2000) from setting I-2 with different $\sigma$ over 100 simulation runs. Bold-faced numbers indicate the best performers.}
	\label{table:dis_sigma}
	\end{table}

    	\begin{table}[t!]
    	\centering
    	\resizebox{0.83\columnwidth}{!}{
    		\begin{tabular}{ccccccccccccccc}
    			\toprule
    			Model & $(p,n)$  & GFR & LFR & RFWLCFR/FRF & RFWLLFR\\
    			\midrule
    			\multirow{8}{*}{I-3}
    			&$(2, 100)$ & 0.2495  (0.2002) & 0.0869  (0.3530) & 0.1179   (0.1745) & \textbf{0.0423}  (0.0551)  \\
    			&$(2, 200)$ & 0.2302  (0.3131) & 0.0390  (0.1558) & 0.0944 (0.2843) & \textbf{0.0347}  (0.1355)  \\
    			&$(5, 200)$ & 0.0893  (0.0641) & NA & 0.0429 (0.0432) & \textbf{0.0342} (0.0261)  \\
    			&$(5, 500)$ & 0.0827  (0.0551) & NA & 0.0248  (0.0272) & \textbf{0.0164}  (0.0107)  \\
    			&$(10, 500)$ & 0.0896  (0.1006) & NA &0.0368  (0.0660) &\textbf{0.0307} (0.0293)  \\
    			&$(10, 1000)$ & 0.0810 (0.0888) & NA &0.0226  (0.0623) & \textbf{0.0187}  (0.0201)  \\
    			&$(20, 1000)$ & 0.0445 (0.0197) & NA & \textbf{0.0181}  (0.0119) & 0.0210  (0.0085)  \\
    			&$(20, 2000)$ & 0.0502  (0.0275) & NA & \textbf{0.0155} (0.0110) & 0.0172  (0.0081)  \\
    			\bottomrule
    		\end{tabular}
    	}
    	\caption{Average MSE (standard deviation) of different methods for setting I-3 over 100 simulation runs. Bold-faced numbers indicate the best performers.}
    	\label{table:cor}
    \end{table}
	
	Since the components of the predictor $X$ are independent in all previous simulation settings, we here add another setting to cover the case that components are correlated. 
	
	Setting I-3:
	We generate $X$ by a multivariate normal distribution 
	\begin{align*}
		X \sim \mathcal{N}\left(0, \Sigma\right),
	\end{align*}
	where the $ij$-th element of $\Sigma$ is $0.5^{\left| i-j \right|}$. Then $Y$ is generated by
	\begin{align*}
		Y =\mathcal{N}\left(\mu_Y, \sigma_Y^2\right),
	\end{align*}
	where    
	\begin{align*}
		\mu_{Y} \sim \mathcal{N}\left(0.1(e_1^{\T}X)^2 \left(2\beta^{\T} X-1\right), 0.2^2\right) \quad \text{and} \quad \sigma_{Y}=1.
	\end{align*}
	The above $e_i$ is a vector of zeros with $1$ in the $i$th element. Consider the following four kinds of dimensions
	
	(i) For $p=2$:	$\beta=\left(0.75, 0.25\right)$.
	
	(ii) For $p=5,10$:	$\beta=\left(0.1, 0.2, 0.3, 0.4,  0, \dots, 0\right)$.
	
	(iii) For $p=20$: $\beta=\left(0.1, 0.2, 0.3, 0.4,  0,\dots, 0, 0.1, 0.2, 0.3, 0.4\right)/2$.
	
	From the results in Table~\ref{table:cor}, we can find that the performance of GFR can not be improved significantly by simply increasing the number of training samples, but it performs better when the number of effective variables increases. RFWLLFR is the most stable among all methods. As the dimension of $X$ becomes larger, the performance of RFWLCFR  gets closer to that of RFWLLFR. Especially in the high-dimensional case, RFWLCFR begins to outperform RFWLLFR. Overall, all methods under the current setting behave similarly to the cases when the components of $X$ are independent.
	
	\subsection{Fr\'echet Regression for Symmetric Positive-definite Matrices}
	For responses being symmetric positive definite matrices, the intrinsic local polynomial regression (ILPR) \citep{yuan2012local} and the manifold additive model (MAM) \citep{lin2022additive} are two promising tools that take advantage of the geometric structure of the Riemannian manifold. We plan to include the two methods for comparisons for Fr\'echet regression with symmetric positive-definite matrices.
	
	The abelian group structure inherited from either the Log-Cholesky metric or the Log-Euclidean metric framework can turn the space of symmetric
	positive-definite matrices into a Riemannian manifold and further a bi-invariant Lie group.  \cite{lin2022additive} further proposed an additive model for the regression of symmetric positive-definite matrix valued responses called the manifold additive model (MAM). Their numerical studies show that the proposed method enjoys superior numerical performance compared with the intrinsic local polynomial regression (ILPR, \cite{yuan2012local}), especially when the underlying model is fully additive. However, \cite{lin2022additive} only considered $p=3,4$ in their simulation studies. In the next, we adopt the settings in \cite{lin2022additive} to make a comprehensive comparison among MAM, ILPR, GFR, FRF, RFWLCFR, and RFWLLFR. MAM can be implemented with the R-package ``matrix-manifold'' \citep{matrix-manifold2020}.
	
	Setting II:
	Let $X \sim \mathcal{U}([0, 1]^p)$.  The response $Y$ is generated via
	\begin{align*}
		Y=\mu \oplus w\left(X\right) \oplus \zeta,
	\end{align*}
	where $\mu$ is the $3 \times 3$ identity matrix, $w\left(X\right)=\mathfrak{e} \mathfrak{x} \mathfrak{p} \tau_{\mu, e} f\left(X\right)$, $e$ is the identity element of the group, $\tau_{\mu, e}$ denotes the parallel transport from $\mu$ to $e$,  $\mathfrak{e} \mathfrak{x} \mathfrak{p} (\cdot)$ denotes the Lie exponential map, $\oplus$ denotes the group operation, and $\zeta$ is the random noise. The noise $\zeta$ is generated according to $\mathfrak{l} \mathfrak{o} \mathfrak{g} \zeta=\sum_{j=1}^{6} Z_{j} v_{j}$, where  $\mathfrak{l} \mathfrak{o} \mathfrak{g} (\cdot)$ denotes the Lie log map, $Z_{1}, \ldots, Z_{6}$ are independently sampled from $\mathcal{N}\left(0, \sigma^{2}\right)$, and $v_{1}, \ldots, v_{6}$ are an orthonormal basis of the tangent space $T_{e} \mathcal{S}_{3}^{+}$.  The signal-to-ratio (SNR) is measured by $\text{SNR} =E\left\|\mathfrak{l o g} w\left(X\right)\right\|_{e}^{2} /E\|\mathfrak{l o g} \zeta\|_{e}^{2}$. Take the value of the parameter $\sigma^{2}$ to cover two choices for the SNR, namely, SNR $=2$ and SNR $=4$. Refer to \cite{lin2022additive} for the details of the notations and concepts here. We consider the following setting about $f(X)$.
	
	II-3: $f\left(X\right)=\sum_{k=1}^{q} f_{k}\left(x_{k}\right)$ with $f_{k}\left(x_{k}\right)$ being an $3 \times 3$ matrix whose $(j, l)$-entry is $g(x_{k} ; j, l, q)$ $=\exp (-|j-l| / q) \sin \left(2 q \pi\left\{x_{k}-(j+l) / q\right\}\right)$.
	
	II-4: $f\left(X\right)=f_{12}\left(x_{1}, x_{2}\right) \prod_{k=3}^{q} f_{k}\left(x_{k}\right)$, where $f_{12}\left(x_{1}, x_{2}\right)$ is an $3 \times 3$ matrix whose $(j, l)$-entry is $\exp \left\{-(j+l)\left(x_{1}+x_{2}\right)\right\}$, and $f_{k}\left(x_{k}\right)$ is an $3 \times 3$ matrix whose $(j, l)$-entry is $\sin \left(2 \pi x_{k}\right)$.
	
	\begin{table}[t!]
		\centering
		\resizebox{0.83\columnwidth}{!}{
			\begin{tabular}{cccccccccccccccc}
				\toprule
				Model & $(p,n)$  & MAM & ILPR & GFR & RFWLCFR/FRF & RFWLLFR\\
				\midrule
				&$(3, 100)$ & \textbf{0.415} (0.023) & 0.922 (0.126) & 0.970 (0.012) &0.714 (0.018)& 0.794 (0.023) \\
				&$(3, 200)$ & \textbf{0.299} (0.017) & 0.796 (0.064) & 0.956 (0.008)  & 0.613 (0.013)& 0.655 (0.017)\\
				&$(4, 100)$ & \textbf{0.527} (0.028) & 0.965 (0.033) & 0.986 (0.013) & 0.805 (0.018)& 0.923 (0.029) \\
				II-3 &$(4, 200)$ & \textbf{0.357} (0.019) & 0.916 (0.021) & 0.970 (0.010) & 0.748 (0.014)& 0.832 (0.018) \\
				(SNR$=2$) &$(10, 500)$ & NA & NA & 0.967 (0.008) & \textbf{0.774} (0.014)& 0.929 (0.015) \\
				&$(10, 1000)$ & NA &NA & 0.959 (0.009) & \textbf{0.742} (0.012)& 0.879 (0.011) \\
				&$(20, 1000)$ &NA & NA  & 0.966 (0.009)  & \textbf{0.778} (0.011)& 0.956 (0.014)\\
				&$(20, 2000)$ & NA & NA  & 0.958 (0.008)  & \textbf{0.752} (0.010) & 0.917 (0.012)  \\
				\midrule
				&$(3, 100)$ & 0.744 (0.054) & 0.672 (0.189) & 0.773 (0.051) & \textbf{0.501} (0.051)& 0.540 (0.058) \\
				&$(3, 200)$ & 0.713 (0.049) & 0.481 (0.087) & 0.761 (0.049)  & \textbf{0.439} (0.037)& 0.465 (0.041)\\
				&$(4, 100)$ & 0.841 (0.065) & 0.834 (0.146) & 0.855 (0.064) & \textbf{0.676} (0.078)& 0.788 (0.125) \\
				II-4 &$(4, 200)$ & 0.835 (0.063) & 0.758 (0.113) & 0.841 (0.060) & \textbf{0.601} (0.073)& 0.684 (0.092) \\
				(SNR$=2$) &$(10, 500)$ & NA & NA & 0.838 (0.061) & \textbf{0.681} (0.076)& 0.773 (0.068) \\
				&$(10, 1000)$ & NA &NA & 0.829 (0.061) & \textbf{0.643} (0.077)& 0.718 (0.070) \\
				&$(20, 1000)$ &NA & NA  & 0.836 (0.061)  & \textbf{0.736} (0.079)& 0.838 (0.074)\\
				&$(20, 2000)$ & NA & NA  & 0.830 (0.060) & \textbf{0.698} (0.078) & 0.783 (0.072) \\
				\midrule
				&$(3, 100)$ & \textbf{0.346} (0.022) & 0.916 (0.136) & 0.965 (0.011) & 0.693 (0.017)& 0.755 (0.019) \\
				&$(3, 200)$ & \textbf{0.229} (0.011) & 0.774 (0.058) & 0.954 (0.008)  & 0.589 (0.012)& 0.616 (0.015)\\
				&$(4, 100)$ & \textbf{0.449} (0.033) & 0.948 (0.030) & 0.979 (0.012) & 0.789 (0.017)& 0.884 (0.025) \\
				II-3 &$(4, 200)$ & \textbf{0.284} (0.012) & 0.902 (0.026) & 0.966 (0.010) & 0.734 (0.013)& 0.802 (0.016) \\
				(SNR$=4$) &$(10, 500)$ & NA & NA & 0.964 (0.008) & \textbf{0.764} (0.013)& 0.894 (0.013) \\
				&$(10, 1000)$ & NA &NA & 0.958 (0.009) & \textbf{0.732} (0.012)& 0.847 (0.011) \\
				&$(20, 1000)$ &NA & NA  & 0.964 (0.009)  & \textbf{0.771} (0.010)& 0.919 (0.013)\\
				&$(20, 2000)$ & NA & NA  & 0.957 (0.008)  & \textbf{0.744} (0.010) & 0.886 (0.011)  \\
				\midrule
				&$(3, 100)$ & 0.736 (0.054) & 0.655 (0.199) & 0.770 (0.051) & \textbf{0.466} (0.054)& 0.477 (0.059) \\
				&$(3, 200)$ & 0.709 (0.049) & 0.439 (0.084) & 0.760 (0.048)  & 0.404 (0.040)& \textbf{0.401}  (0.043)\\
				&$(4, 100)$ & 0.841 (0.066) & 0.853 (0.162) & 0.851 (0.066) & \textbf{0.656} (0.083)& 0.746 (0.126) \\
				II-4 &$(4, 200)$ & 0.834 (0.063) & 0.758 (0.131) & 0.839  (0.060) & \textbf{0.582} (0.076)& 0.648 (0.093) \\
				(SNR$=4$) &$(10, 500)$ & NA & NA & 0.836 (0.061) & \textbf{0.675} (0.076)& 0.749 (0.069) \\
				&$(10, 1000)$ & NA &NA & 0.829 (0.061) & \textbf{0.636} (0.080)& 0.696 (0.074) \\
				&$(20, 1000)$ &NA & NA  & 0.835 (0.061)  & \textbf{0.734} (0.080)& 0.822 (0.078)\\
				&$(20, 2000)$ & NA & NA  & 0.830 (0.060) & \textbf{0.695} (0.076) & 0.769 (0.070) \\ 
				\bottomrule
			\end{tabular}
		}
		\caption{Average MSE (standard deviation) of different methods for setting II-3,4 with SNR = 2, 4 and Log-Cholesky metric over 100 simulation runs. Bold-faced numbers indicate the best performers.}
		\label{table:cov_Lin}
	\end{table}

   To maintain the consistency of the simulations, we use the same way as \cite{lin2022additive} to measure the quality of the estimation. For settings II-3 and II-4, we consider $p=3,4,10,20$ and two choices of $n$ for each $p$. For $p=3,4$, the setting is the same as  \cite{lin2022additive}. For $p=10,20$, we increase the dimension of $X$, but $Y$ is still only related to the first four components of $X$, \emph{i.e.}, $q=4$. Table~\ref{table:cov_Lin} shows the results. For setting II-3 where the underlying model is additive, MAM shows clear advantages when $p$ is relatively small. Although the three methods based on Fr\'echet trees are not optimal, they are significantly better compared to ILPR. For non-additive setting II-4, RFWLCFR/FRF tends to perform the best. This setting also indicates that there are indeed cases where RFWLLFR will perform worse than RFWLCFR. It may be more efficient for complex settings to use RFWLCFR, whose mechanism is similar to that of random forests. Since both MAM and GFR make specific model assumptions, setting II-4 is not suitable for these two methods, although MAM performs slightly better than GFR. In particular, when $p$ is greater than $10$, the implementations of MAM and ILPR often fail to work, and thus they are not feasible when the dimension of $X$ is large. Through this experiment, we again demonstrate the outstanding performance of our methods for relatively high-dimensional Fr\'echet regression. In real applications, we often lack prior knowledge of the structure of the underlying model, so it is important to develop regression methods that suit all situations with excellent performance.

	\section{Additional Materials for New York Taxi Data}\label{sec C}
\begin{figure}[h!]
		\centering
		\includegraphics[width=0.44\linewidth]{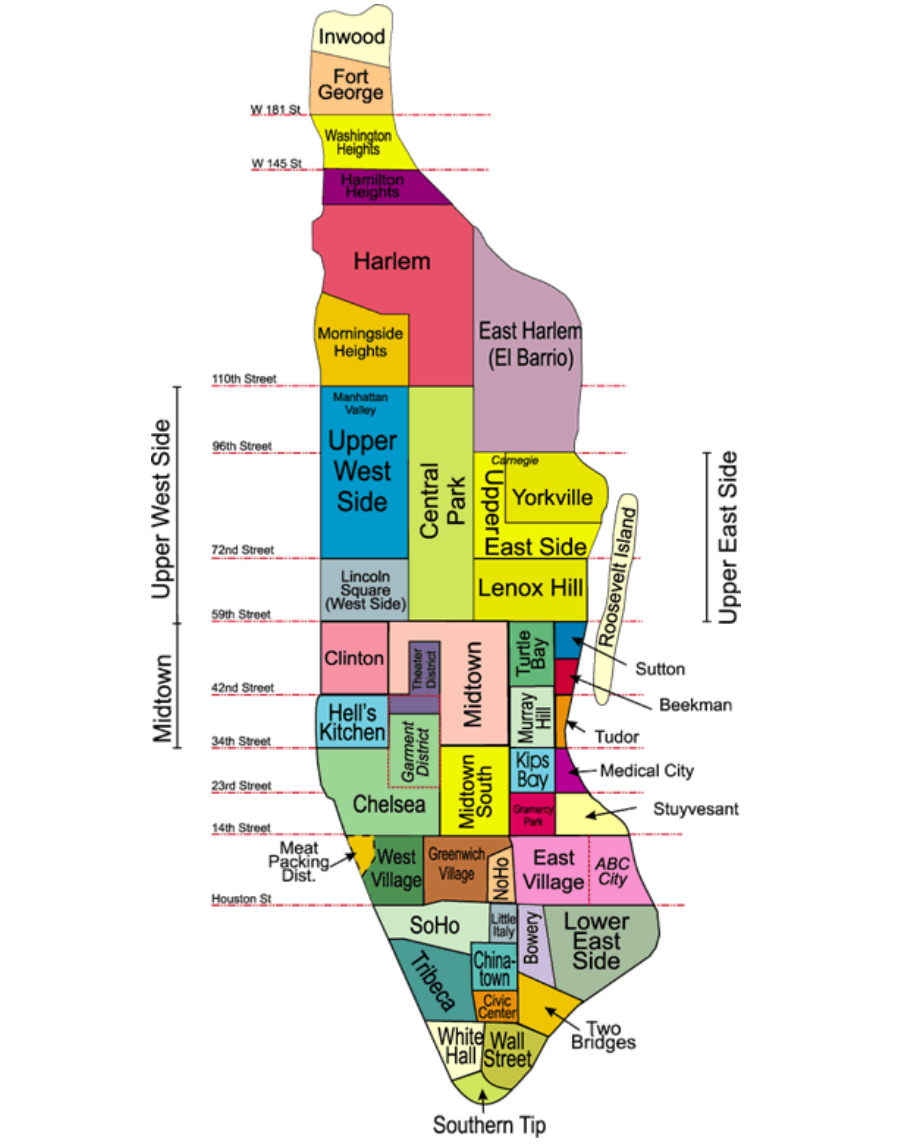}
		\includegraphics[width=0.3\linewidth]{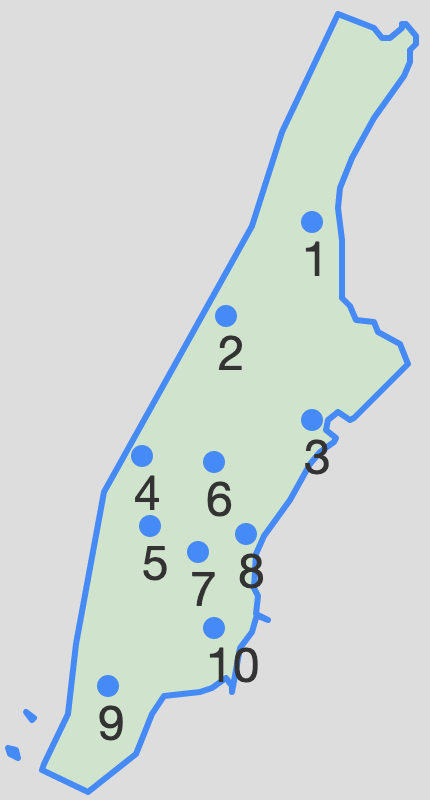}
		\caption{Towns in Manhattan (left) and center points for zones (right)}
		\label{fig:map}
	\end{figure}

	The map for Manhattan excluding the islands is delimited in Figure~\ref{fig:map} (from \cite{dubey2020functional}).  Manhattan can be further grouped into ten distinct zones, as detailed in Table~\ref{zones}, and the center point of each zone is selected as in Figure~\ref{fig:map}.
	\begin{table}[h!]
		\centering
		\resizebox{0.85\columnwidth}{!}{
			\begin{tabular}{|c|l|}
				\hline Zone & Towns \\
				\hline 1 & Inwood, Fort George, Washington Heights, Hamilton Heights, Harlem, East Harlem \\
				2 & Upper West Side, Morningside Heights, Central Park \\
				3 & Yorkville, Lenox Hill, Upper East Side \\
				4 & Lincoln Square, Clinton, Chelsea, Hell's Kitchen \\
				5 & Garment District, Theatre District \\
				6 & Midtown \\
				7 & Midtown South \\
				8 & Turtle Bay, Murray Hill, Kips Bay, Gramercy Park, Sutton, Tudor, Medical City, \\
				& Stuy Town \\
				9 & Meat packing district, Greenwich Village, West Village, Soho, Little Italy, \\
				& China Town, Civic Center, Noho \\
				10 & Lower East Side, East Village, ABD Park, Bowery, Two Bridges, Southern tip, \\
				& White Hall, Tribecca, Wall Street \\
				\hline
		\end{tabular}}
	\caption{10 zones in Manhattan defined for the New York taxi data analysis}
	\label{zones}
	\end{table}
	
	\section{Proofs}\label{sec D}
	In what follows, we prove our main results.
	\subsection{Some Preparation}\label{sec D.1}
	To facilitate the proof of important results in Section~\ref{sec3}, we give some preparatory work as follows. Recall 
	\begin{align} \nonumber
		\hat{r}_{\oplus}(x) &=\underset{y \in \Omega}{\argmin}\hat{R}_{n}(x,y)\\\label{RFWLCFR--1}
		&=\underset{y \in \Omega}{\argmin}\sum_{i=1}^{n}
		\bar{\alpha}_{i}\left(x\right) d^2\left(Y_{i},y\right)\\\label{RFWLCFR--2}
		&= \underset{y \in \Omega}{\argmin}\binom{n}{s_n}^{-1} \sum_{k}E_{\xi \sim \Xi}\Bigg\{\frac{1}{N(L(x;\mathcal{D}_n^k,\xi))} \sum_{i: X_{i} \in L(x;\mathcal{D}_n^k,\xi)}d^2\left(Y_{i},y\right)\Bigg\}.
	\end{align}
	and
	\begin{align}\nonumber
		\tilde{r}_{\oplus}(x)&=\underset{y \in \Omega}{\argmin}\tilde{R}_{n}(x,y)\\\label{P-RFWLCFR--1}
		&=\underset{y \in \Omega}{\argmin} \ n E \big\{\bar{\alpha}_{i}\left(x\right) d^2\left(Y_{i},y\right)\big\}\\\label{P-RFWLCFR--2}
		&=\underset{y \in \Omega}{\argmin}E\Bigg\{\frac{1}{N(L(x;\mathcal{D}_n^k,\xi))} \sum_{i: X_{i} \in L(x;\mathcal{D}_n^k,\xi)}d^2\left(Y_{i},y\right)\Bigg\}.
	\end{align}
	The goal of the Fr\'echet regression is
	\begin{align*}
		m_{\oplus}(x)=\underset{y \in \Omega}{\argmin} M_{\oplus}(x,y)=\underset{y \in \Omega}{\argmin} E\big\{d^{2}(Y, y) \mid X=x\big\}.
	\end{align*}
	We conduct asymptotic analysis  by separating $d(\hat{r}_{\oplus}(x),m_{\oplus}(x))$ into the bias term $d(\tilde{r}_{\oplus}(x), m_{\oplus}(x))$ and the variance term $d(\hat{r}_{\oplus}(x),\tilde{r}_{\oplus}(x))$. As  $\hat{r}_{\oplus}(x)$ have two different expressions, choosing a suitable form will bring great convenience for our theoretical developments. When we adopt  \eqref{RFWLCFR--2} and \eqref{P-RFWLCFR--2}, the theory of infinite order U-statistics and U-processes can be applied. And when we adopt \eqref{RFWLCFR--1} and \eqref{P-RFWLCFR--1}, the perspective of the weighted average is helpful.
	
	Under the honesty assumption, we have
	\begin{align*}
		E\Bigg\{\frac{1}{N(L(x;\mathcal{D}_n^k,\xi))} \sum_{i: X_{i} \in L(x;\mathcal{D}_n^k,\xi)}d^2\left(Y_{i},y\right)\Bigg\}=E\big[E\big\{d^2(Y,y) \mid X\in L(x)\big\} \big],
	\end{align*}
	where $L(x)$ is the leaf node containing $x$ of any honest Fr\'echet tree satisfying the assumption (A3). We emphasize here that $N(L(x;\mathcal{D}_n^k,\xi))$ and $X_{i} \in L(x;\mathcal{D}_n^k,\xi)$ don't involve sample points whose responses have been used to construct the Fr\'echet tree. Then \eqref{P-RFWLCFR--2} can be further rewritten as
	\begin{align*}
		\tilde{r}_{\oplus}(x)=\underset{y \in \Omega}{\argmin}\tilde{R}_{n}(x,y)=\underset{y \in \Omega}{\argmin}E\big[E\big\{d^2(Y,y) \mid X\in L(x)\big\} \big].
	\end{align*}
	
	\subsection{Proofs of Results in  Section~\ref{sec3.1}} \label{sec D.2}
	To prove Theorem~\ref{consistency}, we need to prove the convergence of the bias term $d(\tilde{r}_{\oplus}(x), m_{\oplus}(x))$ and variance term $d(\hat{r}_{\oplus}(x),\tilde{r}_{\oplus}(x))$, respectively. 
	\begin{lemma}
		\label{bias consistency}
		Suppose that for a fixed $x \in [0,1]^p$, (A1)--(A4) hold and the Fr\'echet trees are honest. Then,
		\begin{equation*}
			d(\tilde{r}_{\oplus}(x), m_{\oplus}(x))=o(1).
		\end{equation*}
	\end{lemma}
	\begin{proof}[Proof of Lemma~\ref{bias consistency}]
		By the proof of Theorem 3 in \cite{petersen2019frechet}, we have the relationship that $$\mathrm{d} F_{Y \mid X}(x, y) / \mathrm{d} F_{Y}(y)= g_{y}(x) / f(x)$$ for all $x$ such that $f(x)>0$. To prevent notation confusion, we consider a fixed $x_0 \in [0,1]^p$ and $y_{0}\in\Omega$, then
		\begin{align*}
			M_{\oplus}\left(x_0,y_{0}\right) = E\big\{d^2(Y, y_0) \mid X=x_0\big\} &=\int_{\Omega}d^2(y,y_0)dF_{Y \mid X}(x_0,y)\\
			&=\int_{\Omega}d^2(y,y_0)\frac{g_y(x_0)}{f(x_0)}dF_{Y}(y)
		\end{align*}
		and
		\begin{align*}
			E\big\{d^2(Y,y_0) \mid X\in L(x_0)\big\}
			& = \int_{L(x_0)}\left(\int_{\Omega}d^2(y,y_0)\frac{1}{P\left\{X\in L(x_0)\right\}}dF_{Y \mid X}(x,y)\right) f(x) dx\\
			& = \int_{L(x_0)}\int_{\Omega}d^2(y,y_0)\frac{g_y(x)}{P\left\{X\in L(x_0)\right\}}dF_Y(y) dx\\
			& = \int_{\Omega}d^2(y,y_0)\left(\int_{L(x_0)}\frac{g_y(x)}{P\left\{X\in L(x_0)\right\}}dx\right)dF_Y(y).
		\end{align*}
		It is obvious to have
		\begin{equation*}
			\int_{L(x_0)}\frac{g_y(x)}{P\left\{X\in L(x_0)\right\}}dx=\frac{\int_{L(x_0)}g_y(x)dx}{\int_{L(x_0)}f(x)dx}.
		\end{equation*}
		Since $g_y(x)$ is continuous by the assumption (A2), for every $(x_0,y) \in [0,1]^p\times\Omega$, $\forall$ $ \epsilon>0$, $\exists$ $\delta^1_{\epsilon}>0$ such that when $x \in B(x_0,\delta^1_{\epsilon}) = \left\{x:\|x-x_0\|\leq \delta^1_{\epsilon}\right\}$, we have
		$$\left|g_y(x)-g_y(x_0)\right|\leq \epsilon.$$
		Thus,  $ \exists$ $ \delta^1_{\epsilon}>0$ such that
		\begin{align}\label{fn}
			&\left|\int_{L(x_0)}  \{g_y(x)-g_y(x_0)\}dx \right| \notag \\
			\leq &\int_{L(x_0)}|g_y(x)-g_y(x_0)|dx \notag\\
			=&\int_{L(x_0)\cap B(x_0,\delta^1_{\epsilon})}\left|g_y(x)-g_y(x_0)\right|dx + \int_{L(x_0) \backslash B(x_0,\delta^1_{\epsilon})}\left|g_y(x)-g_y(x_0)\right|dx\notag \\
			\leq &\epsilon\Vol\left(L(x_0)\cap B(x_0,\delta^1_{\epsilon})\right) + 2\|g_y\|_{\infty}\Vol\left(L(x_0)\backslash B(x_0,\delta^1_{\epsilon})\right)\notag\\
			\leq &\epsilon\Vol\left(L(x_0)\right) + 2\|g_y\|_{\infty}\Vol\left(L(x_0)\backslash B(x_0,\delta^1_{\epsilon})\right)
		\end{align}
		where $\|g_y\|_{\infty}$ denotes the supremum norm of $g_y$ and $\Vol$ denotes the volume of subsets in the $[0,1]^p$. Since the marginal density $f$ is also continuous, using the same argument, for above $\epsilon$, $\exists$ $\delta_{\epsilon}^2$ such that
		\begin{equation}\label{gn}
			\int_{L(x_0)}\{f(x)-f(x_0)\}dx\leq \epsilon\Vol(L(x_0))+2\|f\|_{\infty}\Vol\left(L(x_0)\backslash B(x_0,\delta^2_{\epsilon})\right)
		\end{equation}
		We define $\delta_{\epsilon}=\min(\delta_{\epsilon}^1,\delta_{\epsilon}^2)$.  For two sequences of positive functions $\{f_n\}$, $\{g_n\}$ and for another  two  positive functions $\{f_n'\}$ and $\{g_n'\}$, we have
		\begin{equation} \label{fngn}
			\begin{aligned}
				\left|\frac{f_n}{g_n} - \frac{f_n'}{g_n'}\right| = \left|\frac{f_n}{g_n} - \frac{f_n'}{g_n} + \frac{f_n'}{g_n} - \frac{f_n'}{g_n'}\right| = \left|\frac{f_n-f_n'}{g_n}  - f_n'\frac{g_n'-g_n}{g_n'g_n}\right| \leq \frac{\left|f_n-f_n'\right|}{g_n} + f_n'\frac{\left|g_n-g_n'\right|}{g_n'g_n}.
			\end{aligned}
		\end{equation}
		We take \begin{equation*}
			f_n(x_0,y)=\int_{L(x_0)}g_y(x)dx;\qquad f_n'(x_0,y)=\int_{L(x_0)}g_y(x_0)dx;\\
		\end{equation*}
		\begin{equation*}
			g_n(x_0)=\int_{L(x_0)}f(x)dx;\qquad g_n'(x_0)=\int_{L(x_0)}f(x_0)dx.
		\end{equation*}
		By \eqref{fn}, for above $\epsilon$, we have
		\begin{align*}
			\frac{\left|f_n(x_0,y)-f_n'(x_0,y)\right|}{g_n(x_0)} &=\frac{\left|\int_{L(x_0)} g_y(x) d x-\int_{L(x_0)}g_y(x_0)dx\right|}{\int_{L(x_0)} f(x) d x} \\
			& \leq \frac{\left|\int_{L(x_0)}  \{g_y(x)-g_y(x_0)\}dx\right|}{f_{\min } \operatorname{Vol}\left(L(x_0)\right)} \\
			& \leq \frac{\epsilon \operatorname{Vol}\left(L(x_0)\right)+2\|g_y\|_{\infty} \operatorname{Vol}\left(L(x_0) \backslash B\left(x_0, \delta_{\epsilon}\right)\right)}{f_{\min } \operatorname{Vol}\left(L(x_0)\right)} \\
			&=\frac{\epsilon}{f_{\min }}+\frac{2\|g_y\|_{\infty} \operatorname{Vol}\left(L(x_0) \backslash B\left(x_0, \delta_{\epsilon}\right)\right)}{f_{\min } \operatorname{Vol}\left(L(x_0)\right)}.
		\end{align*}
		And similarly by \eqref{gn}, we have
		\begin{align*}
			\frac{\left|g_{n}(x_0)-g_n'(x_0)\right|}{g_{n}(x_0)}&=\frac{\left|\int_{L(x_0)} f(x) d x-\int_{L(x_0)}f(x_0)dx\right|}{\int_{L(x_0)} f(x) d x}
			\leq \frac{\epsilon}{f_{\min }}+\frac{2\left\|f\right\|_{\infty} \operatorname{Vol}\left(L(x_0) \backslash B\left(x_0, \delta_{\epsilon}\right)\right)}{f_{\min } \operatorname{Vol}\left(L(x_0)\right)}.
		\end{align*}
		By the assumption (A3),  $\operatorname{diam}\left(L(x_0)\right) \rightarrow 0$   in probability. Hence,  for $\delta_{\epsilon}$ defined above,
		$$
		\lim_{n \to +\infty}P\left\{\operatorname{diam}\left(L(x_0)\right)<\delta_{\epsilon}\right\}=1.
		$$
		Obviously when $\operatorname{diam}\left(L(x_0)\right)<\delta_{\epsilon}$, $ \operatorname{Vol}\left(L(x_0)\backslash B\left(x_0, \delta_{\epsilon}\right)\right)=0$ holds. Therefore
		$$
		P\left\{\operatorname{diam}\left(L(x_0)\right)<\delta_{\epsilon}\right\}\leq P\left\{\frac{\operatorname{Vol}\left(L(x_0) \backslash B\left(x_0, \delta_{\epsilon}\right)\right)}{\operatorname{Vol}\left(L(x_0)\right)}=0\right\}.
		$$
		Take the limit on both sides of the above formula, we have $\frac{\operatorname{Vol}\left(L(x_0) \backslash B\left(x_0, \delta_{\epsilon}\right)\right)}{\operatorname{Vol}\left(L(x_0)\right)}\rightarrow 0$ in probability. Combine with the assumption (A2), then
		\begin{equation}\label{limit}
			\frac{\left|f_n(x_0,y)-f_n'(x_0,y)\right|}{g_n(x_0)} \stackrel{P}{\rightarrow} \frac{\epsilon}{f_{\min}} \quad \mbox{and} \quad \frac{\left|g_{n}(x_0)-g_n'(x_0)\right|}{g_{n}(x_0)} \stackrel{P}{\rightarrow}  \frac{\epsilon}{f_{\min}}.
		\end{equation}
		Finally, combine \eqref{fngn} and \eqref{limit}, then the following formula holds
		\begin{align*}
			\left|\frac{\int_{L(x_0)}g_y(x)dx}{\int_{L(x_0)}f(x)dx}-\frac{g_y(x_0)}{f(x_0)}\right|
			=&\left|\frac{f_n(x_0,y)}{g_n(x_0)}-\frac{f_n'(x_0,y)}{g_n'(x_0)}\right|\\
			\leq &\frac{\left|f_n(x_0,y)-f_n'(x_0,y)\right|}{g_n(x_0)} + f_n'(x_0,y)\frac{\left|g_{n}(x_0)-g_n'(x_0)\right|}{g_n'(x_0)g_{n}(x_0)}\\ \stackrel{P}{\rightarrow}& \frac{\epsilon}{f_{\min}}+\frac{\epsilon}{f_{\min}}\frac{g_y(x_0)}{f(x_0)}.
		\end{align*}
		Let $\epsilon  \rightarrow 0$, we can get for each $y \in \Omega$
		\begin{equation*}
			\left|\frac{\int_{L(x_0)}g_y(x)dx}{\int_{L(x_0)}f(x)dx}-\frac{g_y(x_0)}{f(x_0)}\right|  \stackrel{P}{\rightarrow}  0.
		\end{equation*}
		Moreover,
		\begin{equation*}
			\left|\frac{\int_{L(x_0)}g_y(x)dx}{\int_{L(x_0)}f(x)dx}\right| \leq \frac{\|g_y\|_{\infty}\Vol(L(x_0))}{f_{\min}\Vol(L(x_0))}<\infty.
		\end{equation*}
		By the dominated convergence theorem and the assumption (A1), we  conclude
		\begin{equation}\label{conditional expectation}
			\begin{aligned}
				&\sup_{y_0 \in \Omega}\left|E\big\{d^2(Y,y_0) \mid X\in L(x_0)\big\}-M_{\oplus}\left(x_0,y_{0}\right)\right|\\
				=&\sup_{y_0 \in \Omega}\left|\int_{\Omega}d^2(y,y_0)\left\{\frac{\int_{L(x_0)}g_y(x)dx}{\int_{L(x_0)}f(x)dx}-\frac{g_y(x_0)}{f(x_0)}\right\}dF_Y(y)\right|\\
				\leq & \sup_{y_0 \in \Omega}\int_{\Omega}d^2(y,y_0)\left|\frac{\int_{L(x_0)}g_y(x)dx}{\int_{L(x_0)}f(x)dx}-\frac{g_y(x_0)}{f(x_0)}\right|dF_Y(y)\\
				\leq &\int_{\Omega} \sup_{y_0 \in \Omega}d^2(y,y_0)\left|\frac{\int_{L(x_0)}g_y(x)dx}{\int_{L(x_0)}f(x)dx}-\frac{g_y(x_0)}{f(x_0)}\right|dF_Y(y)\\
				\stackrel{P}{\rightarrow} &0.
			\end{aligned}
		\end{equation}
		Under the honest condition, we have
		\begin{align*}
			\tilde{R}_{n}(x_0,y_{0})=E\Bigg\{\frac{1}{N(L(x_0;\mathcal{D}_n^k,\xi))} \sum_{i: X_i \in L(x_0;\mathcal{D}_n^k,\xi)}d^2\left(Y_{i},y_0\right)\Bigg\}
			=E\big[E\big\{d^2(Y,y_0) \mid X\in L(x_0)\big\} \big].
		\end{align*}
		By the dominated convergence theorem, we take the expectation about $ L(x_0)$ on both sides of \eqref{conditional expectation} and get
		\begin{align*}
			&\sup_{y_0 \in \Omega}\left|\tilde{R}_{n}(x_0,y_{0})-M_{\oplus}\left(x_0,y_{0}\right)\right| \\
			= & \sup_{y_0 \in \Omega}\left| E\left[E\big\{d^2(Y,y_0)|X\in L(x_0)\big\} - M_{\oplus}\left(x_0,y_{0}\right)\right] \right| \\
			\leq & \sup_{y_0 \in \Omega}E\left| E\big\{d^2(Y,y_0)|X\in L(x_0)\big\} - M_{\oplus}\left(x_0,y_{0}\right)\right| \\
			\leq & E\left(\sup_{y_0 \in \Omega}\left| E\big\{d^2(Y,y_0)|X\in L(x_0)\big\} - M_{\oplus}\left(x_0,y_{0}\right)\right|\right) \\
			\rightarrow &0.
		\end{align*}
		By the assumption (A4), we then get
		$$
		d(\tilde{r}_{\oplus}(x_0),m_{\oplus}(x_0))=o(1).
		$$
	\end{proof}
	
	\begin{lemma}
		\label{variance consistency}
		Suppose that for a fixed $x \in[0,1]^p$, (A1), (A4) hold and the Fr\'echet  trees are symmetric. Then,
		\begin{equation*}
			d(\hat{r}_{\oplus}(x),\tilde{r}_{\oplus}(x))=o_{p}(1).
		\end{equation*}
	\end{lemma}
	
	Before giving the proof of Lemma~\ref{variance consistency}, we need to prove another lemma first. It is the generalization of Corollary 3.2.3 of  \cite{van1996weak}.
	
	\begin{lemma}\label{consistency of  M}
		Let $\mathbb{M}_{n}$ be stochastic processes indexed by a metric space $\Theta$, and let $M_n: \Theta \mapsto \mathcal{R}$ be deterministic functions. Suppose that $\left\|\mathbb{M}_{n}-M_n\right\|_{\Theta} \rightarrow 0$ in probability and that there exists a sequence  $\theta_{n}$ such that
		$$
		\liminf _{n} \inf _{d\left(\theta, \theta_n \right)>\varepsilon}\left\{{M}_{n}(\theta)-{M}_{n}(\theta_n)\right\}>0
		$$
		for every $\varepsilon > 0$. Then any sequence $\hat{\theta}_{n}$, such that $\mathbb{M}_{n}(\hat{\theta}_{n}) \leq \inf _{\theta} \mathbb{M}_{n}(\theta)+o_{P}(1)$, satisfies $d(\hat{\theta}_{n}, {\theta}_{n})\rightarrow0$ in probability.
	\end{lemma}

	\begin{proof}[Proof of Lemma~\ref{consistency of  M}]
		By the requirement of $\hat{\theta}_{n}$, $\mathbb{M}_{n}(\hat{\theta}_{n}) \leq \mathbb{M}_{n}(\theta_n)+o_{P}(1)$. Since $\left\|\mathbb{M}_{n}-M_n\right\|_{\Theta} \rightarrow 0$ in  probability, $\mathbb{M}_{n}(\theta_{n})-M_{n}(\theta_{n}) \rightarrow 0$ in  probability. So we have
		$$\mathbb{M}_{n}(\hat{\theta}_{n}) \leq M_{n}(\theta_n)+o_{P}(1).$$
		Therefore, again by  the uniform convergence, we have
		\begin{equation}\label{the uniform convergence}
			\begin{aligned}
				M_{n}(\hat{\theta}_n)-M_{n}(\theta_n) &\leq M_{n}(\hat{\theta}_{n}) -\mathbb{M}_{n}(\hat{\theta}_{n})+o_{P}(1)\\
				&\leq \left\|\mathbb{M}_{n}-M_n\right\|_{\Theta}+o_{P}(1)\\
				&\stackrel{P}{\rightarrow} 0.
			\end{aligned}
		\end{equation}
		From the requirement of $\theta_{n}$, given $ \epsilon>0$, there exists $\delta>0, N \in \mathcal{N}^{+}$, when $n \geq N$ and $d(\theta, \theta_n) \geq \epsilon$, we have
		$$
		M_n(\theta)-M_n\left(\theta_{n}\right)>\delta.
		$$
		So $\left\{d(\hat{\theta}_n, \theta_n) \geq \epsilon \right\} \subseteq\left\{M_n(\hat{\theta}_n)-M_n\left(\theta_{n}\right)>\delta\right\}$, however, by \eqref{the uniform convergence}
		$$
		P\left\{M_n(\hat{\theta}_{n})-M_n(\theta_n)>\delta\right\} \rightarrow 0.
		$$
		Therefore
		$$
		P\left\{d(\hat{\theta}_n, \theta_n) \geq \epsilon\right\} \rightarrow 0.
		$$
		\emph{i.e.},
		$$d(\hat{\theta}_{n}, {\theta}_{n})\rightarrow0 \ \ \text{in  probability}.$$
		
	\end{proof}

	\begin{proof}[Proof of Lemma~\ref{variance consistency}]
		For a fixed $x \in [0,1]^p$, by Lemma \ref{consistency of  M}  and the assumption (A4),  we only need to prove  convergence of $\sup _{y \in \Omega} \left| \hat{R}_{n}(x,y)-\tilde{R}_{n}(x,y) \right|$ to zero in probability. To implement this, we can show $\hat{R}_{n}(x,\cdot) - \tilde{R}_{n}(x,\cdot)\rightsquigarrow 0$ in $l^{\infty}(\Omega)$ which denotes the space of bounded functions on $\Omega$, and apply Theorem 1.3.6 of  \cite{van1996weak}.  Thanks to Theorem 1.5.4 of  \cite{van1996weak}, this weak convergence is equivalent to $\hat{R}_{n}(x,\cdot) - \tilde{R}_{n}(x,\cdot)$ is asymptotically tight and the marginals converge weakly. By Theorem 1.5.7 of  \cite{van1996weak}, the asymptotically tight continues to be equivalent to two requirements that $\hat{R}_{n}(x,y) - \tilde{R}_{n}(x,y)$ is asymptotically tight in $\mathcal{R}$ for every $y \in \Omega$ and $\hat{R}_{n}(x,\cdot)- \tilde{R}_{n}(x,\cdot)$ is asymptotically uniformly $d$-equicontinuous in probability. So the proof will be finished if the following conditions hold
		
		(i) $\hat{R}_{n}(x,y)-\tilde{R}_{n}(x,y)=o_{p}(1)$ for each $y \in \Omega$,
		
		(ii) For all $\varepsilon, \eta>0$, there exists $\delta>0$ such that
		$$
		\limsup _{n}P\left\{\sup _{d\left(y_{1}, y_{2}\right)<\delta}\left|\left(\hat{R}_{n}-\tilde{R}_{n}\right)\left(x,y_{1}\right)-\left(\hat{R}_{n}-\tilde{R}_{n}\right)\left(x,y_{2}\right)\right|>\varepsilon\right\}<\eta.
		$$
		
		First, prove (i):
		We consider the expressions of \eqref{RFWLCFR--2} and \eqref{P-RFWLCFR--2}. $\hat{R}_{n}(x,y)$ is an infinite order U-statistic  for each  $y\in\Omega$. Since $(\Omega, d)$ is a bounded metric space by the assumption (A1),  the kernels of $\hat{R}_{n}(x,y)$  are uniformly bounded. By  Remark 3.1 of \cite{diciccio2022clt}, we have $\hat{R}_{n}(x,y)-\tilde{R}_{n}(x,y)=o_{p}(1)$ for each $y \in \Omega$.
		
		Then (ii): We consider the expressions of \eqref{RFWLCFR--1} and \eqref{P-RFWLCFR--1}. For any $y_{1}$, $y_{2}\in\Omega$,
		\begin{align*}
			&\left|\left(\hat{R}_{n}-\tilde{R}_{n}\right)\left(x,y_{1}\right)-\left(\hat{R}_{n}-\tilde{R}_{n}\right)\left(x,y_{2}\right)\right| \\
			\leq &\left|\hat{R}_{n}(x,y_{1})-\hat{R}_{n}(x,y_{2})\right|+ \left|\tilde{R}_{n}(x,y_{1})-\tilde{R}_{n}(x,y_{2})\right|\\
			= &\left|\sum_{i=1}^{n}\bar{\alpha}_{i}(x) \left\{d^2\left(Y_{i}, y_{1}\right)-d^2\left(Y_{i}, y_{2}\right)\right\}\right|+ \left|n E\left[\bar{\alpha}_{i}(x) \left\{d^2\left(Y_{i}, y_{1}\right)-d^2\left(Y_{i}, y_{2}\right)\right\}\right]\right|\\
			\leq &\sum_{i=1}^{n}\left|\bar{\alpha}_{i}(x)\left\|d\left(Y_{i}, y_{1}\right)-d\left(Y_{i}, y_{2}\right)\right\| d\left(Y_{i}, y_{1}\right)+d\left(Y_{i}, y_{2}\right)\right|\\ &+nE\left\{\left|\bar{\alpha}_{i}(x)\left\|d\left(Y_{i}, y_{1}\right)-d\left(Y_{i}, y_{2}\right)\right\| d\left(Y_{i}, y_{1}\right)+d\left(Y_{i}, y_{2}\right)\right|\right\}\\
			\leq & 2 \operatorname{diam}(\Omega) d\left(y_{1}, y_{2}\right) \sum_{i=1}^{n}\bar{\alpha}_{i}(x) + 2 \operatorname{diam}(\Omega) d\left(y_{1}, y_{2}\right)\left[nE\left\{\bar{\alpha}_{i}(x)\right\}\right]\\
			= &4 \operatorname{diam}(\Omega) d\left(y_{1}, y_{2}\right)\\
			= &O_{p}\left(d\left(y_{1}, y_{2}\right)\right)
		\end{align*}
		where the $O_{p}$ term is independent of $y_{1}$ and $y_{2}$. The second equality is because $\bar{\alpha}_{i}(x), i=1,\dots,n$ are identically distributed and  a fact
		\begin{align*}
			\sum_{i=1}^{n}\bar{\alpha}_{i}(x)&=\sum_{i=1}^{n}  \binom{n}{s_{n}}^{-1} \sum_{k}E_{\xi \sim \Xi} \frac{1\left\{X_{i} \in L(x; \mathcal{D}_{n}^{k},\xi)\right\}}{N(L(x; \mathcal{D}_{n}^{k},\xi))}\\
			&=\binom{n}{s_{n}}^{-1} \sum_{k}E_{\xi \sim \Xi} \sum_{i=1}^{n} \frac{1\left\{X_{i} \in L(x; \mathcal{D}_{n}^{k},\xi)\right\}}{N(L(x;\mathcal{D}_{n}^{k},\xi))}\\
			&=1.
		\end{align*}
		Hence
		$$
		\sup _{d\left(y_{1}, y_{2}\right)<\delta}\left|\left(\hat{R}_{n}-\tilde{R}_{n}\right)\left(x,y_{1}\right)-\left(\hat{R}_{n}-\tilde{R}_{n}\right)\left(x,y_{2}\right)\right|=O_{p}(\delta)
		$$
		which can deduce (ii).  So, $d(\hat{r}_{\oplus}(x),\tilde{r}_{\oplus}(x))=o_{p}(1)$
	\end{proof}
	
	Based on Lemma~\ref{bias consistency} and Lemma~\ref{variance consistency}, we can prove Theorem~\ref{consistency} easily.
	\begin{proof}[Proof of Theorem~\ref{consistency}]
		Notice that
		$$d(\hat{r}_{\oplus}(x),m_{\oplus}(x))\leq d(\hat{r}_{\oplus}(x),\tilde{r}_{\oplus}(x))+d(\tilde{r}_{\oplus}(x),m_{\oplus}(x)).$$
		By the results of Lemma~\ref{bias consistency} and Lemma~\ref{variance consistency}, we complete the proof.
	\end{proof}
	
	\vspace{0.5cm}
	Before the proof of Theorem~\ref{uniform consistency}, we state the required assumptions (U1)--(U4).
	
	(U1)
	For any $\Vert x \Vert \leq J$, $M_{\oplus}(x, y)$ is equicontinuous, \emph{i.e.},
	$$
	\limsup _{\dot{x} \rightarrow x} \sup _{y \in \Omega}\left|M_{\oplus}(\dot{x}, y)-M_{\oplus}(x, y)\right|=0.
	$$
	
	(U2)
	The marginal density $f$ of $X$, as well as the conditional densities $g_{y}$ of $X \mid Y=y$, exist and are bounded and uniformly continuous, the latter for all $y \in \Omega$. And $f$ is also bounded away from zero such that $0<f_{\min } \leq f$. Additionally, for any open $V \subseteq \Omega$, $\int_{V} \mathrm{~d} F_{Y \mid X}(x, y)$ is continuous as a function of $x$.
	
	(U3)
	$\sup_{\Vert x \Vert \leq J}\operatorname{diam}\left(L(x)\right) \rightarrow 0$  in probability, where $L(x)$ is the leaf node containing $x$ of any Fr\'echet tree in the random forest.
	
	(U4)
	For all $\Vert x \Vert \leq J$, $m_{\oplus}(x)$, $\tilde{r}_{\oplus}(x)$ and $\hat{r}_{\oplus}(x)$ exist and are unique, the latter almost surely. Additionally, for any $\varepsilon>0$,
	$$
	\inf_{\Vert x \Vert \leq J}\inf _{d\left(y, m_{\oplus}(x)\right)>\varepsilon}\left\{M_{\oplus}(x,y)-M_{\oplus}\left(x,m_{\oplus}(x)\right)\right\}>0,
	$$
	$$
	\liminf _{n} \inf_{\Vert x \Vert \leq J} \inf _{d\left(y, \tilde{r}_{\oplus}(x)\right)>\varepsilon}\left\{\tilde{R}_{n}(x,y)-\tilde{R}_{n}\left(x, \tilde{r}_{\oplus}(x)\right)\right\}>0,
	$$
	and there exists $\zeta=\zeta(\varepsilon)>0$ such that
	$$
	P\left\{ \inf _{\|x\| \leq J} \inf _{d\left(y, \hat{r}_{\oplus}(x)\right)>\varepsilon}\left( \hat{R}_{n}(x, y)-\hat{R}_{n}\left(x,\hat{r}_{\oplus}(x)\right) \right)\geq \zeta\right\} \rightarrow 1.
	$$
	
	\begin{proof}[Proof of Theorem~\ref{uniform consistency}]
		We need to prove the following two results:
		
		(i) $\sup_{\Vert x \Vert \leq J} d(\hat{r}_{\oplus}(x),\tilde{r}_{\oplus}(x))=o_p(1)$,
		
		(ii) $\sup_{\Vert x \Vert \leq J} d(\tilde{r}_{\oplus}(x),m_{\oplus}(x))=o(1)$.

		First prove (i): By Lemma~\ref{variance consistency}, given any $x \in [0,1]^p$,  $ d(\hat{r}_{\oplus}(x),\tilde{r}_{\oplus}(x))=o_p(1)$. Now consider the process $D_n(x)= d(\hat{r}_{\oplus}(x),\tilde{r}_{\oplus}(x))$ with $\Vert x \Vert \leq J$. As the proof of Lemma~\ref{variance consistency}, based on Theorems 1.5.4, 1.5.7 and 1.3.6 of \cite{van1996weak}, it suffices to show that for any $S>0$, as $\delta \rightarrow 0$,
		$$
		\limsup _{n \rightarrow \infty} P\left\{\sup _{\|x_1-x_2\|<\delta \atop\|x_1\|,\|x_2\| \leq J}\left|D_{n}(x_1)-D_{n}(x_2)\right|>2 S\right\} \rightarrow 0.
		$$
		Since
		\begin{align*}
			&\left|D_{n}(x)-D_{n}(y)\right|\\ =&\left|d(\hat{r}_{\oplus}(x_1),\tilde{r}_{\oplus}(x_1))-d(\hat{r}_{\oplus}(x_2),\tilde{r}_{\oplus}(x_2))\right|\\
			=&\left|d(\hat{r}_{\oplus}(x_1),\tilde{r}_{\oplus}(x_1))-d(\hat{r}_{\oplus}(x_1),\tilde{r}_{\oplus}(x_2))+d(\hat{r}_{\oplus}(x_1),\tilde{r}_{\oplus}(x_2))-d(\hat{r}_{\oplus}(x_2),\tilde{r}_{\oplus}(x_2))\right|\\
			\leq & d(\tilde{r}_{\oplus}(x_1),\tilde{r}_{\oplus}(x_2))+d(\hat{r}_{\oplus}(x_1),\hat{r}_{\oplus}(x_2)),
		\end{align*}
		it suffices to show that, as  $\delta \rightarrow 0$,
		\begin{align}\label{3.5.1}
			\limsup _{n \rightarrow \infty} \sup _{\|x_1-x_2\|<\delta \atop\|x_1\|,\|x_2\| \leq J} d(\tilde{r}_{\oplus}(x_1),\tilde{r}_{\oplus}(x_2)) \rightarrow 0,
		\end{align}
		and
		\begin{align}\label{3.5.2}
			\limsup _{n \rightarrow \infty} P\left\{\sup _{\|x_1-x_2\|<\delta \atop\|x_1\|,\|x_2\| \leq J} d(\hat{r}_{\oplus}(x_1),\hat{r}_{\oplus}(x_2))>S\right\} \rightarrow 0.
		\end{align}
		Recall that we have proved $\sup_{y \in \Omega}\left|\tilde{R}_{n}(x,y)-M_{\oplus}\left(x,y\right)\right| \rightarrow 0$ for any $x \in [0,1]^p$ in the proof of Lemma~\ref{bias consistency}. Since the density $f$ and $g_f$ are uniformly continuous by the assumption (U2) and $\sup_{\Vert x \Vert \leq J}\operatorname{diam}\left(L(x)\right) \rightarrow 0$  in probability by the assumption (U3), we can get stronger convergence
		\begin{align}\label{3.5.3}
			\sup_{\Vert x \Vert \leq J, y \in \Omega}\left|\tilde{R}_{n}(x,y)-M_{\oplus}\left(x,y\right)\right| \rightarrow 0.
		\end{align}
		Notice that
		\begin{align*}
			&\sup _{\|x_1-x_2\|<\delta \atop \|x_1\|,\|x_2\| \leq J} \sup _{y \in \Omega}\left|\tilde{R}_{n}(x_1, y)-\tilde{R}_{n}(x_2, y)\right| \\
			\leq & \sup _{\|x_1-x_2\|<\delta \atop \|x_1\|,\|x_2\| \leq J} \sup _{y \in \Omega}\left|M_{\oplus}(x_1, y)-M_{\oplus}(x_2, y)\right|+2 \sup_{\Vert x \Vert \leq J, y \in \Omega}\left|\tilde{R}_
			{n}(x,y)-M_{\oplus}\left(x,y\right)\right|.
		\end{align*}
		Combining the assumption (U1) and \eqref{3.5.3}, we can obtain, as $\delta \rightarrow 0$,
		\begin{align*}
			\limsup _{n \rightarrow \infty} \sup _{\|x_1-x_2\|<\delta \atop\|x_1\|,\|x_2\| \leq J} \sup _{y \in \Omega}\left|\tilde{R}_{n}(x_1, y)-\tilde{R}_{n}(x_2, y)\right| \rightarrow 0.
		\end{align*}
		Then \eqref{3.5.1} holds by assumption (U4). Now consider \eqref{3.5.2},
		let $\epsilon > 0$ and suppose $d(\hat{r}_{\oplus}(x_1),\hat{r}_{\oplus}(x_2)) > \epsilon$ with
		$\|x_1\|,\|x_2\| \leq J$. Then the assumption (U4) and the form of $\hat{R}_n(x,y)$ imply that
		\begin{align*}
			\zeta & \leq \sup _{\|x_1-x_2\|<\delta \atop\|x_1\|,\|x_2\| \leq J} \sup _{y \in \Omega}\left|\hat{R}_{n}(x_1, y)-\hat{R}_{n}(x_2, y)\right|=O_p(\delta)
		\end{align*}
		and \eqref{3.5.2} follows as $\delta \rightarrow 0$. At this point, the proof of (i) is finished.
		
		Next prove (ii): By Lemma~\ref{bias consistency}, given any $x \in [0,1]^p$,  $ d(\tilde{r}_{\oplus}(x),m_{\oplus}(x))=o(1)$. Similarly, we consider  $F_n(x)= d(\tilde{r}_{\oplus}(x),m_{\oplus}(x))$. It suffices to show that, as $\delta \rightarrow 0$,
		$$
		\limsup _{n \rightarrow \infty} \sup _{\|x_1-x_2\|<\delta \atop\|x_1\|,\|x_2\| \leq J}\left|F_{n}(x_1)-F_{n}(x_2)\right| \rightarrow 0.
		$$
		Since
		\begin{align*}
			&\left|F_{n}(x_1)-F_{n}(x_2)\right| 
			\leq  d(m_{\oplus}(x_1),m_{\oplus}(x_2))+d(\tilde{r}_{\oplus}(x_1),\tilde{r}_{\oplus}(x_2)),
		\end{align*}
		it suffices to show that, as  $\delta \rightarrow 0$,
		\begin{align}\label{3.5.4}
			\sup _{\|x_1-x_2\|<\delta \atop\|x_1\|,\|x_2\| \leq J} d(m_{\oplus}(x_1),m_{\oplus}(x_2)) \rightarrow 0,
		\end{align}
		and
		\begin{align}\label{3.5.5}
			\limsup _{n \rightarrow \infty} \sup _{\|x_1-x_2\|<\delta \atop\|x_1\|,\|x_2\| \leq J} d(\tilde{r}_{\oplus}(x_1),\tilde{r}_{\oplus}(x_2)) \rightarrow 0.
		\end{align}
		Based on the assumption (U1) and (U4), it is not difficult to prove that $m_{\oplus}(x)$ is continuous at $x$ and hence uniformly continuous on  $\left\{x: \Vert x \Vert \leq J\right\}$ considering the compactness of $\left\{x: \Vert x \Vert \leq J\right\}$. Then \eqref{3.5.4} naturally holds.  And \eqref{3.5.5} has been solved in part (i).
		
		Therefore, based on the results of (i) and (ii), it follows that
		\begin{align*}
			\sup_{\Vert x \Vert \leq J}	d(\hat{r}_{\oplus}(x), m_{\oplus}(x)) 
			\leq &\sup_{\Vert x \Vert \leq J}\left( d(\hat{r}_{\oplus}(x),\tilde{r}_{\oplus}(x))+d(\tilde{r}_{\oplus}(x),m_{\oplus}(x)) \right)\\
			\leq & \sup_{\Vert x \Vert \leq J} d(\hat{r}_{\oplus}(x),\tilde{r}_{\oplus}(x)) +  \sup_{\Vert x \Vert \leq J} d(\tilde{r}_{\oplus}(x),m_{\oplus}(x)) \\
			= &o_{p}(1).
		\end{align*}
	\end{proof}
	
	\subsection{Proofs of Results in Section~\ref{sec3.2}}\label{sec D.3}
	
	\begin{proof}[Proof of Lemma~\ref{bias rate}]
		Since $X \in [0, 1]^p$
		with a density $\rho_X$ bounded away from 0 and infinity. By Lemma 2 of \cite{wager2018estimation},  for any $0<\eta<1$, and for large enough $s_n$,
		\begin{equation}\label{bound of diameter}
			P\left\{\operatorname{diam}_{j}(L(x)) \geq\left(\frac{s_n}{2 k-1}\right)^{-\frac{0.99(1-\eta) \log \left((1-\alpha)^{-1}\right)}{\log \left(\alpha^{-1}\right)} \frac{\pi}{p}}\right\} \leq\left(\frac{s_n}{2 k-1}\right)^{-\frac{\eta^{2}}{2} \frac{1}{\log \left(\alpha^{-1}\right)} \frac{\pi}{p}}.
		\end{equation}
		where  $\operatorname{diam}_{j}(L(x))$ denote the length of the  $j$th dimension of the leaf $L(x)$.  If the honest tree is implemented by the double-sample tree,
		the above argument still holds by simply replacing $s_n$ with $s_n/2$. But it does not affect the final result.
		
		The above bound can derive $\operatorname{diam}\left(L(x)\right) \rightarrow 0$   in probability. Then by Lemma~\ref{bias consistency}, we have $d(\tilde{r}_{\oplus} (x),m_{\oplus}(x))=o(1)$.
		
		Since the Fr\'echet trees are honest,
		\begin{align*}
			\tilde{R}_{n}(x,y)- M_{\oplus}(x,y)
			=&E\left[E\left\{d^2(Y,y) \mid X \in L(x)\right\}\right]-E\left\{d^2(Y,y) \mid X=x\right\}\\
			=&E\left[E\left\{d^2(Y,y) \mid X \in L(x)\right\}-E\left\{d^2(Y,y) \mid X=x\right\}\right].
		\end{align*}
		By  the assumption (A5),
		$$
		\left| E\left\{d^2(Y,y) \mid X \in L(x)\right\}-E\left\{d^2(Y,y) \mid X=x\right\}\right| \leq K \operatorname{diam}(L(x)).
		$$
		Now take the same approach as the proof of Theorem 3 of \cite{wager2018estimation} to bound the diameter of $L(x)$. By plugging in $\eta=\sqrt{\log \left((1-\alpha)^{-1}\right)}$ in the bound from \eqref{bound of diameter}. Since $\alpha \leq 0.2$, we see that $\eta \leq 0.48$ and so $0.99 \cdot(1-\eta) \geq 0.51$; thus, a union bound gives us that, for large enough $s_n$,
		$$
		P\left\{\operatorname{diam}(L(x)) \geq p^{1/2}\left(\frac{s_n}{2 k-1}\right)^{-0.51 \frac{\log \left((1-\alpha)^{-1}\right)}{\log (\alpha^{-1})} \frac{\pi}{p}}\right\} \leq p\left(\frac{s_n}{2 k-1}\right)^{-\frac{1}{2} \frac{\log \left((1-\alpha)^{-1}\right)}{\log (\alpha^{-1})} \frac{\pi}{p}}.
		$$
		The Lipschitz assumption lets us bound $\tilde{R}_{n}(x,y)- M_{\oplus}(x,y)$ base on the  above result about $\operatorname{diam}(L(x))$. Specifically, let
		$$
		a_1=p^{1/2}\left(\frac{s_n}{2 k-1}\right)^{-0.51 \frac{\log \left((1-\alpha)^{-1}\right)}{\log (\alpha^{-1})} \frac{\pi}{p}},\quad a_2 =p\left(\frac{s_n}{2 k-1}\right)^{-\frac{1}{2} \frac{\log \left((1-\alpha)^{-1}\right)}{\log (\alpha^{-1})} \frac{\pi}{p}}.
		$$
		Then
		$$
		P\left\{\operatorname{diam}(L(x)) \geq a_1\right\} \leq a_2.
		$$
		We have
		\begin{align*}
			&\left|\tilde{R}_{n}(x,y)- M_{\oplus}(x,y)\right|\\
			\leq&\left|E\left[E\left\{d^2(Y,y) \mid X \in L(x)\right\}-E\left\{d^2(Y,y) \mid X=x\right\}\right]\right|\\
			\leq&\left|E\left(\left[E\left\{d^2(Y,y) \mid X \in L(x)\right\}-E\left\{d^2(Y,y) \mid X=x\right\}\right] I(\operatorname{diam}(L(x)) \geq a_1)\right)\right|\\
			&+\left|E\left(\left[E\left\{d^2(Y,y) \mid X \in L(x)\right\}-E\left\{d^2(Y,y) \mid X=x\right\}\right] I(\operatorname{diam}(L(x)) < a_1)\right)\right|\\
			\leq&E\left[\left|E\left\{d^2(Y,y) \mid X \in L(x)\right\}-E\left\{d^2(Y,y) \mid X=x\right\}\right| I(\operatorname{diam}(L(x)) \geq a_1)\right]\\
			&+E\left[\left|E\left\{d^2(Y,y) \mid X \in L(x)\right\}-E\left\{d^2(Y,y) \mid X=x\right\}\right| I(\operatorname{diam}(L(x)) < a_1)\right]\\
			\leq & \left(\sup _{x \in[0,1]^{p}}\left[E\left\{d^2(Y,y) \mid X=x\right\}\right]-\inf _{x \in[0,1]^{p}}\left[E\left\{d^2(Y,y) \mid X=x\right\}\right]\right) P\left\{\operatorname{diam}(L(x)) \geq a_1\right\} +K a_1\\
			\leq & \left(\sup _{x \in[0,1]^{p}}\left[E\left\{d^2(Y,y) \mid X=x\right\}\right]-\inf _{x \in[0,1]^{p}}\left[E\left\{d^2(Y,y) \mid X=x\right\}\right]\right) a_2 +K a_1\\
			\lesssim & \left(\sup _{x \in[0,1]^{p}}\left[E\left\{d^2(Y,y) \mid X=x\right\}\right]-\inf _{x \in[0,1]^{p}}\left[E\left\{d^2(Y,y) \mid X=x\right\}\right]\right) a_2,
		\end{align*}
		since $a_1/a_2 \rightarrow 0$.\\
		Due to the Lipschitz condition,
		\begin{align*}
			\sup _{x \in[0,1]^{p}}\left[E\left\{d^2(Y,y) \mid X=x\right\}\right]-\inf _{x \in[0,1]^{p}}\left[E\left\{d^2(Y,y) \mid X=x\right\}\right] 
			&\leq  K \sup _{x_1,x_2 \in[0,1]^{p}} \|x_1-x_2\| \\
			&= K p^{1/2}.
		\end{align*}
		So for large enough $s_n$,
		$$
		\left|\tilde{R}_{n}(x,y)- M_{\oplus}(x,y)\right|=O\left(s_n^{-\frac{1}{2} \frac{\log \left(1-\alpha\right)}{\log (\alpha)} \frac{\pi}{p}}\right).
		$$
		The above bound is uniform over $y \in \Omega$. Let $T_n(x,y)=\tilde{R}_{n}(x,y)-M_{\oplus}(x,y)$, then easily we can get, for any $\delta>0$,
		\begin{align}\label{maximal inequality2}
			\sup _{d\left(y, m_{\oplus}(x)\right)<\delta}\left|T_{n}(x,y)-T_{n}\left(x,m_{\oplus}(x)\right)\right| \lesssim c_1 \delta s_n^{-\frac{1}{2} \frac{\log \left(1-\alpha\right)}{\log (\alpha)} \frac{\pi}{p}}
		\end{align}
		for some constant $c_1>0$.
		
		Now, set $t_{n}=s_n^{\frac{1}{4} \frac{\log \left(1-\alpha\right)}{\log (\alpha)} \frac{\pi}{p} \frac{\beta_1}{\beta_1-1}}$ and define
		\begin{align*}
			S_{j, n}=\left\{y: 2^{j-1}<t_{n} d\left(y, m_{\oplus}(x)\right)^{\beta_1 / 2} \leq 2^{j}\right\}.
		\end{align*}
		Choose $\delta_{1}$ satisfying the assumption (A6). Set $\tilde{\delta}_1:=\left(\delta_{1} \right)^{\beta_{1} / 2}$. For any integer $M$,
		\begin{equation}\label{shells2}
			\begin{aligned}
				&I\left\{t_{n} d\left(\tilde{r}_{\oplus}(x), m_{\oplus}(x)\right)^{\beta_1/ 2}>2^{M}\right\}\\
				=&\sum_{j > M \atop 2^{j} < t_{n} \tilde{\delta}_1} I\left\{2^{j-1}<t_{n} d\left(\tilde{r}_{\oplus}(x), m_{\oplus}(x)\right)^{\beta_1/ 2} \leq 2^{j} \right\}\\
				&\quad \quad \quad \quad +\sum_{j > M \atop 2^{j} \geq t_{n} \tilde{\delta}_1} I\left\{2^{j-1}<t_{n} d\left(\tilde{r}_{\oplus}(x), m_{\oplus}(x)\right)^{\beta_1/ 2}\leq 2^{j} \right\}\\
				\leq & \sum_{j > M \atop 2^{j} < t_{n} \tilde{\delta}_1} I\left\{2^{j-1}<t_{n} d\left(\tilde{r}_{\oplus}(x), m_{\oplus}(x)\right)^{\beta_1/ 2} \leq 2^{j} \right\}
				+I\left\{2 d\left(\tilde{r}_{\oplus}(x), m_{\oplus}(x)\right)^{\beta_{1} / 2} > \tilde{\delta}_1 \right\}.
			\end{aligned}
		\end{equation}
		By the definition of $S_{j, n}(x)$, we have
		\begin{equation}\label{each shell2}
			\begin{aligned}
				I\left\{2^{j-1}<t_{n} d\left(\tilde{r}_{\oplus}(x), m_{\oplus}(x)\right)^{\beta_{1} / 2} \leq 2^{j} \right\}
				=& I\left\{\tilde{r}_{\oplus}(x) \in S_{j, n}(x) \right\}\\
				\leq& I\left\{\inf_{y \in S_{j, n}(x)} \left(\tilde{R}_n(x,y)-\tilde{R}_n(x,m_{\oplus}(x))\right)\leq 0\right\}.
			\end{aligned}
		\end{equation}
		In addition, notice that when $y \in S_{j, n}(x)$, $d(y,m_{\oplus}(x))\leq (\frac{2^j}{t_n})^{\frac{2}{\beta_1}}$. If $2^{j} < t_{n} \tilde{\delta}_1$, we have $d(y,m_{\oplus}(x)) < {\delta}_1$. Then by the assumption (A6),
		\begin{align*}
			M_{\oplus}(x,y)-M_{\oplus}\left(x,m_{\oplus}(x)\right) \geq C_1 d\left(y, m_{\oplus}(x)\right)^{\beta_1}> C_1\left(\frac{2^{2\left(j-1\right)}}{t_n^2}\right).
		\end{align*}
		Therefore,  if  $2^{j} < t_{n} \tilde{\delta}_1$,
		\begin{equation}\label{infsup2}
			\begin{aligned}
				&I\left\{\inf_{y \in S_{j, n}(x)} \left(\tilde{R}_n(x,y)-\tilde{R}_n(x,m_{\oplus}(x))\right)\leq 0\right\} \\
				\leq & I\left\{\sup _{y \in S_{j, n}(x)}\left|T_{n}(x,y)-T_{n}\left(x,m_{\oplus}(x)\right)\right| \geq C_1 \frac{2^{2(j-1)}}{t_{n}^{2}} \right\}.
			\end{aligned}
		\end{equation}
		Combine \eqref{shells2}, \eqref{each shell2} and \eqref{infsup2}, we get
		\begin{align*}
			I\left\{t_{n} d\left(\tilde{r}_{\oplus}(x), m_{\oplus}(x)\right)^{\beta_1/ 2}>2^{M}\right\}
			\leq &I\left\{2 d\left(\tilde{r}_{\oplus}(x), m_{\oplus}(x)\right)^{\beta_{1} / 2} > \tilde{\delta}_1 \right\}\\
			+\sum_{j > M \atop 2^{j} < t_{n} \tilde{\delta}_1} &I\left\{\sup _{y \in S_{j, n}(x)}\left|T_{n}(x,y)-T_{n}\left(x,m_{\oplus}(x)\right)\right| \geq C_1 \frac{2^{2(j-1)}}{t_{n}^{2}} \right\},
		\end{align*}
		where the first term of the right-hand side goes to zero for any $\tilde{\delta}_1>0$ since $d(\tilde{r}_{\oplus} (x),m_{\oplus}(x))=o(1)$. Now we focus on the second term. Obviously,
		\begin{align*}
			&\sum_{j > M \atop 2^{j} < t_{n} \tilde{\delta}_1} I\left\{\sup _{y \in S_{j, n}(x)}\left|T_{n}(x,y)-T_{n}\left(x,m_{\oplus}(x)\right)\right| \geq C_1 \frac{2^{2(j-1)}}{t_{n}^{2}} \right\}\\
			\leq & \sum_{j > M \atop 2^{j} < t_{n} \tilde{\delta}_1} \frac{t_n^2}{C_1 2^{2\left(j-1\right)}} \sup _{y \in S_{j,n}}\left|T_{n}(x,y)-T_{n}\left(x,m_{\oplus}(x)\right)\right|.
		\end{align*}
		As mentioned above, for each $j$ in the sum, we have $d(y,m_{\oplus}(x)) \leq (\frac{2^j}{t_n})^{\frac{2}{\beta_1}} < {\delta}_1$, then \eqref{maximal inequality2} holds with $\delta=(\frac{2^j}{t_n})^{\frac{2}{\beta_1}}$.Therefore, the sum is bounded by
		$$
		4 c_1 C_1^{-1} \sum_{j > M \atop 2^{j} < t_{n} \tilde{\delta}_1} \frac{2^{2 j\left(1-\beta_1\right) / \beta_1}}{t_{n}^{2\left(1-\beta_1\right) / \beta_1}}s_n^{-\frac{1}{2} \frac{\log \left(1-\alpha\right)}{\log (\alpha)} \frac{\pi}{p}}
		\leq 4 c_1 C_1^{-1} \sum_{j > M}\left(\frac{1}{4^{\left(\beta_1-1\right) / \beta_1}}\right)^{j},
		$$
		which converges since $\beta_1>1$. Thus, for some $M>0$, we have
		$$
		d\left(\tilde{r}_{\oplus}(x), m_{\oplus}(x)\right) \leq 2^{2 M / \beta_1} s_n^{-\frac{1}{2} \frac{\log \left(1-\alpha\right)}{\log (\alpha)} \frac{\pi}{p} \frac{1}{\beta_1-1}}
		$$
		for large enough $n$.
	\end{proof}
	
	Before proving the convergence rate of the variance term, we make some notes on the related literature regarding infinite order U-processes. From \eqref{RFWLCFR--2} and \eqref{P-RFWLCFR--2}, let
	\begin{equation*}
		\begin{aligned}
			\hat{R}_{n}(x,y)
			&= \binom{n}{s_n}^{-1} \sum_{k} E_{\xi \sim \Xi}\Bigg\{\frac{1}{N(L(x;\mathcal{D}_n^k,\xi))} \sum_{i: X_{i} \in L(x;\mathcal{D}_n^k,\xi)}d^2\left(Y_{i},y\right)\Bigg\},\\
			\tilde{R}_{n}(x,y)& =E\Bigg\{\frac{1}{N(L(x;\mathcal{D}_n^k,\xi))} \sum_{i: X_{i} \in L(x;\mathcal{D}_n^k,\xi)}d^2\left(Y_{i},y\right)\Bigg\}.
		\end{aligned}
	\end{equation*}
	By Theorem 3.2.5 of \cite{van1996weak},  the key to determining the convergence rate is to establish the upper bound for the expectation of the local maximum value of the centered process $\big\{(\hat{R}_{n}-\tilde{R}_{n})(x, y)-(\hat{R}_{n}-\tilde{R}_{n})(x, \tilde{r}_{\oplus}(x))\big\}$. Recall that $\hat{R}_{n}$ is an infinite order U-statistic for any fixed $y \in \Omega$ if Fr\'echet trees are symmetric.  To figure out such an upper bound, the maximal inequality of the infinite order U-process is the most important tool. Some related papers have studied the maximal inequality of  U-processes. \cite{sherman1994maximal} established the maximal inequality for degenerate U-processes of arbitrary order based on moment inequalities. \cite{arcones1993limit} obtained a similar maximal inequality by exponential inequalities, which is stronger than that of \cite{sherman1994maximal}. However, both results are limited to U-processes of fixed order.  \cite{heilig2001limit} first extended the maximal inequality of \cite{sherman1994maximal} to infinite order U-processes by the complete sign-symmetrization technique. Unfortunately, they are unable to extend the maximal inequality of \cite{arcones1993limit} to the infinite order case because such results rely on a symmetrization inequality of  \cite{de1992decoupling} which incurs upper bounds that grow with the order much too quickly. \cite{chen2020jackknife} gave a local maximal inequality for degenerate infinite order U-processes in the form of uniform entropy integrals. \cite{heilig1997empirical} discovered a stronger maximal inequality by the partial sign-symmetrization technique. Our theoretical studies will adopt this inequality to establish the convergence rate of the variance term.

	\begin{proof}[Proof of Lemma~\ref{variance rate}]
		We consider the expressions of \eqref{RFWLCFR--2} and \eqref{P-RFWLCFR--2}. For the sake of simplicity in notation, let
		$$h_n\left(Z_{i_{k,1}}, \ldots, Z_{i_{k,s_n}}, y\right)=E_{\xi \sim \Xi}\Bigg\{\frac{1}{N(L(x;\mathcal{D}_n^k,\xi))} \sum_{i: X_{i} \in L(x;\mathcal{D}_n^k,\xi)}d^2\left(Y_{i},y\right)\Bigg\}
		$$
		where $Z_i=(X_{i},Y_{i})$ and $\mathcal{D}_n^k=\left(Z_{i_{k,1}}, \ldots, Z_{i_{k,s_n}}\right)$.\\
		Then
		$$
		\hat{r}_{\oplus}(x)=\underset{y \in \Omega}{\argmin}\hat{R}_{n}(x,y)
		=\underset{y \in \Omega}{\argmin} \binom{n}{s_n}^{-1} \sum_{1 \leq i_{1} < \ldots<i_{s_n} \leq n} h_n\left(Z_{i_{1}}, \ldots, Z_{i_{s_n}}, y\right);
		$$
		$$
		\tilde{r}_{\oplus}(x)=\underset{y \in \Omega}{\argmin}\tilde{R}_{n}(x,y)=\underset{y \in \Omega}{\argmin}E h_n\left(Z_{1}, \ldots, Z_{s_n}, y\right).
		$$
		Consider the centered process $V_n(x,y)=\hat{R}_{n}(x,y)-\tilde{R}_{n}(x,y)$, and we have
		\begin{align*}
			&\left| V_{n}(x,y)-V_{n}\left(x,\tilde{r}_{\oplus}(x)\right)\right| \\
			= &\Bigg| \binom{n}{s_n}^{-1} \sum_{1 \leq i_{1} < \ldots<i_{s_n} \leq n} h_n\left(Z_{i_{1}}, \ldots, Z_{i_{s_n}}, y\right)-Eh_n\left(Z_{1}, \ldots, Z_{s_n}, y\right)  \\& -
			\binom{n}{s_n}^{-1} \sum_{1 \leq i_{1} < \ldots<i_{s_n} \leq n} h_n\left(Z_{i_{1}}, \ldots, Z_{i_{s_n}},\tilde{r}_{\oplus}(x) \right)+Eh_n\left(Z_{1}, \ldots, Z_{s_n}, \tilde{r}_{\oplus}(x)\right)
			\Bigg| \\
			=&\Bigg| \binom{n}{s_n}^{-1} \sum_{1 \leq i_{1} < \ldots<i_{s_n} \leq n} \left\{h_n\left(Z_{i_{1}}, \ldots, Z_{i_{s_n}}, y\right)-h_n\left(Z_{i_{1}}, \ldots, Z_{i_{s_n}},\tilde{r}_{\oplus}(x) \right)\right\} \\& -E\left\{h_n\left(Z_{1}, \ldots, Z_{s_n}, y\right)-h_n\left(Z_{1}, \ldots, Z_{s_n}, \tilde{r}_{\oplus}(x)\right) \right\} \Bigg|.
		\end{align*}
		Let
		\begin{align*}
			H_n(Z_{i_{1}}, \ldots, Z_{i_{s_n}}, y)=h_n\left(Z_{i_{1}}, \ldots, Z_{i_{s_n}}, y\right)-h_n\left(Z_{i_{1}}, \ldots, Z_{i_{s_n}},\tilde{r}_{\oplus}(x) \right).
		\end{align*}
		Then
		\begin{align*}
			&\left| V_{n}(x,y)-V_{n}\left(x,\tilde{r}_{\oplus}(x)\right)\right|\\
			=&\left| \binom{n}{s_n}^{-1} \sum_{1 \leq i_{1} < \ldots<i_{s_n} \leq n} H_n(Z_{i_{1}}, \ldots, Z_{i_{s_n}}, y) -E \left\{H_n(Z_{1}, \ldots, Z_{s_n}, y)\right\} \right|.
		\end{align*}
		Next, to control the $\left| V_{n}(x,y)-V_{n}\left(x,\tilde{r}_{\oplus}(x)\right)\right| $ uniformly over small $d\left(y, \tilde{r}_{\oplus}(x)\right)$, we consider the function class $$
		\mathcal{H}_{\delta}=\left\{H_n(z_{1}, \ldots, z_{s_n}, y): d\left(y, \tilde{r}_{\oplus}(x)\right)<\delta\right\}.
		$$ It's not hard to find  $	\left| V_{n}(x,y)-V_{n}\left(x,\tilde{r}_{\oplus}(x)\right)\right|$ is a centered infinite order U-process with index set $\mathcal{H}_{\delta}$. 
		
		Since
		\begin{align*}
			&\left| H_n(Z_{i_{k,1}}, \ldots, Z_{i_{k,s_n}}, y) \right| \\
			=&\left| E_{\xi \sim \Xi}\left[\frac{1}{N(L(x;\mathcal{D}_n^k,\xi))} \sum_{i: X_{i} \in L(x;\mathcal{D}_n^k,\xi)}\left\{d^2\left(Y_{i},y\right)-d^2\left(Y_{i},\tilde{r}_{\oplus}(x)\right)\right\}\right]\right|\\
			\leq &E_{\xi \sim \Xi}\left\{\frac{1}{N(L(x;\mathcal{D}_n^k,\xi))} \sum_{i: X_{i} \in L(x;\mathcal{D}_n^k,\xi)}\left| d\left(Y_{i},y\right)-d\left(Y_{i},\tilde{r}_{\oplus}(x)\right)\right| \left| d\left(Y_{i},y\right)+d\left(Y_{i},\tilde{r}_{\oplus}(x)\right)\right|\right\}\\
			\leq & 2 \operatorname{diam}(\Omega) d(y,\tilde{r}_{\oplus}(x)).
		\end{align*}
		An envelope function for $\mathcal{H}_{\delta}$ is $G_{\delta}(z_{1}, \ldots, z_{s_n})=2 \operatorname{diam}(\Omega) \delta$,
		and  the proof of Theorem 4.7 of \cite{heilig1997empirical} gives that, for small enough $\delta$ and any $\epsilon \in (0,1]$,
		\begin{align*}
			&E\left\{\sup _{d\left(y, \tilde{r}_{\oplus}(x)\right)<\delta}\left|V_{n}(x,y)-V_{n}\left(x,\tilde{r}_{\oplus}(x)\right)\right|\right\} \\
			\leq & 2 K_{1} \varepsilon+K_{2} G_{\delta} \sum_{j=1}^{s_n} E \left\{n^{-1} \log N\left(\varepsilon / s_n, d_{j}, \mathcal{H}_{\delta}\right)\right\}^{1 / 2}
		\end{align*}
		where $K_1, K_2$ are absolute constants.\\
		By the assumption (A7), for small $\delta$,
		$$\sum_{j=1}^{s_n} E\left\{\log N\left(\varepsilon / s_n, d_{j}, \mathcal{H}_{\delta}\right)\right\}^{1 / 2} \leq s_n(V \log s_n+\log A-V \log \varepsilon)^{1 / 2}.$$
		Then
		\begin{align}\label{maximal inequality}
			E\left\{\sup _{d\left(y, \tilde{r}_{\oplus}(x)\right)<\delta}\left|V_{n}(x,y)-V_{n}\left(x,\tilde{r}_{\oplus}(x)\right)\right|\right\}\leq c_2 \delta \frac{s_n (logs_n)^{\frac{1}{2}}}{n^{1/2}}
		\end{align}
		for some constant $c_2>0$.
		
		Now, set $r_{n}=(\frac{n}{s_n^2 logs_n})^{\frac{\beta_{2}}{4\left(\beta_{2}-1\right)}}$ and define
		\begin{align*}
			S_{j, n}(x)=\left\{y : 2^{j-1}<r_{n} d\left(y,\tilde{r}_{\oplus}(x)\right)^{\beta_{2} / 2} \leq 2^{j}\right\}.
		\end{align*}
		Choose $\delta_{2}$ satisfying the assumption (A8) and such that the assumption (A7) is satisfied for any $\delta<\delta_{2}$. Set $\tilde{\delta}_2:=\left(\delta_{2} \right)^{\beta_{2} / 2}$. For any integer $M$,
		\begin{equation}\label{shells}
			\begin{aligned}
				&P\left\{r_{n} d\left(\hat{r}_{\oplus}(x), \tilde{r}_{\oplus}(x)\right)^{\beta_2 / 2}>2^{M}\right\}\\	
				=&\sum_{j > M \atop 2^{j} < r_{n} \tilde{\delta}_2} P\left\{2^{j-1}<r_{n} d\left(\hat{r}_{\oplus}(x),\tilde{r}_{\oplus}(x)\right)^{\beta_{2} / 2} \leq 2^{j} \right\}\\
				& \quad \quad \quad \quad \quad \quad \quad \quad \quad +\sum_{j > M \atop 2^{j} \geq r_{n} \tilde{\delta}_2} P\left\{2^{j-1}<r_{n} d\left(\hat{r}_{\oplus}(x),\tilde{r}_{\oplus}(x)\right)^{\beta_{2} / 2} \leq 2^{j} \right\}\\
				\leq & \sum_{j > M \atop 2^{j} < r_{n} \tilde{\delta}_2} P\left\{2^{j-1}<r_{n} d\left(\hat{r}_{\oplus}(x),\tilde{r}_{\oplus}(x)\right)^{\beta_{2} / 2} \leq 2^{j} \right\}
				+ P\left\{2 d\left(\hat{r}_{\oplus}(x),\tilde{r}_{\oplus}(x)\right)^{\beta_{2} / 2} > \tilde{\delta}_2 \right\}.
			\end{aligned}
		\end{equation}
		By the definition of $S_{j, n}(x)$, we have
		\begin{equation}\label{each shell}
			\begin{aligned}
				P\left\{2^{j-1}<r_{n} d\left(\hat{r}_{\oplus}(x),\tilde{r}_{\oplus}(x)\right)^{\beta_{2} / 2} \leq 2^{j} \right\}
				= & P\left\{\hat{r}_{\oplus}(x) \in S_{j, n}(x) \right\}\\
				\leq &  P\left\{\inf_{y \in S_{j, n}(x)} \left(\hat{R}_n(x,y)-\hat{R}_n(x,\tilde{r}_{\oplus}(x))\right)\leq 0\right\}.
			\end{aligned}
		\end{equation}
		In addition, notice that when $y \in S_{j, n}(x)$, $d(y,\tilde{r}_{\oplus}(x))\leq (\frac{2^j}{r_n})^{\frac{2}{\beta_2}}$. If  $2^{j} < r_{n} \tilde{\delta}_2$, we have $d(y,\tilde{r}_{\oplus}(x)) < {\delta}_2$. Then by the assumption (A8), for large enough $n$,
		\begin{align*}
			\tilde{R}_{n}(x,y)-\tilde{R}_{n}\left(x,\tilde{r}_{\oplus}(x)\right) \geq C_{2} d\left(y, \tilde{r}_{\oplus}(x)\right)^{\beta_{2}}> C_2\left(\frac{2^{2\left(j-1\right)}}{r_n^2}\right).
		\end{align*}
		Therefore,  if  $2^{j} < r_{n} \tilde{\delta}_2$,
		\begin{equation}\label{infsup}
			\begin{aligned}
				&P\left\{\inf_{y \in S_{j, n}(x)} \left\{\hat{R}_n(x,y)-\hat{R}_n(x,\tilde{r}_{\oplus}(x))\right\}\leq 0\right\} \\ \leq & P\left\{\sup _{y \in S_{j, n}(x)}\left|V_{n}(x,y)-V_{n}\left(x,\tilde{r}_{\oplus}(x)\right)\right| \geq C_2 \frac{2^{2(j-1)}}{r_{n}^{2}} \right\}.
			\end{aligned}
		\end{equation}
		Combine \eqref{shells}, \eqref{each shell} and \eqref{infsup}, we get
		\begin{align*}
			P\left\{r_{n} d\left(\hat{r}_{\oplus}(x), \tilde{r}_{\oplus}(x)\right)^{\beta_2 /  2}>2^{M}\right\}
			\leq & P\left\{2 d\left(\tilde{r}_{\oplus}(x), \hat{r}_{\oplus}(x)\right)^{\beta_{2} / 2}>\tilde{\delta}_2\right\} \\
			+\sum_{j > M \atop 2^{j} < r_{n} \tilde{\delta}_2} &P\left\{\sup _{y \in S_{j, n}(x)}\left|V_{n}(x,y)-V_{n}\left(x,\tilde{r}_{\oplus}(x)\right)\right| \geq C_2 \frac{2^{2(j-1)}}{r_{n}^{2}} \right\}.
		\end{align*}
		where the first term of the right-hand side goes to zero for any $\tilde{\delta}_2>0$ by Lemma~\ref{variance consistency}. Now we focus on the second term. By Markov's inequality,
		\begin{align*}
			&\sum_{j > M \atop 2^{j} < r_{n} \tilde{\delta}_2} P\left\{\sup _{y \in S_{j, n}(x)}\left|V_{n}(x,y)-V_{n}\left(x,\tilde{r}_{\oplus}(x)\right)\right| \geq C_2 \frac{2^{2(j-1)}}{r_{n}^{2}} \right\}\\
			\leq & \sum_{j > M \atop 2^{j} < r_{n} \tilde{\delta}_2} \frac{r_n^2}{C_2 2^{2\left(j-1\right)}} E\left\{\sup _{y \in S_{j,n}(x)}\left|V_{n}(x,y)-V_{n}\left(x,\tilde{r}_{\oplus}(x)\right)\right|\right\}.
		\end{align*}
		As mentioned above, for each $j$ in the sum, we have $d(y,\tilde{r}_{\oplus}(x)) \leq (\frac{2^j}{r_n})^{\frac{2}{\beta_2}} < {\delta}_2$, then \eqref{maximal inequality} holds with $\delta=(\frac{2^j}{r_n})^{\frac{2}{\beta_2}}$. Therefore, the sum is bounded by
		$$
		4 c_2 C_2^{-1} \sum_{j > M \atop 2^{j} < r_{n} \tilde{\delta}_2} \frac{2^{2 j\left(1-\beta_{2}\right) / \beta_{2}}}{r_{n}^{2\left(1-\beta_{2}\right) / \beta_{2}}}  \frac{s_n (logs_n)^{\frac{1}{2}}}{n^{1/2}} < 4 c_2 C_2^{-1} \sum_{j > M}\left(\frac{1}{4^{\left(\beta_{2}-1\right) / \beta_{2}}}\right)^{j}.
		$$
		Because $\beta_2>1$, the last series converges, and hence this probability can be made small by choosing $M$ large. Hence
		\begin{align*}
			d\left( \hat{r}_{\oplus}(x), \tilde{r}_{\oplus}(x)\right)=O_{p}\left(r_{n}^{-2 / \beta_{2}}\right)=O_{p}\left(\left(\frac{s_n^2 logs_n}{n}\right)^{\frac{1}{2\left(\beta_{2}-1\right)}}\right).
		\end{align*}
	\end{proof}

	\subsection{Proofs of Results in Section~\ref{sec4}}
	\begin{proof}[Proof of Theorem~\ref{consistency of RFWLLFR}]
		Let
		\begin{equation*}
			\hat{L}_{n}(x,y) = \ e_{1}^{\T} (\tilde{X}^{\T}A\tilde{X})^{-1}\sum_{i=1}^{n}\left(\begin{array}{c}
				1 \\
				X_{i}-x
			\end{array}\right) {\alpha}_{i}(x) d^2(Y_{i},y).
		\end{equation*}
		For a fix $x \in [0,1]^p$, by Corollary 3.2.3 of \cite{van1996weak} and the assumption (A12),  we only need to prove $\sup _{y \in \Omega} \left| \hat{L}_{n}(x,y)-M_{\oplus}(x,y) \right|$ to zero in probability. To do this, we show $\hat{L}_{n}(x,\cdot) \rightsquigarrow M_{\oplus}(x,\cdot)$ in $l^{\infty}(\Omega)$ and apply Theorem 1.3.6 of \cite{van1996weak}.  By Theorem 1.5.4 of  \cite{van1996weak}, this weak convergence is equivalent to $\hat{L}_{n}(x,\cdot)$ is asymptotically tight and the marginals converge weakly. By Theorem 1.5.7 of  \cite{van1996weak}, This  asymptotically tight  is equivalent to $\hat{L}_{n}(x,y)$ is asymptotically tight in $\mathcal{R}$ for every $y \in \Omega$ and $\hat{L}_{n}(x,\cdot)$ is asymptotically uniformly $d$-equicontinuous in probability. So the proof will be finished if the following conditions hold.
		
		(i) $\hat{L}_{n}(x,y)-M_\oplus(x,y)=o_{p}(1)$ for each $y \in \Omega$,
		
		(ii) For all $\varepsilon, \eta>0$, there exists $\delta>0$ such that
		$$
		\limsup _{n} P \left\{\sup _{d\left(y_{1}, y_{2}\right)<\delta}\left|\hat{L}_{n}\left(x,y_{1}\right)-\hat{L}_{n}\left(x,y_{2}\right)\right|>\varepsilon\right\}<\eta.
		$$
		
		First, prove (i):
		By the fact about local linear estimators,
		$$
		e_{1}^{\T} (\tilde{X}^{\T}A\tilde{X})^{-1}\left[\sum_{i=1}^{n}\left(\begin{array}{c}
			1 \\
			X_{i}-x
		\end{array}\right) {\alpha}_{i}(x) E\left\{d^2(Y,y) \mid X=x\right\}\right]=E\left\{d^2(Y,y) \mid X=x\right\}.
		$$
		So we have
		\begin{align*}
			&\hat{L}_n (x,y)-M_{\oplus}(x,y) \\
			=  &e_{1}^{\T} (\tilde{X}^{\T}A\tilde{X})^{-1}\left(\sum_{i=1}^{n}\left(\begin{array}{c}
				1 \\
				X_{i}-x
			\end{array}\right) {\alpha}_{i}(x) \left[d^2(Y_i,y)-E\left\{d^2(Y,y) \mid X=x\right\}\right] \right).
		\end{align*}
		Define
		$$\bar{L}_n(x,y)=e_{1}^{\T} (\tilde{X}^{\T}A\tilde{X})^{-1}\left[\sum_{i=1}^{n}\left(\begin{array}{c}
			1 \\
			X_{i}-x
		\end{array}\right) {\alpha}_{i}(x) E\left\{d^2(Y,y) \mid X=X_i\right\}\right].$$
		Then consider to decompose $\hat{L}_{n}(x,y)-M_{\oplus}(x,y)$ into a variance-type term
		\begin{align*}
			&\hat{L}_{n}(x,y)-\bar{L}_n(x,y)\\
			=&e_{1}^{\T} (\tilde{X}^{\T}A\tilde{X})^{-1}\left(\sum_{i=1}^{n}\left(\begin{array}{c}
				1 \\
				X_{i}-x
			\end{array}\right) {\alpha}_{i}(x) \left[d^2(Y_i,y)-E\left\{d^2(Y,y) \mid X=X_i\right\}\right] \right)
		\end{align*}
		and a bias-type term
		\begin{align*}
			&\bar{L}_n(x,y)-M_{\oplus}(x,y)\\
			=&e_{1}^{\T} (\tilde{X}^{\T}A\tilde{X})^{-1}\left(\sum_{i=1}^{n}\left(\begin{array}{c}
				1 \\
				X_{i}-x
			\end{array}\right) {\alpha}_{i}(x) \left[E\left\{d^2(Y,y) \mid X=X_i\right\}-E\left\{d^2(Y,y) \mid X=x\right\}\right] \right).
		\end{align*}
		Next, we will follow the proof of Theorem 1 in \cite{bloniarz2016supervised}  with a few modifications to show that each of these terms converges to zero in probability for each $y \in \Omega$.
		
		So we begin to prove $\hat{L}_{n}(x,y)-\bar{L}_n(x,y)\stackrel{p}{\rightarrow} 0$. For convenience in notation, let $\alpha_{i}(x;\mathcal{D}_{n}^b,\xi_{b})=1\left\{X_{i} \in L_{b}(x;\mathcal{D}_{n}^b,\xi_{b})\right\}$,  indicating that a training point $X_{i}$ belongs to the same leaf node as $x$ in the $b$th Fr\'echet tree trained with subsample $\mathcal{D}_{n}^b$ and random parameter $\xi_b$. We define the bandwidth matrix of the random forest to be a diagonal matrix with diagonal elements set to be the largest component-wise distances from $x$ to a training point that has a nonzero weight. Let $X_{i}=\left(X_{i}^{(1)}, \ldots, X_{i}^{(p)}\right)$ and $x=\left(x^{(1)}, \ldots, x^{(p)}\right)$. Then define
		\begin{align*}
			h_{j} =\max _{i, b}\left\{\alpha_{i}(x;\mathcal{D}_{n}^b,\xi_{b})\left|X_{i}^{(j)}-x^{(j)}\right|\right\}, \quad 
			H =\operatorname{diag}\left(1, h_{1}, \ldots, h_{p}\right).
		\end{align*}
		By the assumption (A10), the number of training points falling in a leaf node goes to infinity. Under the honesty condition, if we condition on the variables $\alpha_{i}(x;\mathcal{D}_{n}^b,\xi_{b})$, the subset
		of the training data falling in $L_{b}(x;\mathcal{D}_{n}^b,\xi_{b})$ is independent and identically distributed in the rectangle $L_{b}(x;\mathcal{D}_{n}^b,\xi_{b})$. By definition of $H$, for the training point that has nonzero weight, we have
		$$\left\|H^{-1} \left(\begin{array}{c}
			1 \\
			X_{i}-x
		\end{array}\right)\right\|_{\infty} \leq 1.
		$$
		And notice that $E\left[d^2(Y_i,y)-E\left\{d^2(Y,y) \mid X=X_i\right\}\right]=0$. Combining the previous discussion, with the weak law of large numbers we can obtain
		\begin{align*}
			\frac{1}{N(L_{b}(x;\mathcal{D}_{n}^b,\xi_{b}))} \sum_{i=1}^{n} H^{-1} \left(\begin{array}{c}
				1 \\
				X_{i}-x
			\end{array}\right) \alpha_{i}(x;\mathcal{D}_{n}^b,\xi_{b}) \left[d^2(Y_i,y)-E\left\{d^2(Y,y) \mid X=X_i\right\}\right] \stackrel{p}{\rightarrow} 0.
		\end{align*}
		Therefore, averaging over total B Fr\'echet trees, we get
		\begin{align}\label{post-term}
			\sum_{i=1}^{n}  H^{-1} \left(\begin{array}{c}
				1 \\
				X_{i}-x
			\end{array}\right) \alpha_{i}(x) \left[d^2(Y_i,y)-E\left\{d^2(Y,y) \mid X=X_i\right\}\right] \stackrel{p}{\rightarrow} 0.
		\end{align}
		Below we need to study  the  matrix $\tilde{X}^{\T}A\tilde{X}$. Since $A$ is a diagonal matrix,
		\begin{align*}
			\tilde{X}^{\T}A\tilde{X} &=\sum_{i=1}^{n} \alpha_{i}(x)
			\left(\begin{array}{c}
				1 \\
				X_{i}-x
			\end{array}\right) \left(1, \left(X_{i}-x\right)^{\T}\right) \\
			&=\frac{1}{B} \sum_{b=1}^{B}\left[\frac{1}{N(L_{b}(x;\mathcal{D}_{n}^b,\xi_{b}))} \sum_{i=1}^{n}  \alpha_{i}(x;\mathcal{D}_{n}^b,\xi_{b}) \left(\begin{array}{c}
				1 \\
				X_{i}-x
			\end{array}\right) \left(1, \left(X_{i}-x\right)^{\T}\right)\right].
		\end{align*}
		By the same argument of Theorem 1 in \cite{bloniarz2016supervised}, we can have
		\begin{align}\label{pre-item}
			e_{1}^{\T} (\tilde{X}^{\T}A\tilde{X})^{-1} H=\left(O_{p}(1), \ldots, O_{p}(1)\right).
		\end{align}
		which needs the assumption (A9), (A10) and the honesty condition. Combining \eqref{post-term} and \eqref{pre-item}, then
		\begin{align}\label{var-type}
			\hat{L}_{n}(x,y)-\bar{L}_n(x,y) \stackrel{p}{\rightarrow} 0.
		\end{align}
		Now we turn to prove $\bar{L}_n(x,y)-M_{\oplus}(x,y) \stackrel{p}{\rightarrow} 0$. Similarly, we have
		\begin{align*}
			\bar{L}_n(x,y)-M_{\oplus}(x,y)
			=&e_{1}^{\T} (\tilde{X}^{\T}A\tilde{X})^{-1}H\Bigg(\sum_{i=1}^{n} H^{-1}\left(\begin{array}{c}
				1 \\
				X_{i}-x
			\end{array}\right) {\alpha}_{i}(x) \\
		&\quad \quad \quad \quad \left[E\left\{d^2(Y,y) \mid X=X_i\right\}-E\left\{d^2(Y,y) \mid X=x\right\}\right] \Bigg).
		\end{align*}
		By the assumption (A11) and the definition of $H$, we can have
		$$
		\sum_{i=1}^{n} H^{-1}\left(\begin{array}{c}
			1 \\
			X_{i}-x
		\end{array}\right) {\alpha}_{i}(x) \left[E\left\{d^2(Y,y) \mid X=X_i\right\}-E\left\{d^2(Y,y) \mid X=x\right\}\right] \stackrel{p}{\rightarrow} 0.
		$$
		Since $e_{1}^{\T} (\tilde{X}^{\T}A\tilde{X})^{-1} H=\left(O_{p}(1), \ldots, O_{p}(1)\right)$, we get
		\begin{align}\label{bias-type}
			\bar{L}_n(x,y)-M_{\oplus}(x,y) \stackrel{p}{\rightarrow} 0.
		\end{align}
		Hence combine \eqref{var-type} and \eqref{bias-type},  it follows that for each $y \in \Omega$,
		$$\hat{L}_{n}(x,y)-M_\oplus(x,y)=o_{p}(1).$$
		
		Then (ii): For any $y_{1}$, $y_{2}\in\Omega$,
		\begin{align*}
			&\left|\hat{L}_{n}\left(x,y_{1}\right)-\hat{L}_{n}\left(x,y_{2}\right)\right| \\
			= & \left| e_{1}^{\T} (\tilde{X}^{\T}A\tilde{X})^{-1}\left[\sum_{i=1}^{n}\left(\begin{array}{c}
				1 \\
				X_{i}-x
			\end{array}\right) {\alpha}_{i}(x) \left\{d^2(Y_i,y_1)-d^2(Y_i,y_2)\right\} \right]\right|\\
			\leq  &\sum_{i=1}^{n}\left|e_{1}^{\T} (\tilde{X}^{\T}A\tilde{X})^{-1}\left(\begin{array}{c}
				1 \\
				X_{i}-x
			\end{array}\right) {\alpha}_{i}(x)\right|\left|d\left(Y_{i}, y_{1}\right)-d\left(Y_{i}, y_{2}\right)\right| \left|d\left(Y_{i}, y_{1}\right)+d\left(Y_{i}, y_{2}\right)\right|\\
			\leq & 2 \operatorname{diam}(\Omega) d\left(y_{1}, y_{2}\right)\sum_{i=1}^{n}\left|e_{1}^{\T} (\tilde{X}^{\T}A\tilde{X})^{-1}\left(\begin{array}{c}
				1 \\
				X_{i}-x
			\end{array}\right) {\alpha}_{i}(x)\right| \\
			=& 2 \operatorname{diam}(\Omega) d\left(y_{1}, y_{2}\right)\sum_{i=1}^{n}\left|e_{1}^{\T} (\tilde{X}^{\T}A\tilde{X})^{-1}HH^{-1}\left(\begin{array}{c}
				1 \\
				X_{i}-x
			\end{array}\right)  {\alpha}_{i}(x)\right| \\
			\leq & 2 \operatorname{diam}(\Omega) d\left(y_{1}, y_{2}\right)\sum_{i=1}^{n}{\alpha}_{i}(x)\left\|e_{1}^{\T} (\tilde{X}^{\T}A\tilde{X})^{-1}H\right\| \left\|H^{-1}\left(\begin{array}{c}
				1 \\
				X_{i}-x
			\end{array}\right) \right\|  \\
			=&2 \operatorname{diam}(\Omega) d\left(y_{1}, y_{2}\right)O_p(1)\\
			=&O_{p}\left(d\left(y_{1}, y_{2}\right)\right),
		\end{align*}
		where the $O_{p}$ term is independent of $y_{1}$ and $y_{2}$.
		Hence
		$$
		\sup _{d\left(y_{1}, y_{2}\right)<\delta}\left|\hat{L}_{n}\left(x,y_{1}\right)-\hat{L}_{n}\left(x,y_{2}\right)\right|=O_{p}(\delta),
		$$
		which can deduce (ii).  So, $d(\hat{l}_{\oplus}(x),m_{\oplus}(x))=o_{p}(1)$.
		
	\end{proof}
	
	\subsection{Proofs of Results in Appendix ~\ref{sec E}}
	
	\begin{proof}[Proof of Theorem~\ref{Mm normal}]
		Now we generalize the proof of Theorem 2.3 of \cite{bose2018u}. The norm involved in this part is the spectral norm for matrices and the Euclidean norm for vectors.
		
		Since $f_n\left(x, \theta\right)$ is  convex in $\theta$, for all $\alpha$,$\beta$,
		$$
		\begin{aligned}
			&f_n(x, \alpha)+(\beta-\alpha)^{T} g_n(x, \alpha) \leq f_n(x, \beta), \\
			&f_n(x, \beta)+(\alpha-\beta)^{T} g_n(x, \beta) \leq f_n(x, \alpha).
		\end{aligned}
		$$
		Hence,
		\begin{equation}\label{4.1}
			(\beta-\alpha)^{T} g_n(x, \alpha) \leq f_n(x, \beta)-f_n(x, \alpha) \leq(\beta-\alpha)^{T} g_n(x, \beta).
		\end{equation}
		Then
		\begin{equation}\label{4.2}
			\begin{aligned}
				0 \leq f_n(x, \beta)-f_n(x, \alpha) -(\beta-\alpha)^{T} g_n(x, \alpha) 
				\leq(\beta-\alpha)^{T}[g_n(x, \beta)-g_n(x, \alpha)].
			\end{aligned}
		\end{equation}
		By the assumption (ii), we know $E g_n\left(Z_{1}, Z_{2}, \ldots, Z_{m_n}, \theta\right) < \infty$ from (\ref{4.1}).
		Moreover, based on (\ref{4.2}), note that $E g_n\left(Z_{1}, Z_{2}, \ldots, Z_{m_n}, \theta\right)$ serves as a subgradient of $Q_n(\theta)$. Now, when $Q_n(\theta)$ is differentiable, it follows that
		$$
		\nabla Q_n(\theta_n)=E g_n\left(Z_{1}, Z_{2}, \ldots, Z_{m_n}, \theta_n\right)=0.
		$$
		Let $W_n=\left\{w_n=\left(i_{1}, i_{2}, \ldots, i_{m_n}\right): 1 \leq i_{1}<i_{2} \cdots<i_{m_n} \leq n\right\}$. For any $w_n \in W_n$, let
		$Z_{w_n}=\left(Z_{i_{1}}, \ldots, Z_{i_{m_n}}\right)$ and
		$$
		Z_{n, w_n}=f_n\left(Z_{w_n}, n^{-1 / 2}\Lambda_n^{\frac{1}{2}} \alpha+\theta_n\right)-f_n\left(Z_{w_n}, \theta_n\right)-n^{-1 / 2}\alpha^T\Lambda_n^{\frac{1}{2}}  g_n\left(Z_{w_n}, \theta_n\right)
		$$
		where
		$
		\Lambda_n=m_n^2H_n^{-1}K_nH_n^{-1}
		$.
		Note that $V_{n}=\binom{n}{m_n}^{-1} \sum_{w_n \in W_n} Z_{n, w_n}$ is an infinite order $U$-statistic. Using (\ref{4.2}), it follows that
		$$
		\begin{aligned}
			\Var\left(V_n\right) & \leq \frac{m_n}{n} \Var\left(Z_{n, w_n}\right) \\
			& \leq \frac{m_n}{n} E Z_{n, w_n}^{2} \\
			& \leq \frac{m_n}{n^{2}} E\left\{\alpha^{T}\Lambda_n^{\frac{1}{2}} \left[g_n(Z_{w_n}, n^{-1 / 2}\Lambda_n^{\frac{1}{2}} \alpha+\theta_n)-g_n(Z_{w_n}, \theta_n)\right]\right\}^{2}\\
			& \leq \frac{m_n}{n^{2}} E\left[\Vert\alpha\Vert \Vert\Lambda_n^{\frac{1}{2}} \Vert \Vert
			g_n(Z_{w_n}, n^{-1 / 2}\Lambda_n^{\frac{1}{2}} \alpha+\theta_n)-g_n(Z_{w_n}, \theta_n)\Vert\right]^{2}\\
			&= \frac{m_n\Vert\Lambda_n \Vert}{n^{2}}\Vert\alpha\Vert^2 E\left[\Vert
			g_n(Z_{w_n}, n^{-1 / 2}\Lambda_n^{\frac{1}{2}} \alpha+\theta_n)-g_n(Z_{w_n}, \theta_n)\Vert\right]^{2}.
		\end{aligned}
		$$
		By Taylor's expansion, we have
		$$g_n(Z_{w_n}, n^{-1 / 2}\Lambda_n^{\frac{1}{2}} \alpha+\theta_n)=g_n(Z_{w_n}, \theta_n)+\frac{\partial g_n(Z_{w_n}, \tilde{\theta}_n)}{\partial \theta}n^{-1 / 2}\Lambda_n^{\frac{1}{2}} \alpha,
		$$
		where $\tilde{\theta}_n$ is between $\theta_n$ and $n^{-1 / 2}\Lambda_n^{\frac{1}{2}} \alpha+\theta_n$. Then
		$$
		\begin{aligned}
			&E\left[\Vert
			g_n(Z_{w_n}, n^{-1 / 2}\Lambda_n^{\frac{1}{2}} \alpha+\theta_n)-g_n(Z_{w_n}, \theta_n)\Vert\right]^{2}\\
			=&	E\left[\Vert
			\frac{\partial g_n(Z_{w_n}, \tilde{\theta}_n)}{\partial \theta}n^{-1 / 2}\Lambda_n^{\frac{1}{2}} \alpha\Vert\right]^{2}\\
			\leq &\frac{1}{n}\Vert\Lambda_n \Vert \Vert\alpha\Vert^2	E\left[\Vert
			\frac{\partial g_n(Z_{w_n}, \tilde{\theta}_n)}{\partial \theta}\Vert\right]^{2}.
		\end{aligned}
		$$
		Since $\lambda_{\min }(H_n)\nrightarrow 0$ by the assumption (v), $\Vert H_n^{-1}\Vert = \frac{1}{\lambda_{\min }(H_n)} \nrightarrow \infty$. And due to $m_n K_n \leq \Var\left(g_n\left(Z_{1}, \ldots, Z_{m_n}, \theta_n\right)\right) < \infty$, we have $\Vert m_n K_n \Vert < \infty $. Hence,
		$$\frac{\Vert\Lambda_n \Vert}{n}=\frac{\Vert m_n^2H_n^{-1}K_nH_n^{-1}\Vert}{n}\leq \frac{m_n \Vert H_n^{-1}\Vert^2\Vert m_nK_n\Vert}{n}\rightarrow 0.$$
		And we get
		$$
		E\left[\Vert
		g_n(Z_{w_n}, n^{-1 / 2}\Lambda_n^{\frac{1}{2}} \alpha+\theta_n)-g_n(Z_{w_n}, \theta_n)\Vert\right]^{2} \rightarrow 0.
		$$
		By Markov's inequality, it follows that for each fixed $\alpha$,
		$$
		\frac{n}{\sqrt{m_n \Vert\Lambda_n \Vert}} \left(V_n-E(V_n)\right)\stackrel{P}{\rightarrow}  0.
		$$
		Specifically,
		\begin{equation}\label{4.3}
			\begin{aligned}
				&\frac{n}{\sqrt{m_n \Vert\Lambda_n \Vert}}\binom{n}{m_n} ^{-1} \sum_{w_n \in W_n}\left(Z_{n, w_n}-E Z_{n, w_n}\right) \\
				=&\frac{n}{\sqrt{m_n \Vert\Lambda_n \Vert}} \hat{Q}_{n}\left(n^{-1 / 2}\Lambda_n^{\frac{1}{2}} \alpha+\theta_n\right)-\frac{n}{\sqrt{m_n \Vert\Lambda_n \Vert}} \hat{Q}_{n}(\theta_n)-\frac{\sqrt{n}}{\sqrt{m_n \Vert\Lambda_n \Vert}}\alpha^{T} \Lambda_n^{\frac{1}{2}}U_{n}\\
				&-\frac{n}{\sqrt{m_n \Vert\Lambda_n \Vert}} Q_n\left(n^{-1 / 2}\Lambda_n^{\frac{1}{2}} \alpha+\theta_n\right)+\frac{n}{\sqrt{m_n \Vert\Lambda_n \Vert}} Q_{n}(\theta_n) \\
				\stackrel{P}{\rightarrow} & 0.
			\end{aligned}
		\end{equation}
		On the other hand, for each fixed $\alpha$, by Taylor's expansion and the  assumption (iv), (v),
		\begin{equation}\label{4.4}
			\frac{n}{\sqrt{m_n \Vert\Lambda_n \Vert}} Q_n\left(n^{-1 / 2}\Lambda_n^{\frac{1}{2}} \alpha+\theta_n\right) \rightarrow\frac{n}{\sqrt{m_n \Vert\Lambda_n \Vert}} Q_{n}(\theta_n) +\frac{\alpha^T \Lambda_n^{\frac{1}{2}} H_n \Lambda_n^{\frac{1}{2}}\alpha}{2\sqrt{m_n \Vert\Lambda_n \Vert}}.
		\end{equation}
		The reason why there are only two items on the right side of the above formula is
		$$\frac{n\Vert \Lambda_n^{\frac{1}{2}} \Vert ^3 n^{-\frac{3}{2}}}{\sqrt{m_n \Vert \Lambda_n\Vert}}=\frac{\Vert \Lambda_n \Vert} {\sqrt{m_n}\sqrt{n}}\leq \frac{m_n \Vert H_n^{-1}\Vert^2\Vert m_nK_n\Vert}{\sqrt{m_n}\sqrt{n}}\rightarrow 0.
		$$
		By Lemma 2.1 of \cite{bose2018u}, the convergences in (\ref{4.3}) and (\ref{4.4}) are uniform on compact sets due to convexity. Thus for every $\epsilon>0$ and every $M>0$,
		\begin{equation}\label{4.5}
			\begin{aligned}
				\sup _{\Vert \alpha \Vert \leq M}&\left|\frac{n}{\sqrt{m_n \Vert\Lambda_n \Vert}} \hat{Q}_{n}\left(n^{-1 / 2}\Lambda_n^{\frac{1}{2}} \alpha+\theta_n\right)-\frac{n}{\sqrt{m_n \Vert\Lambda_n \Vert}} \hat{Q}_{n}(\theta_n)\right.\\
				&\left.-\frac{\sqrt{n}}{\sqrt{m_n \Vert\Lambda_n \Vert}}\alpha^{T} \Lambda_n^{\frac{1}{2}}U_{n}-\frac{\alpha^T \Lambda_n^{\frac{1}{2}} H_n \Lambda_n^{\frac{1}{2}}\alpha}{2\sqrt{m_n \Vert\Lambda_n \Vert}} \right|<\epsilon.
			\end{aligned}
		\end{equation}
		holds with probability at least $(1-\epsilon / 2)$ for large $n$.
		
		Define the quadratic form
		$$
		B_{n}(\alpha)=\frac{\sqrt{n}}{\sqrt{m_n \Vert\Lambda_n \Vert}}\alpha^{T} \Lambda_n^{\frac{1}{2}}U_{n}+\frac{\alpha^T \Lambda_n^{\frac{1}{2}} H_n \Lambda_n^{\frac{1}{2}}\alpha}{2\sqrt{m_n \Vert\Lambda_n \Vert}}.
		$$
		Its minimizer is $\alpha_{n}=-\Lambda_n^{-\frac{1}{2}} H_n^{-1} n^{1 / 2} U_{n}$. Since for any $c \neq 0$,
		$$
		c^{T}U_{n} =\binom{n}{m_n}^{-1} \sum_{1 \leq i_{1}< i_{2}< \ldots<i_{m_n} \leq n} c^{T}g_n\left(Z_{i_{1}}, Z_{i_{2}},\ldots, Z_{i_{m_n}}, \theta_{n}\right)
		$$
		And
		$$
		m_n \Var\left[E\left(c^{T}g_n\left(Z_{1}, \ldots, Z_{m_n}, \theta_{n}\right) \mid Z_{1}\right)\right]=	c^{T}m_n K_{n}c.
		$$
		By Rayleigh-Ritz theorem and  assumption (iv), $c^{T}m_n K_{n}c \nrightarrow 0$ for any $c \neq 0$. Utilizing Theorem 1 of \cite{peng2019asymptotic}, $c^{T}U_{n}$  is asymptotically normal. Hence, by Cram\'er-Wold device,
		$$
		\frac{\sqrt{n}}{m_n}K_n^{-\frac{1}{2}}U_n \stackrel{d}{\longrightarrow} \mathcal{N}\left(0, I \right).
		$$
		So
		$$
		\alpha_{n}=-\Lambda_n^{-\frac{1}{2}} H_n^{-1} n^{1 / 2} U_{n} \stackrel{d}{\longrightarrow} \mathcal{N}\left(0, I\right).
		$$
		The minimum value of the quadratic form is
		$$
		B_{n}\left(\alpha_{n}\right)=-n U_{n}^{T} H_n^{-1}  U_{n}/ 2\sqrt{m_n \Vert \Lambda_n \Vert}.
		$$
		Further, $\alpha_n$ is bounded in probability. So we can select an $M$ such that
		\begin{equation}\label{4.6}
			P\left(\Vert \alpha_{n}\Vert<M-1\right) \geq 1-\epsilon / 2.
		\end{equation}
		The rest of the argument is on the intersection of the two events in (\ref{4.5}) and (\ref{4.6}), which has a probability of at least $1-\epsilon$.
		
		Consider the convex function
		$$
		A_{n}(\alpha)=\frac{n}{\sqrt{m_n \Vert\Lambda_n \Vert}} \hat{Q}_{n}\left(n^{-1 / 2}\Lambda_n^{\frac{1}{2}} \alpha+\theta_n\right)-\frac{n}{\sqrt{m_n \Vert\Lambda_n \Vert}} \hat{Q}_{n}(\theta_n).
		$$
		From (\ref{4.5}),
		\begin{equation}\label{4.7}
			A_{n}\left(\alpha_{n}\right) \leq B_{n}\left(\alpha_{n}\right)+\epsilon=-n U_{n}^{T} H_n^{-1}  U_{n}/ 2\sqrt{m_n \Vert \Lambda_n \Vert}+\epsilon.
		\end{equation}
		Again by using (\ref{4.5}), we know the value of $A_n(\alpha)$ is at least
		\begin{equation}\label{4.8}
			B_{n}(\alpha)-\epsilon.
		\end{equation}
		By simple calculation, the bound in (\ref{4.8}) is always strictly larger than the one in (\ref{4.7}) when
		$\Vert \alpha-\alpha_n\Vert \geq T (\epsilon \sqrt{m_n \Vert \Lambda_n\Vert})^{\frac{1}{2}}/\Vert \Lambda_n^{\frac{1}{2}} \Vert$, where $T=4\left[\lambda_{\min }(H_n)\right]^{-1 / 2}$ and $\lambda_{\min}$ denotes the minimum eigenvalue.
		
		On the other hand $A_{n}$ has the minimizer $\sqrt{n}\Lambda_n^{-\frac{1}{2}} (\hat{\theta}_n-\theta_n)$. So, using the fact that $A_{n}$ is convex, it follows that its minimizer satisfies
		\begin{equation}\label{4.9}
			\Vert \sqrt{n}\Lambda_n^{-\frac{1}{2}} (\hat{\theta}_n-\theta_n)-\alpha_{n}\Vert<T (\epsilon \sqrt{m_n \Vert \Lambda_n\Vert})^{\frac{1}{2}}/\Vert \Lambda_n^{\frac{1}{2}} \Vert.
		\end{equation}
		Note that  $\frac{( \sqrt{m_n \Vert \Lambda_n\Vert})^{\frac{1}{2}}}{\Vert \Lambda_n^{\frac{1}{2}} \Vert}   =\frac{m_n^{\frac{1}{4}}}{\Vert \Lambda_n \Vert ^{\frac{1}{4}}}< \infty$. Since (\ref{4.9}) holds with probability at least $(1-\epsilon)$ where $\epsilon$ is arbitrary, we get
		$$
		\sqrt{n}\Lambda_n^{-\frac{1}{2}} (\hat{\theta}_n-\theta_n)-\alpha_{n}=o_p(1).
		$$
		The first part of the theorem has been proved, and the second part follows from the multivariate version of the Central Limit Theorem of $U_n$.
	\end{proof}
	
	\begin{proof}[Proof of Theorem~\ref{asymptotic normal}]
		Since the assumption (A1) and assumption (A4) hold, by Lemma~\ref{variance consistency}, we have $d(\hat{r}_{\oplus}(x),\tilde{r}_{\oplus}(x))=o_{p}(1)$. Then $P(\hat{r}_{\oplus}(x) \in G) \rightarrow 1$ as $n \rightarrow \infty$.
		
		For large $n$, let
		$$u_n(x)=\phi(\tilde{r}_{\oplus}(x)),\quad \hat{u}_n(x)=\phi(\hat{r}_{\oplus}(x)).$$
		Since
		$$Ef_n(Z_{1}, Z_{2}, \ldots, Z_{s_n},u)= E h_n\left(Z_{1}, Z_{2}, \ldots, Z_{s_n}, \phi ^{-1}(u)\right)=\tilde{R}_{n}(\phi ^{-1}(u)), $$
		we have
		\begin{align*}
			u_n(x)=\underset{u \in U}{\argmin} Ef_n(Z_{1}, Z_{2}, \ldots, Z_{s_n},u)\triangleq \underset{u \in U}{\argmin}  Q_n(u).
		\end{align*}
		
		Similarly, since
		\begin{align*}
			&\binom{n}{s_n}^{-1} \sum_{1 \leq i_{1}<i_{2} <\ldots<i_{s_n} \leq n} f_n(Z_{i_1}, Z_{i_2}, \ldots, Z_{i_{s_n}},u)\\
			=& \binom{n}{s_n}^{-1} \sum_{1 \leq i_{1}<i_{2} <\ldots<i_{s_n} \leq n} h_n\left(Z_{i_1},Z_{i_2}, \ldots, Z_{i_{s_n}}, \phi ^{-1}(u)\right)\\
			=&\hat{R}_{n}(\phi ^{-1}(u)),
		\end{align*}
		we have
		\begin{align*}
			\hat{u}_n(x)=\underset{u \in U}{\argmin}\binom{n}{s_n}^{-1} \sum_{1 \leq i_{1}<i_{2} <\ldots<i_{s_n} \leq n} f_n(Z_{i_1}, Z_{i_2}, \ldots, Z_{i_{s_n}},u)\triangleq \underset{u \in U}{\argmin}  \hat{Q}_n(u).
		\end{align*}
		When we consider $M_{s_n}$-estimator $\hat{u}_n(x)$ of $u_n(x)$,  the assumptions of Theorem \ref{Mm normal} are satisfied by the assumptions (A13)--(A18). Hence by Theorem \ref{Mm normal}, we get
		$$\sqrt{n} \Lambda_n^{-1/2}\left\{\hat{u}_{n}(x)-u_{n}(x)\right\} \stackrel{d}{\longrightarrow} \mathcal{N}\left(0, I\right)$$
		where
		$$
		\Lambda_n=s_n^2H_n^{-1}K_nH_n^{-1}.
		$$	
	\end{proof}
	

	\vskip 0.2in
	\bibliography{sample}
	
\end{document}